\documentclass[11pt]{article}

\usepackage[preprint]{acl}
\usepackage{times}
\usepackage{latexsym}
\usepackage[T1]{fontenc}
\usepackage[utf8]{inputenc}
\usepackage{microtype}
\usepackage{inconsolata}
\usepackage{graphicx}

\usepackage{hyperref}
\usepackage{url}
\usepackage{booktabs}
\usepackage{subcaption}
\usepackage{booktabs}
\usepackage{enumitem}
\usepackage{multirow}
\usepackage{soul}
\usepackage{amsmath}
\usepackage{amssymb}
\usepackage{mathtools}
\usepackage{amsthm}
\usepackage[breakable]{tcolorbox}

\usepackage[scaled=.8]{beramono}
\usepackage[T1]{fontenc}

\usepackage[capitalize,noabbrev]{cleveref}

\usepackage{fontawesome5}

\definecolor{darkblue}{rgb}{0, 0, 0.5}
\hypersetup{colorlinks=true, citecolor=darkblue, linkcolor=darkblue, urlcolor=darkblue}

\newcommand{\authorsep}[0]{\ \ }

\newcommand{\parahead}[1]{\vspace{1.5mm}\noindent\textbf{#1}.\ }
\setlist[itemize]{noitemsep,topsep=0pt}
\setlist[enumerate]{noitemsep,topsep=0pt}

\title{
\begin{minipage}{0.12\textwidth}
\includegraphics[width=\textwidth]{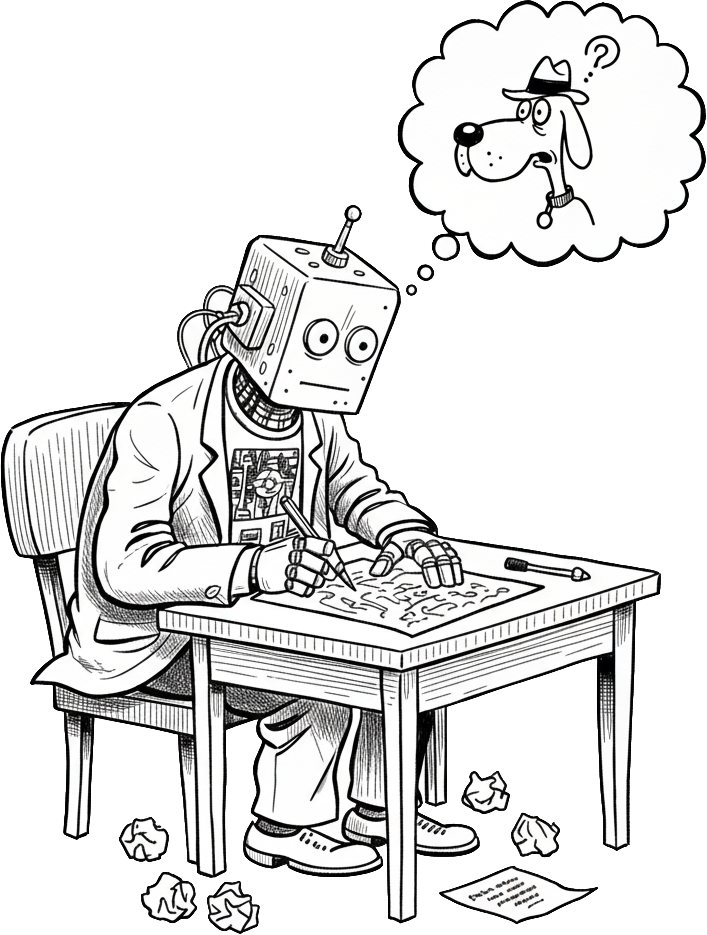}
\end{minipage}%
\hfill
\begin{minipage}{0.85\textwidth}
\raggedright
Learning to Think Like a Cartoon Captionist: \mbox{Incongruity--Resolution} Supervision for Multimodal \mbox{Humor} Understanding
\end{minipage}
}

\author{Hatice Merve Vural$^{1,\dagger}$ \authorsep Doga Kukul$^{1,2}$ \authorsep Ege Erdem Ozlu$^1$ \authorsep Demir Ekin Arikan$^1$ \\ 
 \textbf{Bob Mankoff}$^3$ \authorsep \textbf{Erkut Erdem}$^{2,4}$ \authorsep \textbf{Aykut Erdem}$^{1,2}$\\ 
 $^1$Ko\c{c} University, Istanbul, Turkey \authorsep 
$^2$KUIS AI Center, Istanbul, Turkey \\
$^3$Air Mail and Cartoon Collections \authorsep  $^4$Hacettepe University, Ankara, Turkey\\
$^\dagger$\small{\textbf{Correspondence:} \href{mailto:havural24@ku.edu.tr}{\texttt{havural24@ku.edu.tr}}}}

\begin{document}

\maketitle

\begin{center}
    \faGithub\ \href{https://cyberiada.github.io/NYCC-Thinking/}{\textbf{Project page}}
\end{center}

\begin{abstract}
Humor is one of the few cognitive tasks where getting the reasoning right matters as much as getting the answer right. While recent work evaluates humor understanding on benchmarks such as the New Yorker Cartoon Caption Contest (NYCC), it largely treats it as black-box prediction, overlooking the structured reasoning processes underlying humor comprehension. We introduce IRS (Incongruity--Resolution Supervision), a framework that decomposes humor understanding into three components: \emph{Incongruity Modeling}, which identifies mismatches in the visual scene; \emph{Resolution Modeling}, which constructs coherent reinterpretations of these mismatches; and \emph{Preference Alignment}, which evaluates candidate interpretations under human judgments. Grounded in incongruity-resolution theory and expert captionist practice, IRS supervises intermediate reasoning process through structured traces that make the path from visual perception to humorous interpretation explicit and learnable. Across 7B, 32B, and 72B models on NYCC, IRS improves performance across caption matching and ranking, with IRS-72B achieving the strongest model performance on ranking (76.10\%), surpassing both non-expert human performance and all evaluated open- and closed-source multimodal baselines. Zero-shot transfer further shows that IRS learns generalizable reasoning patterns.
\end{abstract}
\section{Introduction}
\label{introduction}
Humor is a demanding form of human intelligence that requires visual perception, cultural knowledge, and creative reasoning \cite{kazemi-etal-2025-big}. Unlike standard benchmarks, humor depends not only on choosing an answer but on identifying a mismatch between expectation and observation, and resolving it in a coherent yet surprising way. This incongruity-resolution dynamic, long studied in cognitive science \cite{suls1972twostage,attardo1994linguistic}, suggests that humor understanding is fundamentally a structured reasoning problem, yet most computational approaches treat it as a prediction task, training models to select or rank captions without modeling the interpretive steps that make a caption funny.
Cartoon captioning makes this gap concrete. A strong caption does not merely describe a scene; it identifies a salient visual tension and reframes it into a meaningful punchline. This requires knowing what the incongruity is, how it can be resolved, and why one resolution is funnier than another; capabilities that current multimodal language models only partially capture, as reflected in their gap to human performance on humor benchmarks \cite{hessel-etal-2023-androids,zhou2025bridgingcreativityunderstandinggap}.

\emph{The New Yorker Cartoon Caption Contest} (NYCC) provides a structured setting for this problem, pairing each image with thousands of crowd-submitted captions, expert editorial curation, and large-scale audience judgments and creating a rare alignment between visual input, linguistic creativity, expert reasoning, and human preferences. While prior work treats NYCC as a classification or ranking benchmark, it also offers insight into the reasoning processes underlying visual humor.

In this work, we propose \emph{Incongruity-Resolution Supervision (IRS)}, a framework that decomposes humor understanding into three components: \emph{Incongruity Modeling}, which identifies mismatches in the visual scene; \emph{Resolution Modeling}, which constructs coherent reinterpretations of those mismatches; and \emph{Preference Alignment}, which evaluates candidate interpretations under human judgments. Grounded in incongruity-resolution theory and informed by expert analyses of professional captionist practice \cite{wood2024your,mankoff2002nakedcartoonist}, IRS uses \emph{captionist reasoning traces} to supervise intermediate reasoning steps.
Across 7B, 32B, and 72B 
models, IRS improves performance on NYCC matching and ranking tasks. Our largest model approaches expert-level ranking performance, and zero-shot transfer to external humor benchmarks suggests that IRS learns generalizable reasoning patterns rather than dataset-specific heuristics.

Our contributions are as follows:
\begin{itemize}[leftmargin=*]
\item We introduce Incongruity-Resolution Supervision (IRS), a theoretically grounded framework that decomposes humor understanding into three explicit, learnable stages bridging cognitive theory and multimodal learning. IRS is implemented through \emph{captionist reasoning traces}, structured supervision signals derived from expert practice that make intermediate reasoning over visual humor explicit and learnable.
\item We show that IRS substantially improves both performance and reasoning quality across model scales on NYCC, and that the learned reasoning patterns transfer zero-shot to out-of-domain humor benchmarks, establishing explicit reasoning supervision as a scalable and generalizable approach to creative multimodal understanding.
\end{itemize}

\section{Related Work}
\label{related_work}

\parahead{Computational Humor}
Humor is commonly modeled through incongruity-resolution theory \cite{suls1972twostage, attardo1994linguistic}, which frames it as the violation of an expectation followed by a coherent reinterpretation. Related perspectives, including script opposition \cite{attardo1991script}, frame shifting \cite{coulson2001getting}, and Theory of Mind accounts \cite{samson2012influence}, further highlight its structured cognitive nature.  While large language models (LLMs) capture surface-level patterns of humor, they remain limited on tasks requiring deeper reasoning \cite{trott-et-al-2025-turing}. Early computational approaches relied on rule-based templates for jokes and puns \cite{ritchie2001computational, mihalcea2006learning, doogan-etal-2017-idiom}, later replaced by data-driven methods with curated humor corpora \cite{hossain-etal-2019-president, hossain-etal-2020-stimulating} and large-scale datasets \cite{West_Horvitz_2019, horvitz-etal-2024-getting}. Recent benchmarks extend to visually grounded humor, including HumorDB \cite{jain2025humordbaiunderstandgraphical}, YesBut \cite{yesbut-neurips2024}, DeepEval \cite{yang-etal-2024-large}, and Oogiri \cite{Zhong_2024_CVPR}. However, these datasets primarily evaluate outcomes (e.g., selecting the funnier caption) rather than modeling the underlying reasoning processes emphasized in cognitive theory.

\parahead{The New Yorker Cartoon Caption Contest}
The New Yorker Cartoon Caption Contest (NYCC) is a key resource for studying multimodal humor, aligning visual input, linguistic creativity, expert judgment, and crowd preferences. \citet{hessel-etal-2023-androids} introduced a benchmark derived from NYCC spanning caption matching, ranking, and explanation tasks, with later work extending evaluation through harder ranking splits \cite{zhou2025bridgingcreativityunderstandinggap} and large-scale preference analysis \cite{zhang-etal-2024-humor}. Recent approaches incorporate humor theory into modeling. \citet{shang2025homer} propose HOMER, a generation framework grounded in the General Theory of Verbal Humor \cite{attardo1991script}, showing improved caption quality over prompting baselines. IRS is complementary: while HOMER focuses on caption generation with closed models, IRS targets humor understanding via explicit supervision of intermediate reasoning in open-weight models.

\parahead{Multimodal Reasoning and Alignment}
Recent work explores equipping multimodal LLMs with explicit reasoning via structured traces distilled from stronger models, improving performance and reducing hallucination \cite{xu2025llavacotletvisionlanguage, thawakar2025llamavo1rethinkingstepbystepvisual, liu2025xreasonergeneralizablereasoningmodalities, liao2025longperceptualthoughts}. Yet, these approaches generally rely on generic chain-of-thought formats rather than task-specific reasoning structures. Reinforcement learning with grounded rewards offers a complementary alignment strategy. Methods such as Vision-R1 \cite{zhan2025visionr1evolvinghumanfreealignment}, Visual-RFT \cite{liu2025visualrftvisualreinforcementfinetuning}, and Perception-R1 \cite{xiao2025advancingmultimodalreasoningcapabilities} improve visual grounding and reduce spurious correlations through task-specific reward signals, but do not explicitly model the structure of reasoning itself.

\begin{figure*}[!t]
\begin{center}
\includegraphics[width=\linewidth]{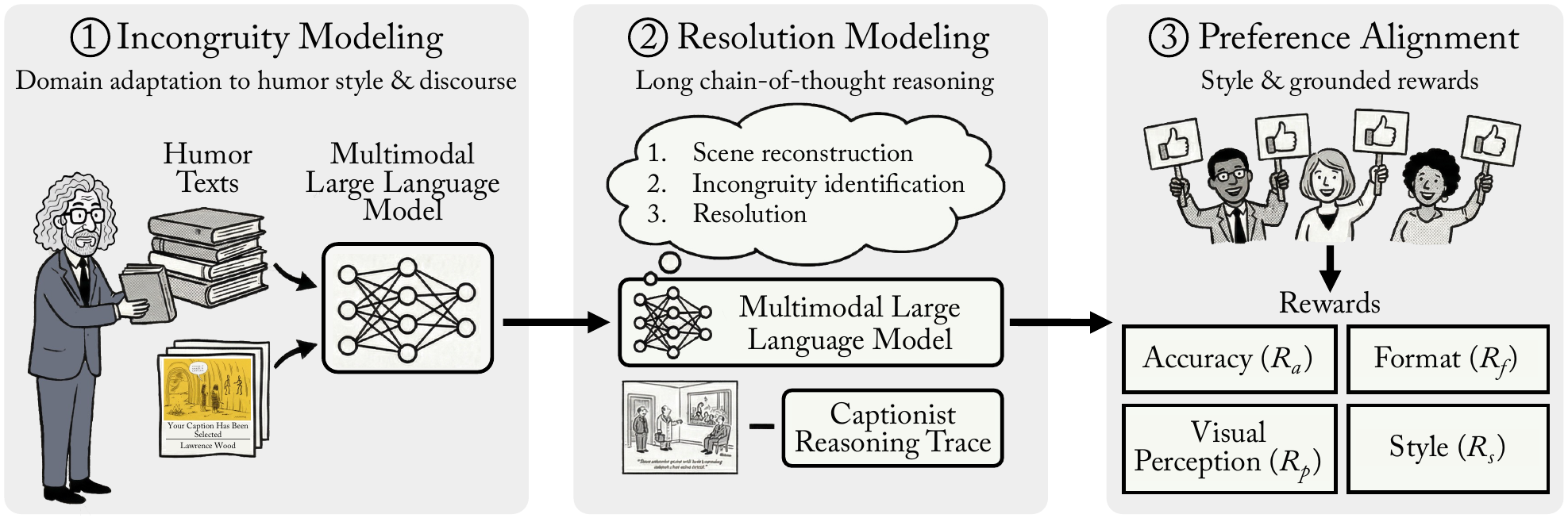} 
\end{center}
\caption{\textbf{Overview of Incongruity-Resolution Supervision (IRS).}
IRS models humor understanding as a structured reasoning process with three components: (1)~\textit{Incongruity Modeling (IM)}, which identifies mismatches in the visual scene; (2)~\textit{Resolution Modeling (RM)}, which constructs coherent reinterpretations of these mismatches; and (3)~\textit{Preference Alignment (PA)}, which evaluates candidate interpretations under human judgments. The training stages, continual pretraining, supervised learning on captionist reasoning traces, and reinforcement learning with grounded rewards, support these components.}
\label{fig:overview}
\end{figure*}

\section{Approach}
Analyses of professional captionist practice \citep{wood2024your,mankoff2002nakedcartoonist} reveal a consistent interpretive process: identifying a salient visual incongruity, constructing a coherent reinterpretation, and evaluating punchlines by how effectively they resolve the tension in a surprising yet meaningful way. IRS models this process through three learnable components: \emph{Incongruity Modeling (IM)}, which identifies mismatches in the visual scene; \emph{Resolution Modeling (RM)}, which constructs coherent reinterpretations of those mismatches; and \emph{Preference Alignment (PA)}, which evaluates candidate interpretations against human judgments of humor (Fig.~\ref{fig:overview}). We implement this via \emph{captionist reasoning traces}, structured sequences that make  intermediate reasoning from visual perception to humorous interpretation explicit. These traces capture how incongruities are identified, interpretations are constructed, and captions are evaluated, enabling direct supervision of reasoning.

\subsection{Incongruity Modeling via Domain-Adaptive Pretraining}
Recognizing humor-relevant incongruities requires background knowledge often absent from standard pretraining corpora: stylistic conventions of cartoon humor, editorial heuristics for what makes a caption work, and culturally grounded expectations about narrative and visual surprise. In the IM stage, we perform \emph{continual pretraining} (CPT) on a curated corpus of captionist discussions, editorial analyses, caption-writing guides, and complementary general-knowledge text. CPT uses a standard causal language modeling objective and does not optimize downstream tasks directly. Its role is preparatory: biasing the model's representations toward humor-relevant concepts, incongruity-resolution structure, lexical economy, narrative framing, and culturally grounded references, so that subsequent supervised training can build on a more appropriate prior. Models incorporating this stage show more stable and coherent reasoning during fine-tuning, suggesting that domain adaptation meaningfully shapes what the model attends to rather than merely what it predicts. 
\textit{Full details of the IM corpus are provided in App.~\ref{sec:cpt-corpora}.}

\begin{figure*}[!t]
\centering
\includegraphics[width=\linewidth]{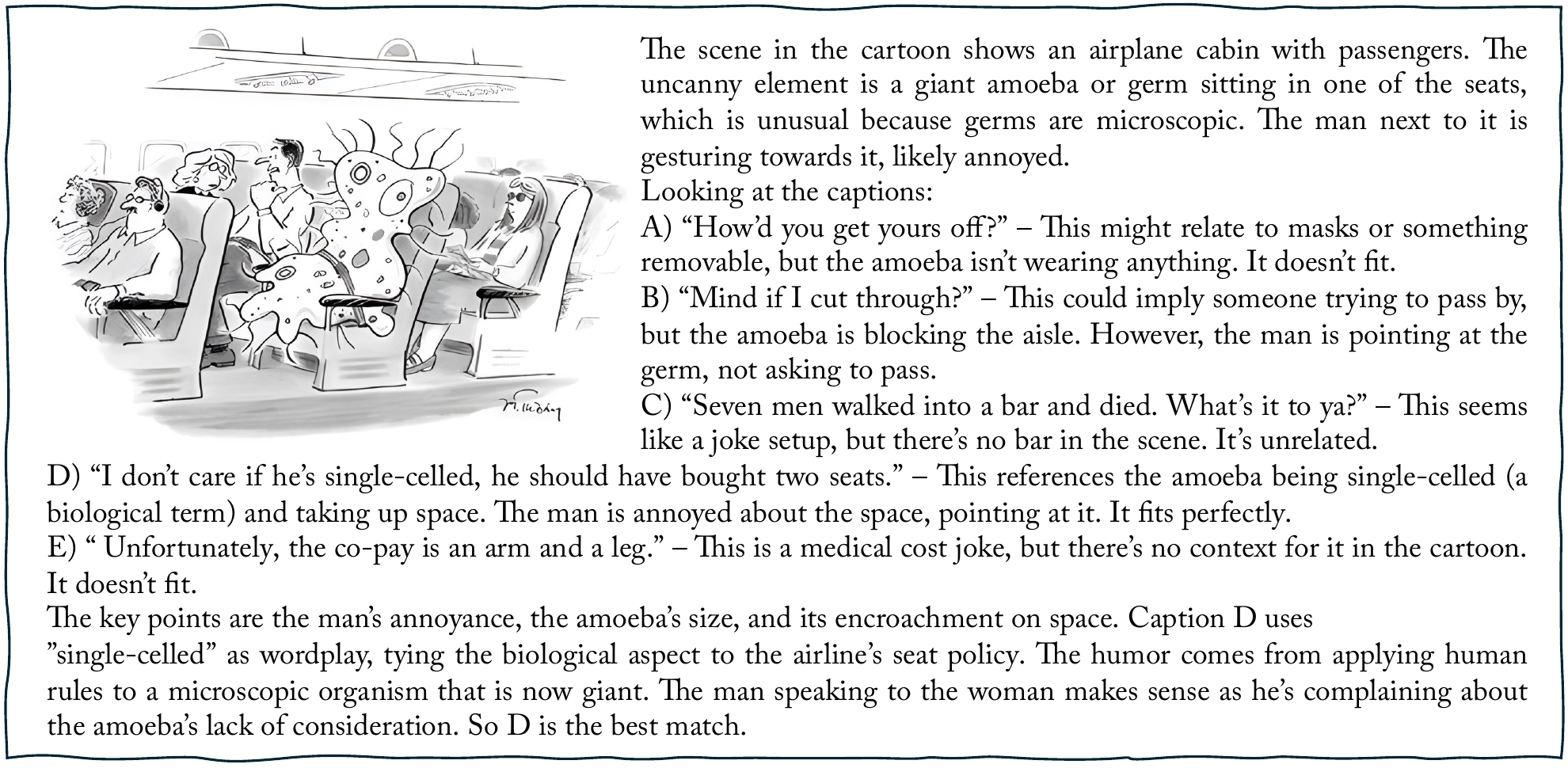} 
\caption{\textbf{Example captionist reasoning trace under IRS (matching).}
Given a cartoon and five candidate captions (A-E), the trace shows structured reasoning: reconstructing the scene, identifying the key incongruity, evaluating alternatives, and selecting the caption best resolving the mismatch. The final choice links the character’s annoyance to the wordplay on ``single-celled'', illustrating both visual grounding and humor-specific reasoning.}
\label{fig:cot_matching_example}
\end{figure*}

\subsection{Resolution Modeling via Captionist Reasoning Traces}
\label{ssec:SFT}
The core of IRS is the supervision of resolution modeling; teaching the model how incongruities are reinterpreted into coherent, humorous readings. We construct a dataset of \emph{captionist reasoning traces} for both caption matching and ranking tasks. Starting from human-annotated cartoon descriptions provided by \citet{hessel-etal-2023-androids}, we generate structured step-by-step analyses using DeepSeek-R1 \citep{deepseekai2025deepseekr1incentivizingreasoningcapability}, following a fixed expert-derived template that covers scene reconstruction, incongruity identification, and narrative resolution. The generated traces are further verified under human supervision to ensure consistency with expert reasoning patterns. To reflect professional captionist discourse, traces are rephrased using GPT-4o \citep{openai2024gpt4o} to emphasize concise, observational, image-grounded commentary, replacing description-based phrasing with direct visual reference while preserving the underlying reasoning structure. In rare cases where annotations do not fully support the correct caption, we minimally adjust the teacher prompt to ensure alignment with the intended reasoning. We assess final trace quality using a structured cross-model audit with DeepSeek-R1 \citep{deepseekai2025deepseekr1incentivizingreasoningcapability}, NVLM-D-72B \citep{dai2024nvlmopenfrontierclassmultimodal}, and Gemini-2.5-Flash \citep{comanici2025gemini25pushingfrontier} as independent external judges. The traces receive strong audit scores, suggesting that RM provides substantive humor-reasoning supervision rather than generic rationales.

Formally, given an input pair $(I, Q)$ consisting of a cartoon image and its associated question, the model is supervised to produce outputs of the form: $\langle \texttt{think} \rangle \textit{ reasoning steps } \langle/\texttt{think}\rangle$
$\langle \texttt{answer} \rangle \textit{ final choice } \langle/\texttt{answer}\rangle$.
This structured supervision encourages models to internalize captionist-style reasoning rather than relying on surface-level patterns. Fig.~\ref{fig:cot_matching_example} shows a representative trace: the model reconstructs the scene (a giant amoeba on an airplane), rules out off-topic captions, and identifies the correct option by recognizing the wordplay on ``single-celled'' as the resolution of the visual incongruity. 
\textit{Prompt templates for reasoning trace generation and the self-audit analysis are respectively provided in App.~\ref{sec:prompts} and App.~\ref{sec:audit}, while additional examples are given in App.~\ref{sec:examples}.}

\begin{figure*}[!t]
\centering
\includegraphics[width=\linewidth]{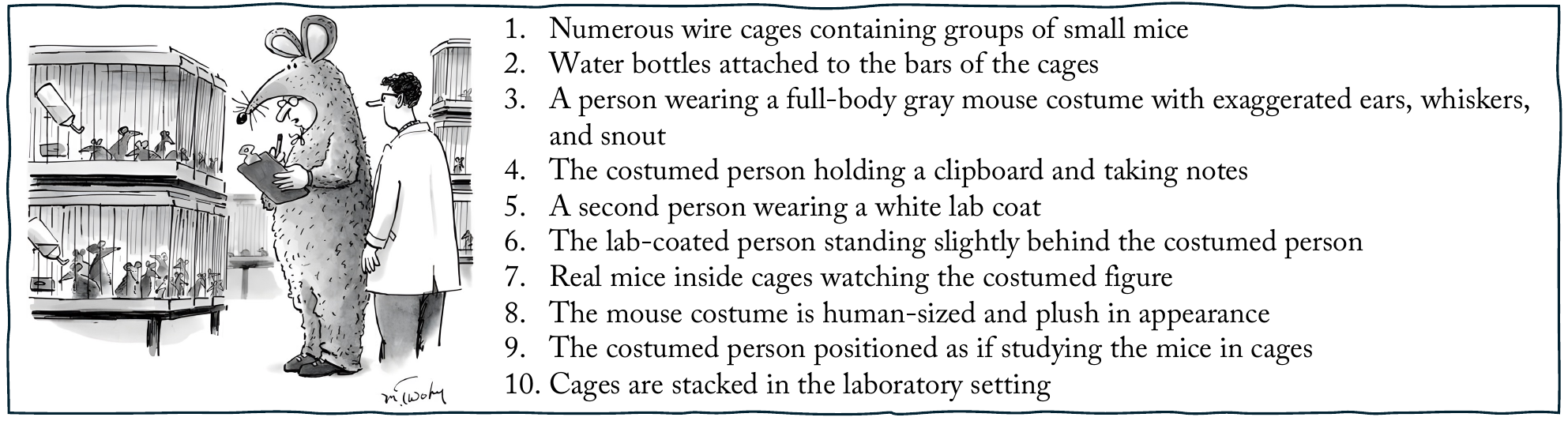}
\caption{\textbf{Curated visual references for perception alignment in IRS.} For each cartoon, we collect concise descriptions of entities, scene context, and key incongruities. These references serve as anchors 
for evaluating visual grounding and computing the perception reward.
}
\label{fig:visual_references}
\end{figure*}

\begin{figure*}[!t]
  \centering
  \includegraphics[width=\linewidth, keepaspectratio]{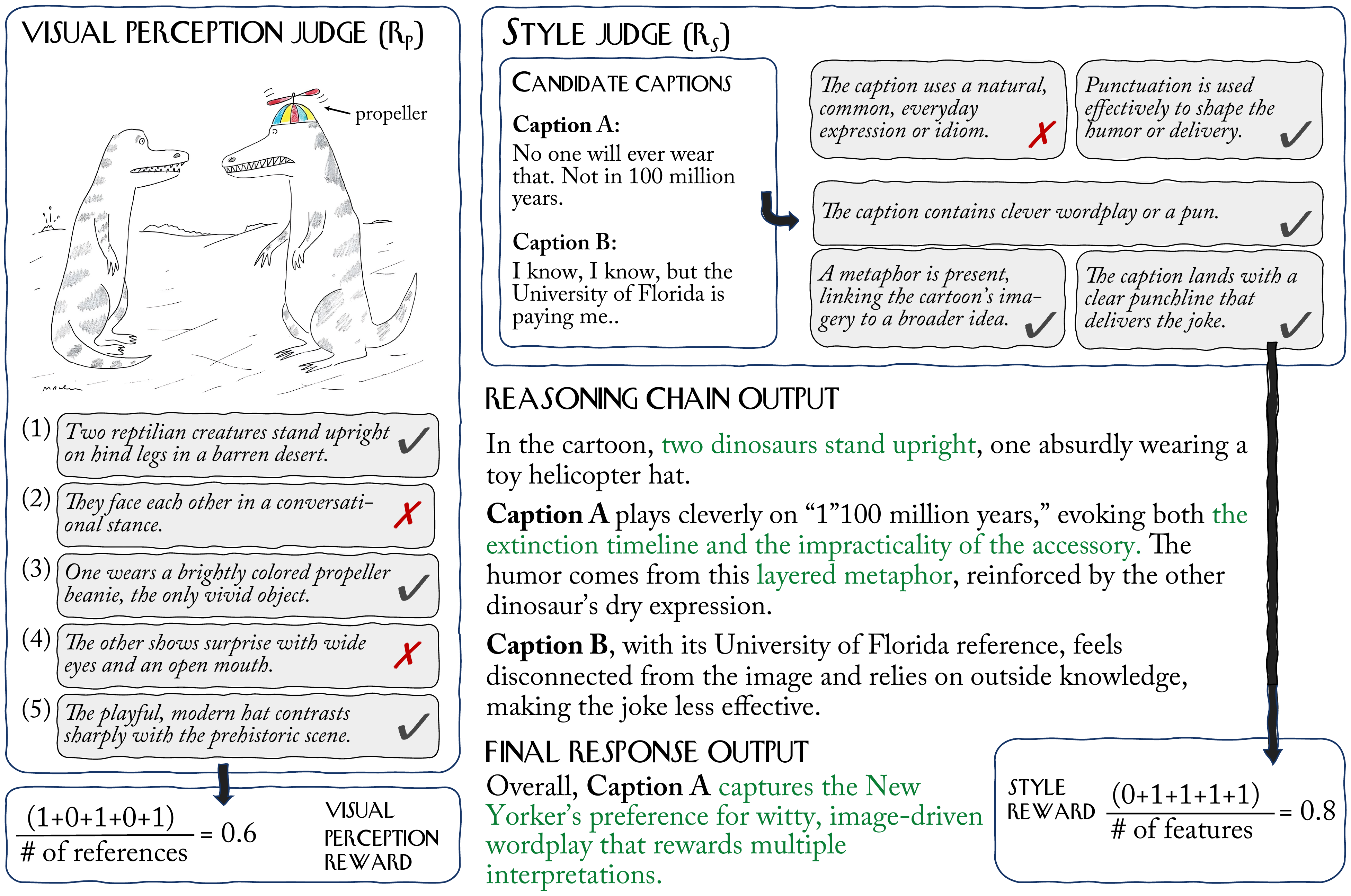} 
  \caption{\textbf{Preference alignment in IRS via judge-based rewards.} Given a cartoon and candidate captions, two judges evaluate reasoning quality: a \emph{visual perception} judge checks grounding in salient elements and incongruities, while a \emph{style} judge assesses linguistic quality. Their binary outputs are aggregated into perception and style rewards ($R_p$, $R_s$), guiding learning toward visually grounded, captionist-consistent reasoning.}
  \label{fig:judge_rewards}
\end{figure*}

\subsection{Preference Alignment via Humor-Aware Rewards}
\label{ssec:RL}
Resolution Modeling provides reasoning structure, but does not ensure visual grounding or stylistic consistency. We address this through reinforcement learning using humor-specific rewards, optimized via GRPO  \citep{deepseekai2025deepseekr1incentivizingreasoningcapability}, which directly optimizes the reasoning process without a value network, using the following objective:
\begin{eqnarray}
\begin{split}
\mathcal{J}(\theta)
&= \mathbb{E}_{(I, q)\sim \mathcal{D},\; \{o_i\}_{i=1}^{G}\sim \pi_{\theta_{\mathrm{old}}}(O \mid I, q)} \notag\\
&
   \frac{1}{G}\sum_{i=1}^{G} \frac{1}{|o_i|}\sum_{t=1}^{|o_i|}
   \min\!\Biggl(
      \frac{\pi_{\theta}(o_{i,t}\mid q,\,o_{i,<t})}
           {\pi_{\theta_{\mathrm{old}}}(o_{i,t}\mid q,\,o_{i,<t})}\,\hat A_{i,t},\Biggr.
\\ & \Biggl.
      \text{clip}\!\Biggl(
         \frac{\pi_{\theta}(o_{i,t}\mid q,\,o_{i,<t})}
              {\pi_{\theta_{\mathrm{old}}}(o_{i,t}\mid q,\,o_{i,<t})},
         1-\epsilon,\,1+\epsilon
      \Biggr)\hat A_{i,t}
   \Biggr) \notag \\ & 
 - \beta\,\text{KL}\Biggl[\pi_\theta \,\Vert\, \pi_{\mathrm{ref}}\Biggr],
\end{split}
\end{eqnarray}

where $\epsilon$ is the clipping hyperparameter, $\beta$ is the KL penalty, and $\pi_{\mathrm{ref}}$ is the reference model. Advantages are computed by normalizing rewards across sampled responses:
\[
\hat A_i = \frac{r_i - \mathrm{mean}(\{r_i\}_{i=1}^{G})}{\mathrm{std}(\{r_i\}_{i=1}^{G})}.
\]
Our composite reward combines standard correctness signals with humor-specific components. The first two use common multimodal RLHF practice:
\begin{enumerate}[label=(\roman*),leftmargin=*]
\item \textbf{Accuracy} ($R_a$): verifies whether the final caption  is correct (gold caption for matching; crowd-preferred caption for ranking).
\item \textbf{Format} ($R_f$): enforces adherence to the structured reasoning format, ensuring outputs remain parseable and interpretable.
\end{enumerate}
These signals provide necessary scaffolding but are insufficient for humor understanding. Our key additions are two humor-specific rewards:  
\begin{enumerate}[leftmargin=*,label=(\roman*)]
\setcounter{enumi}{2}
\item \textbf{Visual Perception} ($R_p$): rewards reasoning grounded in salient visual elements and incongruities. As illustrated in Fig.~\ref{fig:visual_references}, we curate up to ten reference descriptions per cartoon, extending the annotations of \citet{hessel-etal-2023-androids}, and use Qwen2.5-7B-Instruct~\citep{qwen25technicalreport_pre} as an LLM judge to assess visual grounding.

\item \textbf{Style} ($R_s$): evaluates linguistic quality using an LLM-as-judge based on captionist guidelines~\citep{wood2024your}, assessing everyday phrasing, punctuation, wordplay, metaphor, and punchline placement via an LLM-as-judge.  A binary score is aggregated into the style reward using the same Qwen2.5-7B-Instruct judge.
\end{enumerate}

Fig.~\ref{fig:judge_rewards} illustrates how these judges operate in practice.
The four signals are combined as:
\begin{equation}
R
= \lambda_a R_a
+ \lambda_f R_f
+ \mathbb{I}[R_a = 1]\left( \lambda_p R_p + \lambda_s R_s \right),
\end{equation}
conditioning perception and style rewards on correctness to ensure they reinforce rather than distract from accurate caption selection. Following prior work~\citep{liu2025xreasonergeneralizablereasoningmodalities,xiao2025advancingmultimodalreasoningcapabilities}, we remove the KL penalty, which improves reasoning coherence. When either $R_p$ or $R_s$ exceeds 80\% of its maximum, over 70\% of cases are correct, indicating that the LLM-based judge provides meaningful guidance rather than spurious optimization signals. Importantly, the judge operates solely on reasoning traces, without access to ground-truth captions, ensuring reward signals reflect reasoning quality rather than answer matching. 
\textit{Reward-judge prompts are provided in App.~\ref{sec:prompts}.}
\section{Experimental Setup}
\label{sec:experimental_setup}
We evaluate IRS in two settings: in-domain NYCC caption selection and zero-shot cross-dataset generalization. The in-domain evaluation covers four benchmark settings spanning caption matching and humor ranking, while the cross-dataset evaluation tests whether the learned reasoning patterns transfer beyond NYCC. \textit{Training configurations and optimization details are provided in App.~\ref{sec:training}.}

\paragraph{Datasets and Tasks.}
Our in-domain evaluation uses four NYCC-based settings. From \citet{hessel-etal-2023-androids}, we evaluate: (i) \textbf{Matching}, where a model selects the gold caption from five candidates given a cartoon image; and (ii) \textbf{Ranking}, where a model selects the crowd-preferred caption from two options. From \citet{zhou2025bridgingcreativityunderstandinggap}, we evaluate: (iii) \textbf{10-vs-1000}, which contrasts a top-10 caption with one ranked 1000--1009; and (iv) \textbf{30-vs-300}, which contrasts a caption ranked 30--39 with one ranked 300--309.
These tasks probe complementary aspects of humor reasoning. Matching tests whether models can identify the visual interpretation that best matches the cartoon, while ranking emphasizes preference alignment between plausible captions. The 30-vs-300 split is especially challenging because both captions are crowd-ranked and stylistically competent, requiring fine-grained preference discrimination.

For cross-dataset generalization, we additionally evaluate IRS zero-shot on two external humor-related benchmarks: \textbf{YesBut}~\citep{yesbut-neurips2024}, a two-panel visual humor dataset centered on contrastive reasoning, where we use the classification-based \emph{Philosophy} and \emph{Title} subtasks; and \textbf{DeepEval}~\citep{yang-etal-2024-large}, a broader multimodal benchmark with a small humor subset, where we evaluate \emph{DeepSemantics}, \emph{Description}, and \emph{Title}. These datasets differ from NYCC in visual structure, task format, and annotation scheme, allowing us to test whether IRS learns transferable reasoning patterns rather than NYCC-specific heuristics.
To prevent data leakage, all evaluation cartoons are excluded from the continual pretraining corpus, and we verify negligible n-gram overlap between the CPT corpus and evaluation caption text.

\paragraph{Models.}
We compare IRS against a diverse set of baselines spanning text-only, closed-source, and open-weight systems, allowing us to assess gains attributable to explicit reasoning supervision independently of scale and data access advantages.

\begin{itemize}[leftmargin=*]

\item \textbf{Text-only reasoning.}
DeepSeek-R1~\citep{deepseekai2025deepseekr1incentivizingreasoningcapability} is evaluated on ground-truth textual annotations of cartoons rather than raw images. Since it receives curated scene descriptions that multimodal models must infer from pixels, its performance represents an approximate upper bound on reasoning given perfect visual perception rather than a true multimodal comparison.

\item \textbf{Closed multimodal models.} We include o3 and o4-mini \citep{openai_o3_system_card_2025} as representative high-performing proprietary systems. Their performance provides a useful reference point, although such models may benefit from data and scale advantages unavailable to open-weight models.

\item \textbf{Open multimodal reasoning models.} We benchmark against recent open-weight models trained for multimodal reasoning: GLM-4.1V-9B-Thinking \citep{vteam2025glm45vglm41vthinkingversatilemultimodal}, Kimi-VL-A3B-Thinking-2506 \citep{kimiteam2025kimivltechnicalreport}, LlamaV-o1 \citep{thawakar2025llamavo1rethinkingstepbystepvisual}, and Qwen2.5-VL at 7B, 32B, and 72B scales \citep{qwen2025qwen25technicalreport}. These models apply general-purpose reasoning strategies without humor-specific supervision.

\item \textbf{IRS (ours).} We apply IRS to Qwen2.5-VL backbones at 7B, 32B, and 72B scales, implementing all three training stages: domain-adaptive pretraining, supervised fine-tuning on captionist reasoning traces, and preference-based alignment with perceptual and stylistic rewards. Comparing IRS against the corresponding Qwen2.5-VL base models isolates the contribution of IRS-specific supervision from backbone capacity.

\item \textbf{Human baselines.}
We report expert and non-expert human performance on a subset of tasks. The expert captionist, a professional with sustained experience in NYCC, serves as a qualitative reference for what human-level humor reasoning looks like, while a user study with 21 participants provides a measure of alignment with crowd preferences.

\end{itemize}
\textit{Evaluation prompts are provided in App.~\ref{sec:prompts}.}
\section{Experimental Results}
\label{sec:experiment_results}

\begin{table}[!t]
\centering
\caption{\textbf{Performance across models.} Accuracy (\%) on caption matching and ranking tasks. Bold values indicate the best performance in each column, and underlined values indicate the second-best performance (excluding human baselines).}
\resizebox{\linewidth}{!}{%
\begin{tabular}{@{}l@{}c@{$\,$}c@{$\,\,$}c@{$\,$}c@{}}
\toprule
\multirow{2}{*}{\textbf{Model}} & \multicolumn{2}{c}{\textbf{(Hessel et al., 2023)}} & \multicolumn{2}{c}{\textbf{(Zhou et al., 2025)}}\\
& \textbf{Matching} & \textbf{Ranking} & \textbf{10-vs-1000} & \textbf{30-vs-300}\\%
\midrule
Human (Expert) & 100.00 & 100.00 & 60.00 & 40.00\\[4.5pt]
\textcolor{black}{Human (Non-Expert)} & \textcolor{black}{53.03} & \textcolor{black}{65.61} & \textcolor{black}{54.70} & \textcolor{black}{52.27}\\[4.5pt]
DeepSeek-R1 & 74.00  & 64.67 & 56.86 & 47.14\\[4.5pt]

o4-mini & \underline{75.08} &  62.59 & 60.17 & 51.42\\[1pt]
o3 & \textbf{83.33} & 62.85 &  \textbf{69.05} & \textbf{54.57} \\[4.5pt]
GLM-4.1V-9B-Think & 59.40 & 55.60 & 52.28 & 49.14\\[1pt]
Kimi-VL-A3B-Think & 54.00 & 57.45 & 52.30 & 52.01 \\[1pt]
LlamaV-o1 & 43.67 & 50.90 & 48.57 & 49.42 \\[1pt]
Qwen2.5-VL-7B-Inst & 42.67 & 55.06 & 50.57 & 47.99\\[1pt]
Qwen2.5-VL-32B-Inst & 46.67 & 49.87 & 52.00 & 44.57\\[1pt]
Qwen2.5-VL-72B-Inst & 56.00 & 55.58 & 53.71 & 50.29\\[4.5pt]
IRS-7B (Our model)& 59.67 & 64.42 & 56.29 & \underline{53.14}\\[1pt]
\textcolor{black}{IRS-32B (Our model)} & \textcolor{black}{62.67} & \underline{\textcolor{black}{68.05}} & \textcolor{black}{\underline{62.86}} & \textcolor{black}{\underline{53.14}}\\
[1pt]
\textcolor{black}{IRS-72B (Our model)} & \textcolor{black}{69.33} & \textcolor{black}{\textbf{76.10}} & \textcolor{black}{62.57} & \textcolor{black}{50.86}\\
[1pt]
\bottomrule
\end{tabular}}
\label{table:comparison}
\end{table}

\subsection{Comparison with Baselines}
Table~\ref{table:comparison} reports accuracy across all models.

\paragraph{Human performance reveals task-specific difficulty.} The expert captionist achieves perfect scores on \emph{matching} and \emph{ranking}, tasks built from NYCC finalist captions — but shows lower agreement with crowd preferences on the 10-vs-1000 and 30-vs-300 settings (60\% and 40\%, respectively). This is consistent with the known divergence between editorial judgment and popular vote: expert captionists optimize for originality and craft, while crowd preferences reflect broader accessibility. Non-expert human performance is substantially lower across all tasks, confirming that NYCC-style humor reasoning requires sustained domain familiarity rather than general common sense.

\paragraph{Closed models benefit from data advantages that open models lack.} Among baselines, o3 and o4-mini remain highly competitive, with o3 achieving the highest matching accuracy overall. Their strong performance may reflect both architectural sophistication and possible data advantages, including broader familiarity with New Yorker-style editorial content unavailable to open-weight models.\footnote{OpenAI had announced a partnership with Condé Nast to display content from brands including \textit{The New Yorker} within their products~\citep{openai_conde_nast_2024}. This does not imply direct training on NYCC captions, but it may provide broader exposure to New Yorker-style editorial language and humor.} DeepSeek-R1, despite operating on curated textual annotations rather than images, performs competitively on matching and ranking, confirming that high-quality perceptual grounding remains a bottleneck for open multimodal models. Among open-weight baselines, scaling alone provides limited benefit: Qwen2.5-VL improves only modestly from 7B to 72B, and even at 72B falls well short of closed-model performance without task-specific reasoning supervision.

\paragraph{IRS consistently improves performance across scales.} Models trained with IRS outperform their base counterparts at every scale, with improvements growing as backbone capacity increases. At 7B, IRS narrows the gap to closed models substantially, demonstrating that explicit reasoning supervision transfers even at smaller scales. At 32B, IRS produces the strongest open-weight model across most tasks, with particularly large gains on ranking — the task most sensitive to preference alignment between competing interpretations. At 72B, IRS achieves the highest ranking accuracy overall (76.10\%), surpassing all baselines including o3, and approaches expert-level performance on that task. Performance on the 30-vs-300 setting is comparatively less stable, as semantically similar candidates make fine-grained preference distinctions inherently ambiguous.
\textit{Qualitative results demonstrate that IRS-trained models can generate plausible, captionist-style captions despite not being explicitly trained for this task (App.~\ref{sec:captioning})}. \textit{We also audit inference-time reasoning quality, showing that IRS improves humor-specific rationale dimensions beyond answer accuracy (App.~\ref{sec:audit}).}

\subsection{Comparison with Inference-Time CoT Prompting}
A natural question is whether the gains from IRS could be recovered simply by eliciting step-by-step reasoning at inference time, without any of the proposed supervision. To test this, we evaluate controlled zero-shot and few-shot CoT baselines on the same Qwen2.5-VL-7B and 32B backbones, using identical evaluation sets, decoding protocol, and structured output format. As Table~\ref{table:test_prompting_comparison} shows, IRS achieves the best performance on all four tasks at both backbone scales, improving over the strongest CoT configuration per task by 2.00--13.34 points at 7B and 4.28--12.00 points at 32B. Notably, few-shot CoT does not consistently outperform zero-shot CoT, indicating that additional in-context reasoning demonstrations alone provide no reliable benefit on these tasks.

These controls indicate that inference-time reasoning elicitation alone cannot account for the gains. Rather, the improvement comes from explicitly supervising \emph{what} the model reasons about—the visual incongruity, its plausible resolution, and the comparative quality of humorous interpretations—and aligning that reasoning with humor-specific grounding and stylistic signals.

\begin{table}[!t]
\centering
\caption{\textbf{Comparison of different prompting strategies and IRS.} Accuracy (\%) on caption matching and ranking tasks. Bold values indicate the best performance in each column.}
\resizebox{\linewidth}{!}{%
\begin{tabular}{@{}lc@{$\,\,$}c@{$\,\,\,$}c@{$\,\,$}c@{}}
\toprule
\multirow{2}{*}{\textbf{Method}} & \multicolumn{2}{c}{\textbf{(Hessel et al., 2023)}} & \multicolumn{2}{c}{\textbf{(Zhou et al., 2025)}}\\
& \textbf{Matching} & \textbf{Ranking} & \textbf{10-vs-1000} & \textbf{30-vs-300}\\%
\midrule
\multicolumn{5}{l}{\textit{Qwen2.5-VL-7B}} \\
Direct Prompting & 42.67 & 55.06 & 50.57 & 47.99\\[1pt]
Zero-shot CoT  & 46.33 & 55.32 & 48.29 & 50.86\\[1pt]
Few-shot CoT  & 44.67 & 51.69 & 47.43 & 51.14\\[1pt]
IRS (Ours) & \textbf{59.67} & \textbf{64.42} & \textbf{56.29} & \textbf{53.14} \\[4.5pt]
\midrule
\multicolumn{5}{l}{\textit{Qwen2.5-VL-32B}} \\
Direct Prompting & 46.67 & 49.87 & 52.00 & 44.57\\[1pt]
Zero-shot CoT  & 49.33 & 57.66 & 49.71 & 48.86\\[1pt]
Few-shot CoT  & 50.67 & 54.03 & 55.43 & 48.00\\[1pt]
IRS (Ours) & \textbf{62.67} & \textbf{68.05} & \textbf{62.86} & \textbf{53.14}\\[4.5pt]
\bottomrule\vspace{-1cm}
\end{tabular}}
\label{table:test_prompting_comparison}
\end{table}

\subsection{Ablation on Sources of Supervision}
\label{ssec:ablation}
Table~\ref{tab:training_ablation} isolates the contribution of each IRS component on the 7B backbone. Three findings stand out. 
RM is the main source of improvement. Adding RM to the base model leads to consistent gains across all tasks, highlighting the importance of explicitly modeling how incongruities are interpreted into coherent humorous readings. IM, by contrast, is only helpful when combined with RM. On its own, it does not improve accuracy and can even degrade performance, suggesting that humor-specific priors are most effective when paired with a structured reasoning process. When combined with RM, however, it provides additional gains, particularly on the more challenging ranking tasks \citep{zhou2025bridgingcreativityunderstandinggap}. In contrast, IM+PA underperforms Base+PA, indicating that domain adaptation without resolution supervision can introduce instability. PA provides complementary benefits, especially on ranking tasks that depend on perceptual grounding and stylistic quality. The full IRS pipeline (IM+RM+PA) achieves the strongest overall performance, indicating that each component contributes a distinct and necessary aspect of humor reasoning.

\begin{table}[!t]
\centering
\caption{\textcolor{black}{\textbf{Ablation on IRS components.} Each component contributes complementary gains; the full IRS pipeline yields the strongest results.}}
\color{black}{
\resizebox{\linewidth}{!}{%
\begin{tabular}{@{}l@{}c@{$\,$}c@{$\,\,$}c@{$\,$}c@{}}
\toprule
\multirow{2}{*}{\textbf{Approach}} & \multicolumn{2}{c}{\textbf{(Hessel et al., 2023)}} & \multicolumn{2}{c}{\textbf{(Zhou et al., 2025)}}\\
& \textbf{Matching} & \textbf{Ranking} & \textbf{10-vs-1000} & \textbf{30-vs-300} \\
\midrule
Base Model & 42.67 & 55.06 & 50.57 & 47.99\\[1pt]
+ IM  & 41.00 & 51.69 & 50.29 & 51.43 \\[1pt]
+ RM  & 47.00 & 56.88 & 49.71 & 50.86 \\[1pt]
+ IM + RM  & 49.00 & 56.88 & 54.57 & 49.14 \\[1pt]
+ PA  & 56.67 & 58.96 & 54.86 & 49.14 \\[1pt]
+ RM + PA & 57.33 & 58.18 & 56.29 & 44.86 \\[1pt] 
+ IM + PA  & 46.00 & 51.95 & 49.43 & 50.29 \\[1pt]
+ IM + RM + PA (IRS)  & 59.67 & 64.42 & 56.29 & 53.14 \\[1pt]
\bottomrule
\end{tabular}
}
}
\label{tab:training_ablation}
\end{table}

\subsection{Ablation on Reward Functions}
Table~\ref{tab:reward_ablation} isolates the contribution of reward signals within Preference Alignment (PA). Accuracy and format rewards ($R_a+R_f$) already provide a strong baseline, improving matching performance over the IM+RM model. Adding humor-aware rewards yields more targeted gains rather than uniform improvements across all tasks. The visual perception reward ($R_p$) improves ranking and 10-vs-1000 performance, suggesting that explicit grounding is most useful when the model must compare plausible captions. The style reward ($R_s$) provides smaller but consistent benefit on ranking, aligning with its role in encouraging captionist-consistent expression. These results indicate that $R_p$ and $R_s$ should be interpreted primarily as reasoning-shaping rewards: they are designed to improve visual grounding and stylistic quality, rather than to yield uniform accuracy improvements across all tasks. While $R_a$ directly supervises caption selection, $R_p$ encourages explanations grounded in visual incongruities and $R_s$ encourages stylistically appropriate humor analysis. Thus, the humor-aware rewards complement correctness-based RL by guiding the structure and quality of the rationale.

\begin{table}[!t]
\centering
\caption{\textbf{Ablation on reward functions.} Humor-aware rewards improve ranking and generalization.}
\small
\resizebox{\linewidth}{!}{%
\begin{tabular}{@{}l@{$\;\,$}c@{$\;\,$}c@{$\;\;$}c@{$\;\,$}c@{}}
\toprule
\multirow{2}{*}{\textbf{Approach}} & \multicolumn{2}{c}{\textbf{(Hessel et al., 2023)}} & \multicolumn{2}{c}{\textbf{(Zhou et al., 2025)}}\\
& \textbf{Matching} & \textbf{Ranking} & \textbf{10-vs-1000} & \textbf{30-vs-300} \\
\midrule
Base Model+IM+RM & 49.00 & 56.88 & 54.57 & 49.14\\[1pt]
+ PA (w/ $R_a$+$R_f$) & 60.67 & 57.99 & 54.57 & 50.86\\[1pt]
+ PA (w/ $R_a$+$R_f$+$R_p$) & 58.00 & 60.78 & 56.57 & 49.43\\[1pt]
+ PA (w/ $R_a$+$R_f$+$R_s$) & 60.00 & 58.44 & 54.57 & 49.43\\[1pt]
\bottomrule
\end{tabular}
}
\label{tab:reward_ablation}
\end{table}

\subsection{Cross-Dataset Generalization}
\label{sec:cross-dataset}
We evaluate zero-shot transfer beyond NYCC using the external benchmarks described in Sec.~\ref{sec:experimental_setup}. Despite differences in visual structure, task format, and annotation scheme, IRS consistently improves over its base model. On YesBut \citep{yesbut-neurips2024}, it gains +31.7 and +34.2 points on \emph{Philosophy} and \emph{Title}, respectively (Table~\ref{tab:yesbut}), indicating transfer to contrastive humor reasoning. On DeepEval \citep{yang-etal-2024-large}, it improves across all evaluated tasks (Table~\ref{tab:deepeval}), especially semantic alignment and descriptive accuracy, despite minimal stylistic overlap with NYCC.

\begin{table}[!t]
\centering
\caption{\color{black}{\textbf{Zero-shot Generalization to YesBut \citep{yesbut-neurips2024}}  
Accuracy (\%) on the Philosophy and Title subtasks.  
Our models are trained only on NYCC data yet show strong transfer, particularly for the 7B backbone.}}
\small
\color{black}{
\begin{tabular}{lcc}
\toprule
\textbf{Model} & \textbf{Philosophy} & \textbf{Title} \\
\midrule
Qwen2.5-VL-7B-Inst (Base) & 43.19 & 29.11 \\
IRS-7B (Ours) & \textbf{74.90} & \textbf{63.32}\\ 
\bottomrule
\end{tabular}}
\label{tab:yesbut}
\end{table}

\begin{table}[!t]
\centering
\caption{\color{black}{\textbf{Zero-shot Generalization to DeepEval \citep{yang-etal-2024-large}}  
Performance on the \emph{humorous} subset (29 images, 2.9\% of the benchmark).  
We report accuracy on three tasks: DeepSemantics, Description, and Title.}}
\color{black}{
\resizebox{\linewidth}{!}{%
\begin{tabular}{lccc}
\toprule
\textbf{Model} & \textbf{DeepSem} & \textbf{Descr} & \textbf{Title} \\
\midrule
Qwen2.5-VL-7B-Inst (Base) & 10.34 & 24.13 & 34.48 \\
IRS-7B (Ours) & \textbf{54.43} & \textbf{100.00} & \textbf{63.18} \\
\bottomrule
\end{tabular}}}
\label{tab:deepeval}
\end{table}

\section{Conclusion}
We introduced Incongruity-Resolution Supervision (IRS), a framework that models humor understanding as a structured reasoning process grounded in incongruity-resolution theory and expert captionist practice. Rather than supervising caption selection alone, IRS supervises the intermediate steps that make a caption funny--identifying visual mismatches, constructing coherent reinterpretations, and evaluating candidate interpretations against human judgments--through domain-adaptive pretraining, captionist reasoning traces, and preference alignment with perceptual and stylistic rewards. Across 7B, 32B, and 72B models, IRS consistently outperforms strong baselines on matching and ranking tasks, with the largest model approaching expert-level performance and generalizing to out-of-domain benchmarks. These results suggest that explicitly supervising reasoning structure, rather than relying on scale alone, is key for complex, subjective tasks such as humor.

\section*{Limitations}
\label{sec:limitations}
While our framework improves multimodal humor understanding, several limitations remain that highlight open challenges in reasoning-based alignment.

\parahead{Visual perception errors}
Despite reinforcement learning with perceptual rewards, the model can misidentify salient visual entities, which propagates through the reasoning process and leads to incorrect caption selection. As shown in Fig.~\ref{fig:lim-perception}, the model confuses a Viking warrior with the Grim Reaper (scythe) due to superficial visual similarity (e.g., weapon shape), causing a complete mismatch between perceived scene and intended incongruity. Scaling the backbone to 32B substantially reduces such errors, indicating that perceptual grounding benefits from increased representational capacity. However, this improvement is not uniform: stylized cartoons require abstraction over sparse and exaggerated visual cues, which remain underrepresented in standard vision pretraining corpora. This suggests that reasoning supervision alone cannot compensate for systematic perception errors, and that advances in vision encoders or domain-specific visual pretraining are necessary to fully address this limitation.

\parahead{Shallow cultural grounding} The model sometimes produces fluent but brittle analyses that miss culturally embedded references. 
As shown in Fig.~\ref{fig:lim-culture}, an RM-only model latches onto surface wordplay (\textit{``plane''}) and ignores the cartoon’s Superman motif-a popular-culture cue. 
After PA, the model gestures toward the pop-culture context but over-attributes visual evidence (e.g., inventing a \textit{``flying-pose''} link) and still fails to articulate the catchphrase-based joke. 
This illustrates the challenge of grounding cultural knowledge without hallucination.

\parahead{Model size and scaling} Our experiments use Qwen2.5-VL backbones with 7B, 32B, and 72B parameters. While this range provides a fair and reproducible open baseline, it generally underperforms large proprietary systems such as o3 and GPT-4o on matching and harder ranking tasks. Our focus in this work is not raw scale, but the development of humor-specific priors, reasoning traces, and reward functions. \textcolor{black}{Scaling IRS to the Qwen2.5-VL-72B backbone further improves performance, particularly on the standard ranking task, as presented in the Table~\ref{table:comparison}.}

\parahead{Cultural and linguistic specificity} Our dataset is centered on the New Yorker Cartoon Caption Contest, which reflects predominantly Anglophone and U.S.-centric humor. As a result, the learned reasoning patterns may not transfer to other cultural contexts, where humor relies on different conventions, references, and linguistic structures. Extending this framework to multilingual and culturally diverse datasets is necessary to evaluate the generality of humor-aware reasoning.

\parahead{Subjectivity in evaluation} Humor is inherently subjective, and even within NYCC, crowd judgments and editorial selections represent only a subset of possible interpretations. Our reward functions approximate these preferences through perceptual and stylistic criteria, but cannot fully capture the diversity of human responses. This introduces an unavoidable gap between automated evaluation and human judgment, motivating future work on more robust and diverse human evaluation protocols.

\parahead{Data access and licensing} Parts of our IM corpus are derived from copyrighted sources (e.g., podcast transcripts, published books). Due to licensing restrictions, we cannot release this material directly, though we will provide evaluation splits and full preprocessing pipelines. This ensures reproducibility while respecting intellectual property.

Despite these limitations, our work represents a step toward aligning multimodal reasoning models with one of the most elusive facets of human intelligence: humor. Addressing cultural, subjective, and computational challenges offers fertile ground for future research.

\section*{Ethical considerations}
Modeling humor raises additional ethical concerns. Humor often relies on cultural stereotypes, sensitive topics, or implicit social norms, and models trained on such data may reproduce or amplify biases. Furthermore, systems capable of generating humor-like captions could be misused to produce offensive or inappropriate content. While our work focuses on controlled evaluation settings, deploying such models in real-world applications requires careful consideration of content moderation, cultural sensitivity, and user context.

\section*{Acknowledgements}
We thank all reviewers for their valuable comments and suggestions, which helped improve the final version of this work. This work was supported in part by the KUIS AI Center Research Awards to Doga Kukul. Erkut Erdem acknowledges funding from the TUBITAK 2247-A Program under Award 123C542. The authors gratefully acknowledge that the numerical computations reported in this work were performed in part using the TRUBA resources of the TUBITAK ULAKBIM High Performance and Grid Computing Center. The authors also thank Lambda AI for providing compute credits, and Google for the Google Cloud Platform credits awarded through its Gemini Academic Program.

\bibliography{emnlp2026_conference}
\newpage
\appendix

\clearpage
\setcounter{figure}{0}
\setcounter{table}{0}
\renewcommand{\thefigure}{\thesection.\arabic{figure}}
\renewcommand{\thetable}{\thesection.\arabic{table}}
\counterwithin{equation}{section}

\section*{Appendix}
This appendix provides extra material to support transparency, reproducibility, and a deeper understanding of our approach. It expands on key components of the main paper, including dataset construction, training procedures, prompt design, statistical, qualitative, and quantitative analyses, as well as additional experiments that further validate our findings.

The content is organized as follows:

\begin{itemize}[leftmargin=*]

\item \textbf{Section~\ref{sec:all-data-stats} -- Dataset and Benchmark Statistics:}
A consolidated summary of the data used at every training stage and
evaluation split, reporting the number of examples, task composition, and
unique cartoons for the IM, RM, and PA training sets, the in-domain NYCC
matching and ranking benchmarks, and the zero-shot transfer benchmarks.

\item \textbf{Section~\ref{sec:cpt-corpora} -- Incongruity Modeling Corpora:}  
A detailed description of the Incongruity Modeling (IM) dataset, including its composition, sources, temporal coverage, and licensing considerations. This section provides insight into how domain-relevant knowledge is incorporated to support incongruity modeling.

\item \textbf{Section~\ref{sec:sft-corpora} -- Resolution Modeling Corpora:}  
A detailed description of the Resolution Modeling (RM) dataset, including its composition, sources, human-in-the-loop pipeline, cross-family trace rewriting, quality-filtering, and annotator qualifications. This section provides insight into how captionist reasoning traces are generated to support resolution modeling.

\item \textbf{Section~\ref{sec:reward-weights} -- Reward Composition and Weight Allocation:}
An analysis of the reward components used for Preference Alignment (PA), including ablations of accuracy, format, perception, and style rewards and comparisons of alternative weight configurations. This section provides insight into the contribution of each reward and the basis for the final reward allocation.

\item \textbf{Section~\ref{sec:training} -- Training Details:}  
Complete documentation of model configurations, hyperparameters, optimization strategies, and implementation details across all training stages. This includes specifics for continual pretraining, supervised learning with reasoning traces, and reinforcement learning with humor-aware rewards.

\item \textbf{Section~\ref{sec:prompts} -- Prompt Templates:}  
Full prompt specifications used throughout the pipeline, including those for generating captionist reasoning traces, evaluating outputs via reward models, and performing inference during testing. These templates are critical for reproducing the structured reasoning behavior induced by IRS.

\item \textbf{Section~\ref{sec:inference-conf} -- Inference Configurations:}  
Inference configurations for all models used in evaluation, including model identifiers, access methods, and decoding configurations. API-accessed models use provider defaults, while configurations for locally hosted models are reported where available. 

\item \textbf{Section~\ref{sec:stat-analysis} -- Statistical Analysis:}  
Bootstrap confidence intervals for model performance across evaluation settings.

\item \textbf{Section~\ref{sec:examples} -- Qualitative Examples:}  
Additional examples illustrating model behavior at different stages of training. These include generated reasoning traces, outputs refined through reinforcement learning, curated visual references, judge responses, and comparisons with expert captionist reasoning. They provide a qualitative view of how structured reasoning~evolves. Additional qualitative examples illustrating the limitations discussed in Section~\ref{sec:limitations} are also provided.

\item \textbf{Section~\ref{sec:error_taxonomy} -- Quantitative Error Taxonomy:} A breakdown of IRS-72B model errors across the benchmark splits, with seven error categories and their frequencies reported separately for Matching and Ranking tasks.

\item \textbf{Section~\ref{sec:audit} -- Reasoning-Chain Quality Audit:} A structured audit of reasoning-chain quality using a shared set of humor and reasoning-specific evaluation dimensions, assessing evaluator consistency across independent model judges and comparing IRS against zero-shot and few-shot CoT baselines, with full audit prompts and metric definitions.

\item \textbf{Section~\ref{sec:captioning} -- Zero-shot Caption Generation:}  
Qualitative results demonstrating that IRS-trained models can generate plausible, captionist-style captions despite not being explicitly trained for this task. These examples highlight how structured reasoning learned through IRS transfers to open-ended caption generation, producing outputs grounded in visual incongruity and consistent with editorial humor conventions.

\item \textbf{Section~\ref{app:userstudy} -- User Study Details:}
Documentation of the user study, including participant instructions, consent, recruitment, and how the expert and non-expert references were collected.

\end{itemize}

\section{Dataset and Benchmark Statistics}
\label{sec:all-data-stats}

\renewcommand{\arraystretch}{1.05}
\begin{table*}[t]
\centering
\caption{\textbf{Dataset and evaluation benchmark statistics.} Summary of data volume, task composition, and unique cartoons used across all training stages and evaluation splits.}
\label{tab:dataset_stats}
\begin{tabular}{@{}llllc@{}}
\toprule
\textbf{Component / Split} & \textbf{Purpose} & \textbf{Size} & \begin{tabular}[t]{@{}l@{}}\textbf{Task}\\\textbf{composition}\end{tabular} & \begin{tabular}[t]{@{}l@{}}\textbf{Unique}\\\textbf{Cartoons}\end{tabular} \\
\midrule
\multicolumn{5}{l}{\textbf{Training Datasets}} \\
IM corpus & Continual pretraining & \begin{tabular}[t]{@{}l@{}}4,123 docs;\\6.29M tokens\end{tabular} & \begin{tabular}[t]{@{}l@{}}Humor-specific\\and general text\end{tabular} & -- \\
\addlinespace[2pt]
RM training & Reasoning-trace SFT & 5,163 traces & \begin{tabular}[t]{@{}l@{}}2,208 Matching;\\2,955 Ranking\end{tabular} & 736 \\
\addlinespace[2pt]
PA training & GRPO alignment & 5,163 samples & \begin{tabular}[t]{@{}l@{}}2,208 Matching;\\2,955 Ranking\end{tabular} & 736 \\
\midrule
\multicolumn{5}{l}{\textbf{In-Domain Evaluation}} \\
NYCC test & In-domain evaluation & 685 examples & \begin{tabular}[t]{@{}l@{}}300 Matching;\\385 Ranking\end{tabular} & 100 \\
\addlinespace[2pt]
10-vs-1000 & Hard ranking evaluation & 350 comparisons & Ranking & 35 \\
\addlinespace[2pt]
30-vs-300 & \begin{tabular}[t]{@{}l@{}}Fine-grained\\ranking evaluation\end{tabular} & 350 comparisons & Ranking & 35 \\
\midrule
\multicolumn{5}{l}{\textbf{Zero-Shot Transfer Evaluation}} \\
YesBut & Zero-shot transfer & \begin{tabular}[t]{@{}l@{}}1,044 Philosophy\\1,044 Title\end{tabular} & \begin{tabular}[t]{@{}l@{}}Underlying\\philosophy; Title\end{tabular} & 348 \\
\addlinespace[2pt]
\begin{tabular}[t]{@{}l@{}}DeepEval humor\\subset\end{tabular} & Zero-shot transfer & 29 images & \begin{tabular}[t]{@{}l@{}}DeepSemantics;\\Description; Title\end{tabular} & 29 \\
\bottomrule
\end{tabular}
\end{table*}

Table~\ref{tab:dataset_stats} consolidates the data volumes used across all training stages and evaluation splits, reporting for each the number of examples, its task composition, and the number of unique cartoons it covers.
We report unique-cartoon counts alongside example counts because several benchmark questions or caption comparisons may derive from the same underlying cartoon; separating the two makes the effective visual diversity of each split explicit.

The humor-specific training data consists of NYCC-derived sources, supplemented with general-purpose corpora for broad linguistic and world-knowledge coverage. The Incongruity Modeling (IM) corpus consists of 4,123 documents (6.29M tokens) of humor-specific and general text used for continual pretraining (its composition and temporal coverage are detailed in Appendix~\ref{sec:cpt-corpora}). Resolution Modeling (RM) and Preference Alignment (PA) share the same pool of 5,163 samples (2,208 matching and 2,955 ranking), spanning 736 unique cartoons.

In-domain evaluation uses the full established test splits rather than subsampled subsets: the NYCC test set of 685 examples (300 matching and 385 ranking) over 100 unique cartoons from \citet{hessel-etal-2023-androids}, together with the harder 10-vs-1000 and 30-vs-300 ranking splits, similar to those from \citet{zhou2025bridgingcreativityunderstandinggap} (350 comparisons each, over 35 cartoons). The zero-shot transfer benchmarks contribute no training data: IRS is trained only on the NYCC-derived data above and evaluated directly on YesBut \citep{yesbut-neurips2024} (1,044 Philosophy and 1,044 Title queries over 348 image pairs) and the DeepEval humor subset \citep{yang-etal-2024-large} (29 images). To prevent leakage, all evaluation cartoons are excluded from the continual pretraining corpus, and we verify negligible $n$-gram overlap between the CPT corpus and the evaluation caption text.

\setcounter{figure}{0}
\setcounter{table}{0}

\section{Incongruity Modeling Corpora}
\label{sec:cpt-corpora}
\parahead{Corpus Composition} Our corpus is designed to support the \emph{incongruity modeling} component of \textbf{Incongruity-Resolution Supervision (IRS)} by exposing the model to the discourse, reasoning patterns, and stylistic conventions underlying cartoon humor. Table~\ref{tab:corpus} summarizes the main sources, which fall into three categories:

\begin{itemize}[leftmargin=*]
\item \textbf{Contest deliberations \& roundtables (audio/video)} capture conversational reasoning processes involved in caption generation and evaluation. Sources such as the \textit{New Yorker Cartoon Caption Contest Podcast}, \textit{Official New Yorker Cartoon Podcast}, \textit{CartoonStock YouTube Panel Discussions}, and \textit{The Cartoon Pad} provide rich examples of how incongruities are identified, debated, and refined in practice.
\item \textbf{Editorial \& captionist commentary (written)} provides reflective analyses explaining why specific captions succeed or fail. We include Lawrence Wood’s commentaries on CartoonStock.com, which offer explicit reasoning about humor effectiveness and stylistic choices.
\item \textbf{Books: craft, process, and editorial perspective} distill expert knowledge into concise heuristics and principles. Works such as \textit{The Naked Cartoonist} \citep{mankoff2002nakedcartoonist}, \textit{Your Caption Has Been Selected} \citep{wood2024your}, and \textit{How About Never-Is Never Good for You?} \citep{mankoff2014howaboutnever} provide structured guidance on creativity, interpretation, and editorial decision-making.
\end{itemize}

\renewcommand{\arraystretch}{1.05} 
\begin{table*}[t]
\centering
\caption{\textbf{Composition of the IM corpus.} The corpus is dominated by conversational and editorial sources, reflecting the importance of reasoning processes and stylistic conventions in humor understanding.}
\begin{tabular}{lrrrr}
\toprule
\textbf{Source} & \textbf{Instances} & \textbf{Words} & \textbf{Tokens} & \textbf{\% of Tokens} \\
\midrule
\multicolumn{5}{l}{\textbf{Contest Deliberations and Commentaries}} \\
New Yorker Caption Contest Podcast & 173 & 2,292,458 & 2,921,356 & 46.43\% \\
The Cartoon Pad & 43 & 443,774 & 568,849 & 9.04\% \\
Official New Yorker Cartoon Podcast & 34 & 349,588 & 465,701 & 7.40\% \\
CartoonStock YouTube Panel Discussions & 36 & 224,864 & 281,295 & 4.47\% \\
CartoonStock Lawrence Wood Commentaries & 175 & 128,635 & 170,978 & 2.72\% \\
\midrule
\multicolumn{5}{l}{\textbf{Books on Caption Writing}} \\
How About Never & 1 & 40,621 & 56,010 & 0.89\% \\
Your Caption Has Been Selected & 94 & 8,087 & 11,057 & 0.18\% \\
The Naked Cartoonist & 1 & 3,062 & 4,076 & 0.06\% \\
\midrule
\multicolumn{5}{l}{\textbf{General-Purpose Corpora}} \\
FineWeb & 1759 & 979,418 & 1,308,241 & 20.79\% \\
Olmo-Mix-1124 & 1807 & 353,705 & 504,177 & 8.01\% \\
\midrule
\textbf{Total} & 4123 & 4,824,212 & 6,291,740 & 100.00\% \\
\bottomrule
\end{tabular}
\label{tab:corpus}
\end{table*}
\renewcommand{\arraystretch}{1.05} 
\begin{table*}[!h]
\caption{\textbf{Temporal coverage of humor-specific sources.}
The corpus spans nearly a decade of caption contest discourse, capturing both stable stylistic conventions and evolving humor preferences.}
\centering
\begin{tabular}{l@{$\;\;\;$}c@{$\;\;\;$}c@{$\;\;\;$}c}
\toprule
\textbf{Source} & \textbf{First episode} & \textbf{Last episode} & \textbf{Coverage} \\
\midrule
NYCC Podcast & Mar 18, 2021 (Ep. 1) & Sep 25, 2024 (Ep. 173) & 3.5 years \\
The Cartoon Pad & Apr 2, 2021 (Ep. 1) & Jul 24, 2024 (Ep. 44) & 3 years \\
Official NYer Cartoon Podcast & Nov 14, 2015 (Ep. 29) & Jan 22, 2021 (Ep. 289) & 5+ years \\
Cartoon Stock YouTube Podcasts & Dec 20, 2022 & Jul 25, 2025 & 2.5 years \\
Lawrence Wood Commentaries & Jul 12, 2019 & Oct 22, 2024 & 5 years \\
\bottomrule
\end{tabular}
\label{tab:temporal}
\end{table*}

To ensure stable optimization and prevent overfitting to narrow humor-specific distributions, we additionally incorporate subsets of FineWeb~\citep{penedo2024the} and OLMo-Mix-1124~\citep{allenai_olmo_mix_1124_2024}. These data provide broad linguistic coverage and general world knowledge, which are essential for interpreting cultural references and contextual nuances frequently used in cartoon captions. All items overlapping evaluation cartoons or captions are removed to avoid leakage.

\parahead{Temporal Coverage} 
As summarized in Table~\ref{tab:temporal}, our podcast and commentary sources span nearly a decade of caption contest discourse. The \textit{Official New Yorker Cartoon Podcast} provides the earliest coverage (2015-2021), while more recent series such as the NYCC Podcast (2021-2024) and \textit{The Cartoon Pad} (2021-2024) extend into the present. More recent commentary comes from CartoonStock YouTube Panels through July 2025, complemented by Lawrence Wood’s weekly contest analyses (2019-2024).

CartoonStock is an independent cartoon archive and marketplace that hosts a caption contest similar to NYCC. Its panel discussions and Wood’s critiques are particularly valuable, as they explicitly analyze finalist and winning captions, highlighting both successful and unsuccessful interpretations. This complements NYCC data by exposing the model to diverse reasoning styles and evaluative criteria across platforms.

\parahead{Licensing note} Parts of the IM corpus are derived from copyrighted sources (e.g., books and podcast transcripts). Due to licensing restrictions, these materials cannot be directly released. To ensure reproducibility, we will release all train/test splits used for evaluation, along with preprocessing code and prompt templates required to reconstruct the pipeline. This balances transparency with respect for intellectual property constraints.

\renewcommand{\arraystretch}{1.05}
\begin{table*}[!h]
\renewcommand{\thetable}{D.1}
\centering
\caption{\textbf{Reward composition and weight allocation.} Performance under different reward components and weight configurations. Reward configurations are reported as $(R_a,R_p,R_s,R_f)$, corresponding to accuracy, perception, style, and format rewards, respectively.}
\label{tab:reward-comp-weights}
\setlength{\tabcolsep}{5pt}
\begin{tabular}{llrrrr}
\toprule
\multirow{2}{*}{\textbf{Reward Configuration}} & \multirow{2}{*}{\textbf{Configuration}} & \multicolumn{2}{c}{\textbf{(Hessel et al., 2023)}} & \multicolumn{2}{c}{\textbf{(Zhou et al., 2025)}}\\
& & \textbf{Matching} & \textbf{Ranking} & \textbf{10-vs-1000} & \textbf{30-vs-300} \\
\midrule
$(0.8,0,0,0.2)$ & Accuracy + Format & 60.67 & 57.99 & 54.57 & 50.86 \\
$(0.4,0.4,0,0.2)$ & + Perception Reward & 58.00 & 60.78 & 56.57 & 49.43 \\
$(0.4,0,0.4,0.2)$ & + Style Reward & 60.00 & 58.44 & 54.57 & 49.43 \\
$(0.3,0.3,0.3,0.1)$ & Equal Reward Weights & 61.00 & 59.74 & 54.57 & 50.29 \\
$(0.4,0.2,0.2,0.2)$ & Final & 59.67 & \textbf{64.42} & \textbf{56.29} & \textbf{53.14} \\
\bottomrule
\end{tabular}
\end{table*}

\section{Resolution Modeling Corpora}
\label{sec:sft_corpora}

To construct high-quality, reliable reasoning traces without relying on unverified model outputs or same-family self-audits, we implement a multi-stage human-in-the-loop curation and verification pipeline.

\parahead{Corpus Composition} Rather than directly adopting raw outputs from DeepSeek-R1 \citep{deepseekai2025deepseekr1incentivizingreasoningcapability}, reasoning chains undergo rigorous annotation, filtering, and cross-family refinement:
\begin{enumerate}
    \item \textbf{Visual and Contextual Annotation Review:} Annotators manually inspect and correct the base NYCC scene descriptions, entity tags, and incongruity annotations to resolve ambiguities or missing visual details.
    \item \textbf{Visual Grounding Verification:} Automatically expanded descriptions are audited to eliminate hallucinations and retain only details explicitly grounded in the image.
    \item \textbf{Initial Trace Generation:} DeepSeek-R1 generates preliminary reasoning traces conditioned on candidate captions and the human-verified visual metadata.
    \item \textbf{Automated Heuristic Screening:} Traces are screened via automated checks to filter out answer leakage, shortcut cues, visual inconsistencies, and deviations from prescribed reasoning structures.
    \item \textbf{Targeted Human Correction:} Flagged traces undergo manual inspection and are corrected, regenerated, or rejected. Revised traces are re-screened prior to inclusion.
    \item \textbf{Cross-Family Rewriting:} To prevent family-specific bias, GPT-4o \citep{openai2024gpt4o} rewrites accepted traces into concise, image-grounded captionist-style prose. A subset of rewritten traces is manually reviewed to verify reasoning preservation.
    \item \textbf{Final Quality Pass:} A final automated audit validates formatting compliance, reasoning coherence, and label consistency across the dataset.
\end{enumerate}

\paragraph{Annotation Team and Quality Control.}
The annotation and verification process involved three annotators: one graduate and two undergraduate researchers with backgrounds in computer science and prior experience with multimodal language models and NYCC-style humor data. All annotators were proficient in English. Uncertain cases were discussed and resolved jointly.
\label{sec:sft-corpora}

\section{Reward Composition and Weight Allocation}
\label{sec:reward-weights}
\setlength{\tabcolsep}{5pt}
\renewcommand{\arraystretch}{1.05} 
\begin{table*}[h]
\renewcommand{\thetable}{E.1}
\centering
\caption{\textbf{Training hyperparameters across IRS (7B) stages.} Each stage supports a different component of IRS, with distinct optimization regimes and resource requirements.}
\label{tab:hyperparams}
\begin{tabular}{ll@{$\;\;\;$}c@{$\;\;\;$}c@{$\;\;\;$}c@{$\;\;\;$}c@{$\;\;\;$}c@{$\;\;\;$}c@{$\;\;\;$}c}
\toprule
\textbf{Stage} & \textbf{Base Model} & \textbf{Adapter} & \textbf{Epochs} & \textbf{Batch} & \textbf{LR} & \textbf{Precision} & \textbf{GPUs} & \textbf{Duration} \\
\midrule
IM & Qwen2.5-VL-7B & \begin{tabular}[t]{@{}l@{}}LoRA\\(rank 32)\end{tabular} & 50 & 1 & 1e-4 & bf16 & 2$\times$H100 & $\sim$4 h \vspace{0.15cm}\\
RM & IM model & \begin{tabular}[t]{@{}l@{}}LoRA\\(rank 64)\end{tabular} & 7 & 16 & 1e-4 & bf16 & 2$\times$ H100 & $\sim$3 h \vspace{0.15cm} \\
PA  & RM model & Full model & 5 & 16 (roll.) & 1e-6 & bf16 & 4$\times$H100 & $\sim$1.5 d \\
\bottomrule
\end{tabular}
\label{table:hyperparams}
\end{table*}

Table~\ref{tab:reward-comp-weights} reports the reward-component ablation used to determine the final GRPO reward composition. The baseline combines the accuracy and format rewards. Adding the perception reward produces the largest individual improvement, increasing Ranking accuracy from 57.99 to 60.78 (+2.79) and 10-vs-1000 accuracy from 54.57 to 56.57 (+2.00). Adding the style reward alone yields a smaller improvement on Ranking (+0.45), indicating that the two auxiliary rewards provide complementary supervision.

We further compare two configurations that combine all reward components. Equal weighting of the three rewards (0.3, 0.3, 0.3, 0.1) achieves the highest Matching accuracy but underperforms on Ranking and both larger candidate-set evaluations. The final configuration (0.4, 0.2, 0.2, 0.2) yields the best overall trade-off, improving Ranking by 4.68 points, 10-vs-1000 by 1.72 points, and 30-vs-300 by 2.85 points over the equal-weight configuration, while incurring a 1.33-point decrease in Matching accuracy. We therefore assign the largest weight to the accuracy reward, while equally weighting the perception and style rewards. This allocation prioritizes answer correctness while retaining complementary supervision for visual incongruity and captionist-style reasoning.

\section{Training Details}
\label{sec:training}
We describe the training procedures used to implement \textit{Incongruity-Resolution Supervision (IRS)}. These consist of three stages: Incongruity Modeling (IM), Resolution Modeling (RM), and Preference Alignment (PA). Table~\ref{tab:hyperparams} summarizes hyperparameters across the stages.

\begin{figure*}[!h]
    \renewcommand{\thefigure}{E.1}
    \centering
    \begin{minipage}{0.32\textwidth}
        \centering
        \includegraphics[width=\linewidth]{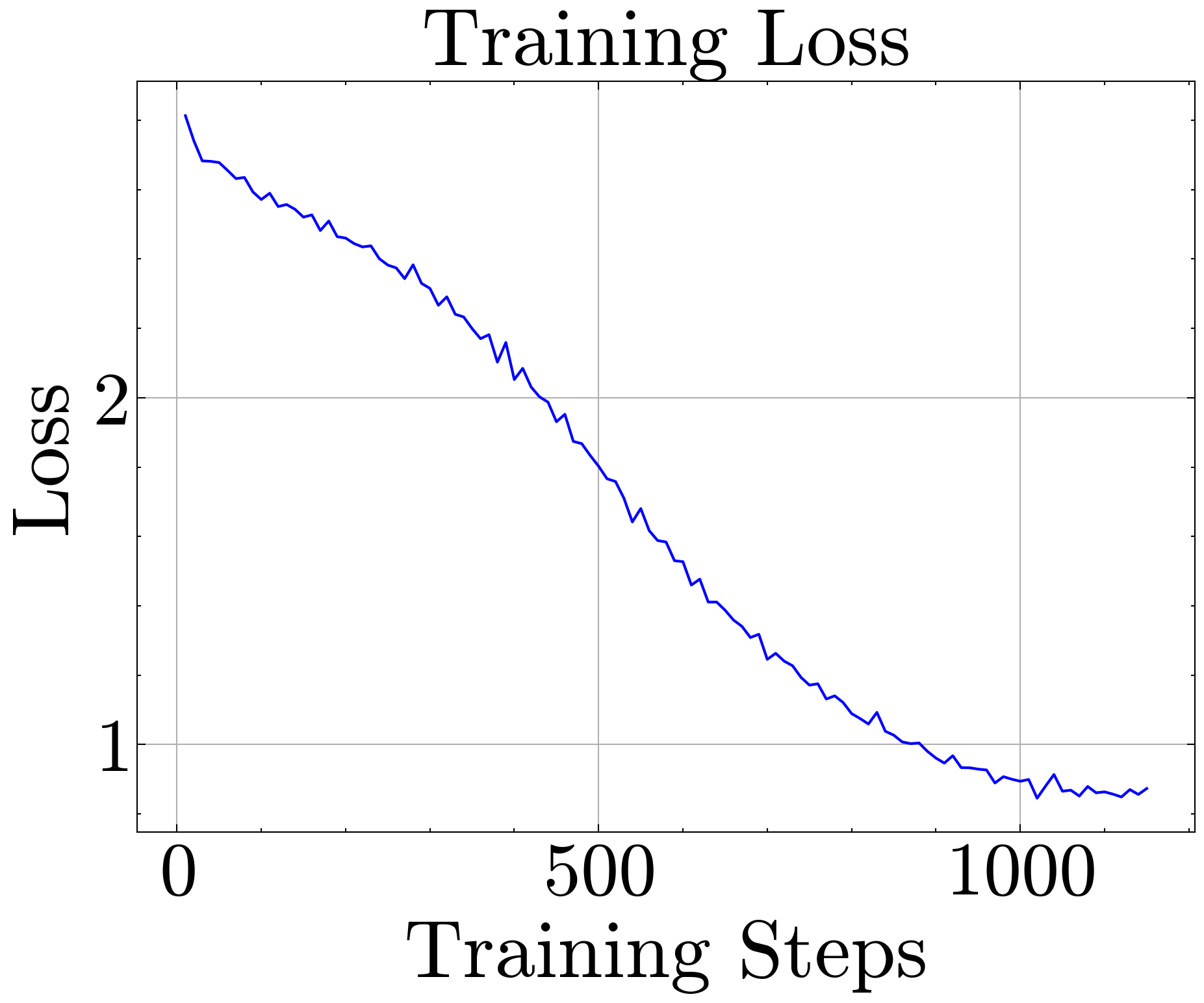}
    \end{minipage}
    \hfill
    \begin{minipage}{0.32\textwidth}
        \centering
        \includegraphics[width=\linewidth]{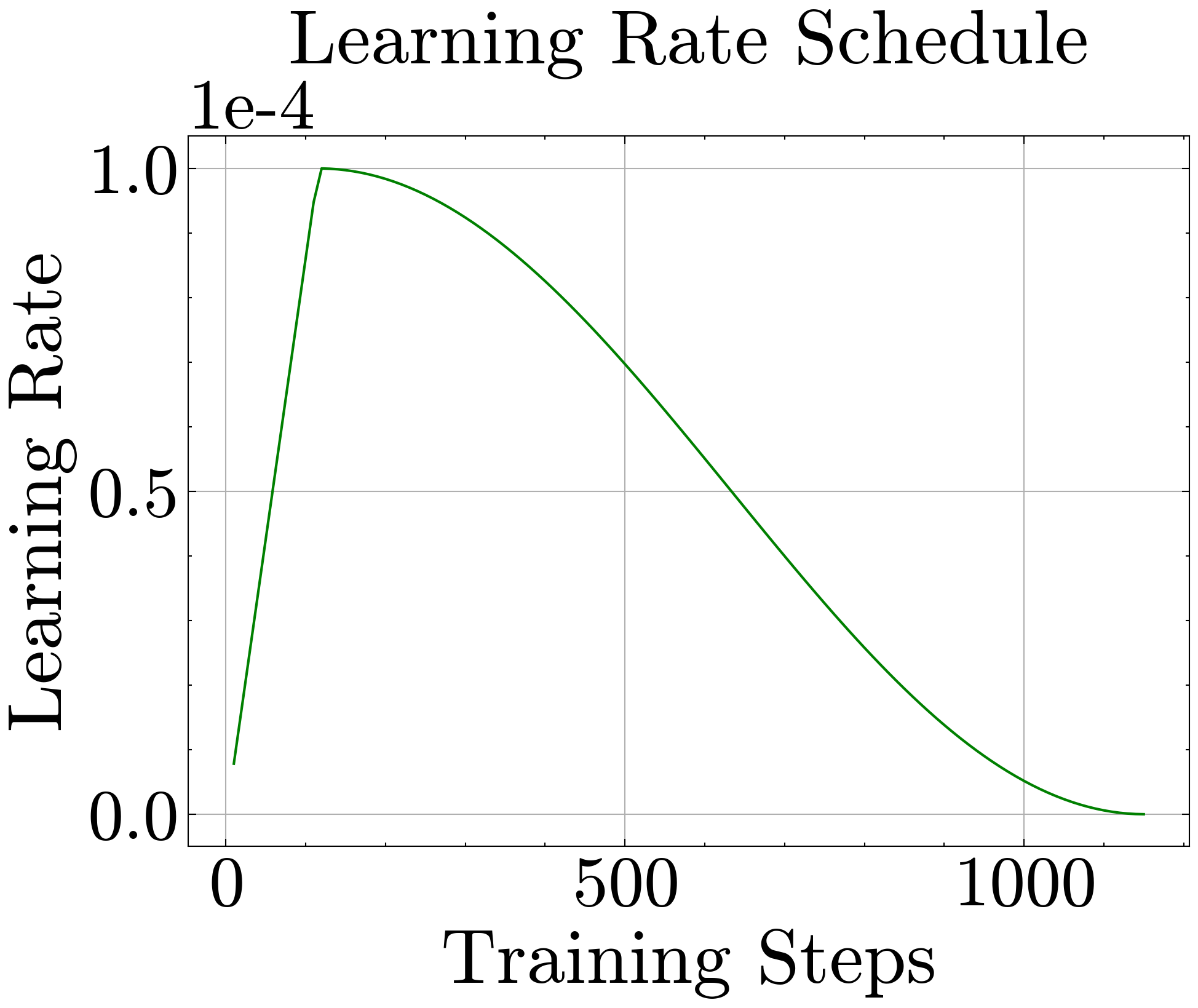}
    \end{minipage}
    \hfill
    \begin{minipage}{0.32\textwidth}
        \centering
        \includegraphics[width=\linewidth]{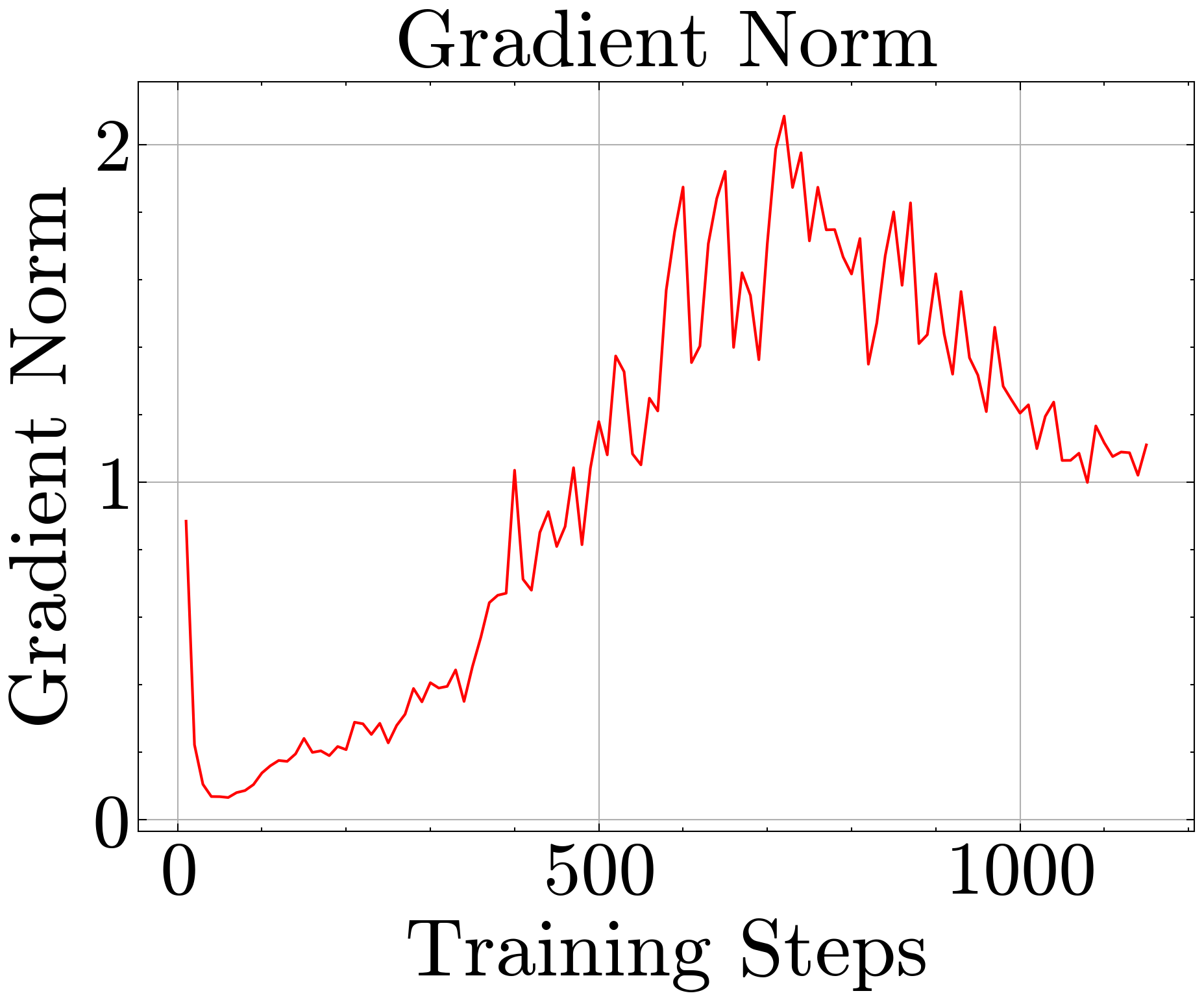}
    \end{minipage}\vspace{-1em}
    \caption{\textbf{IM dynamics of the 7B model.} Loss decreases steadily over training, indicating successful adaptation to humor-specific discourse. The gradient norm remains controlled overall, with larger spikes appearing later in training.}\vspace{-1em}
    \label{fig:CPT}
\end{figure*}
\begin{figure*}[!h]
\renewcommand{\thefigure}{E.2}
    \centering
    \begin{minipage}{0.32\textwidth}
        \centering
        \includegraphics[width=\linewidth]{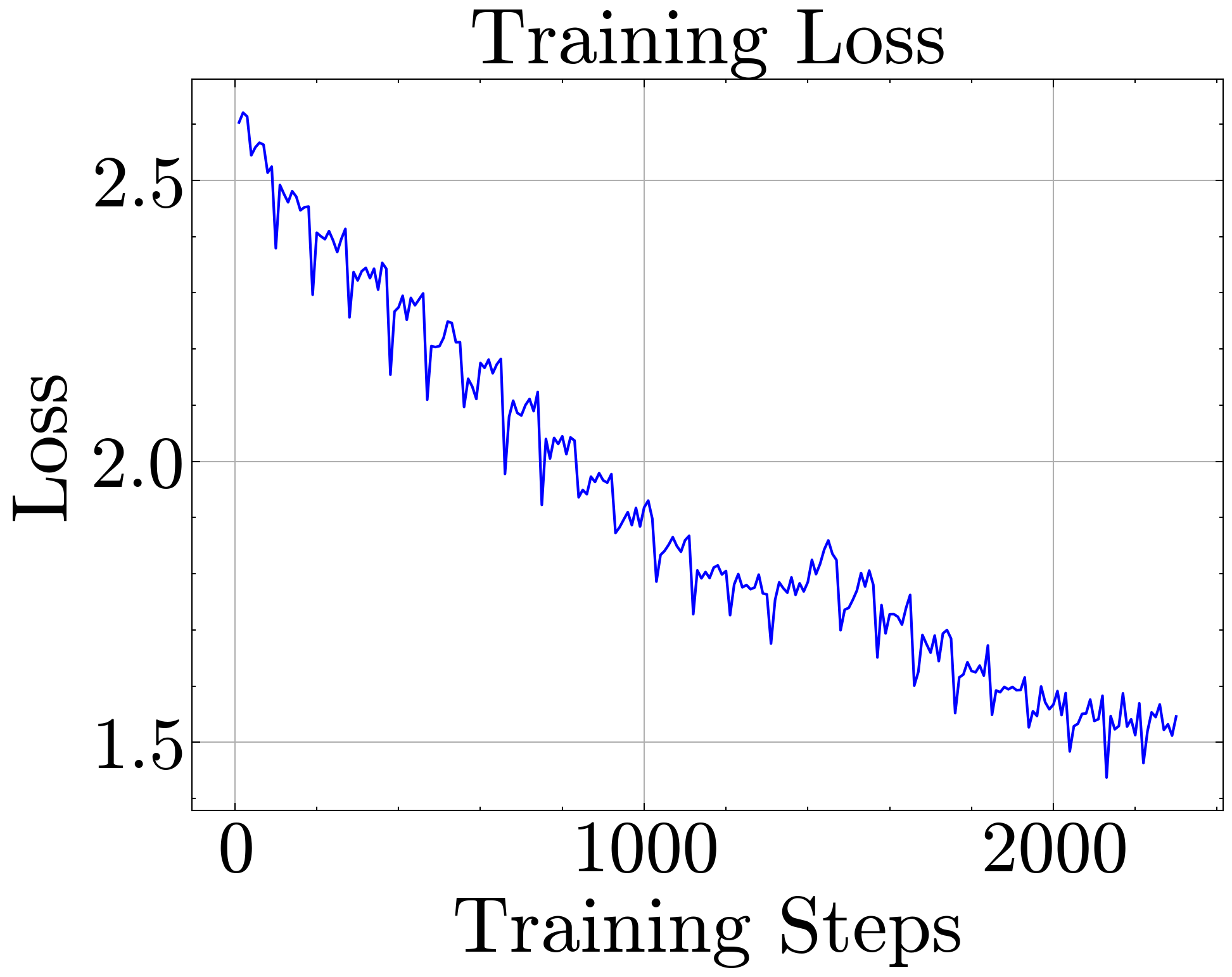}
    \end{minipage}
    \hfill
    \begin{minipage}{0.32\textwidth}
        \centering
        \includegraphics[width=\linewidth]{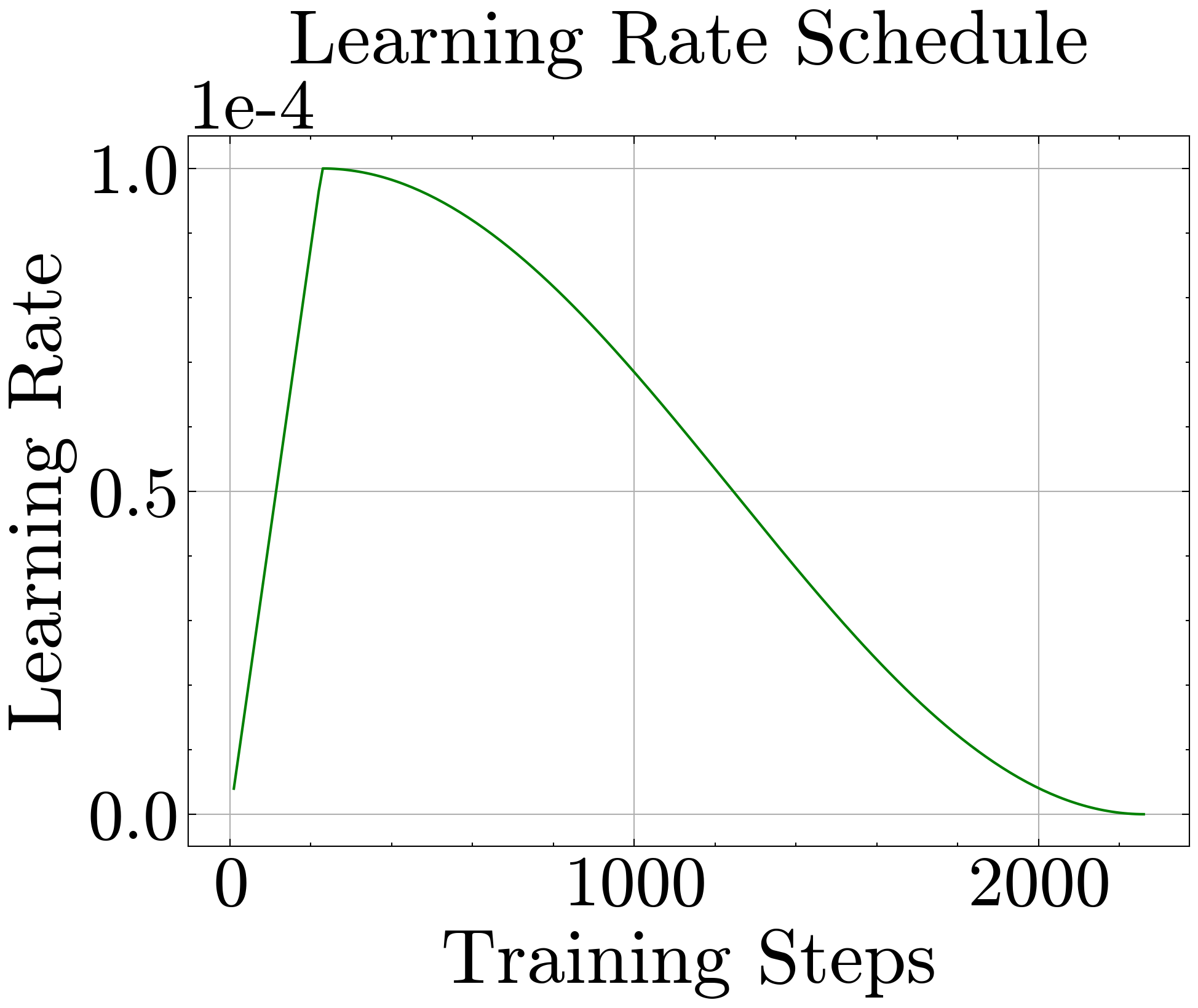}
    \end{minipage}
    \hfill
    \begin{minipage}{0.32\textwidth}
        \centering
        \includegraphics[width=\linewidth]{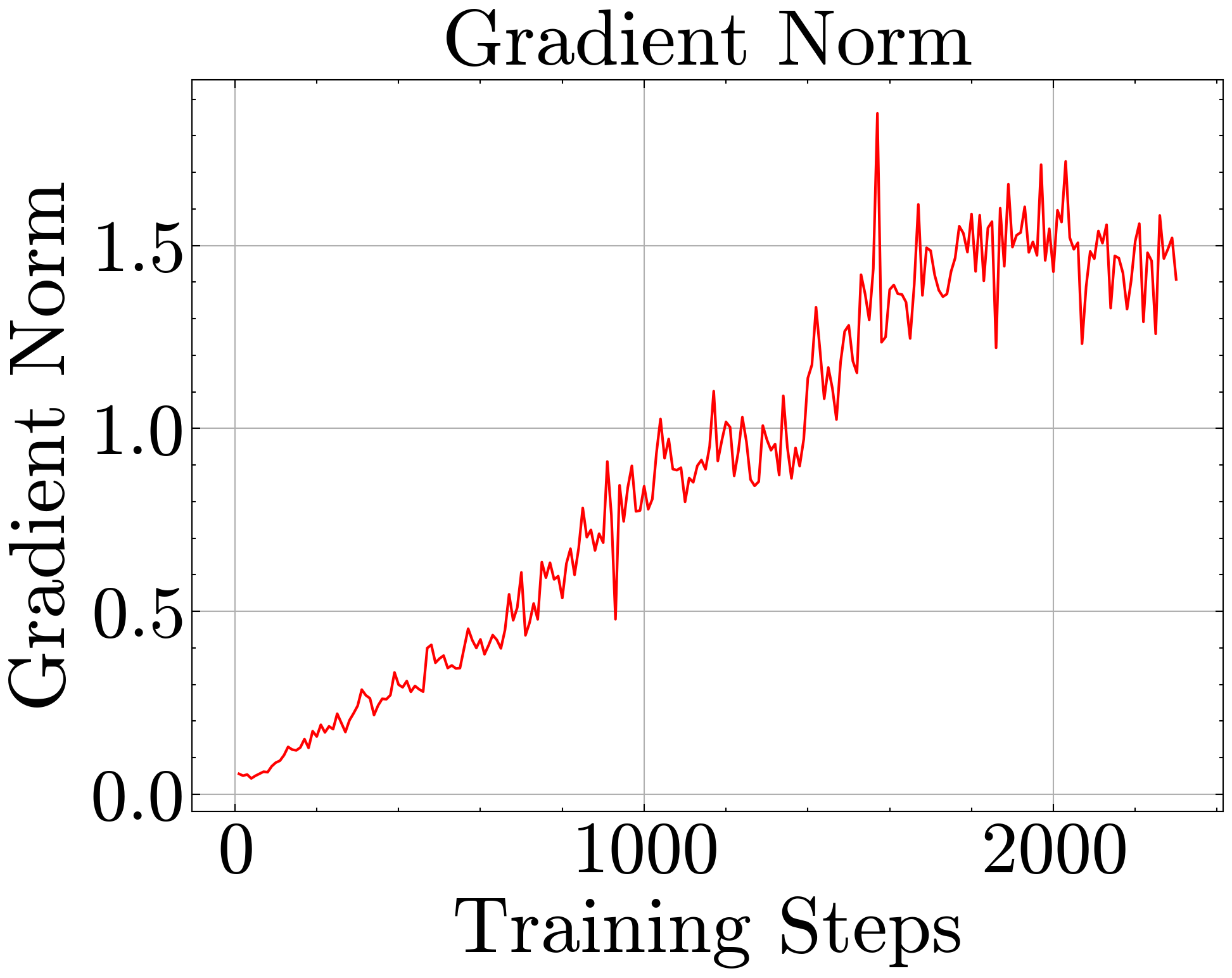}
    \end{minipage}\vspace{-1em}
    \caption{\textbf{IM dynamics of the 32B model.} Loss decreases steadily despite mild early fluctuations, indicating successful adaptation to humor-specific discourse. The gradient norm increases gradually over training while remaining stable overall.}\vspace{-1em}
    \label{fig:CPT32B}
\end{figure*}
\begin{figure*}[!h]
\renewcommand{\thefigure}{E.3}
    \centering
    \begin{minipage}{0.32\textwidth}
        \centering
        \includegraphics[width=\linewidth]{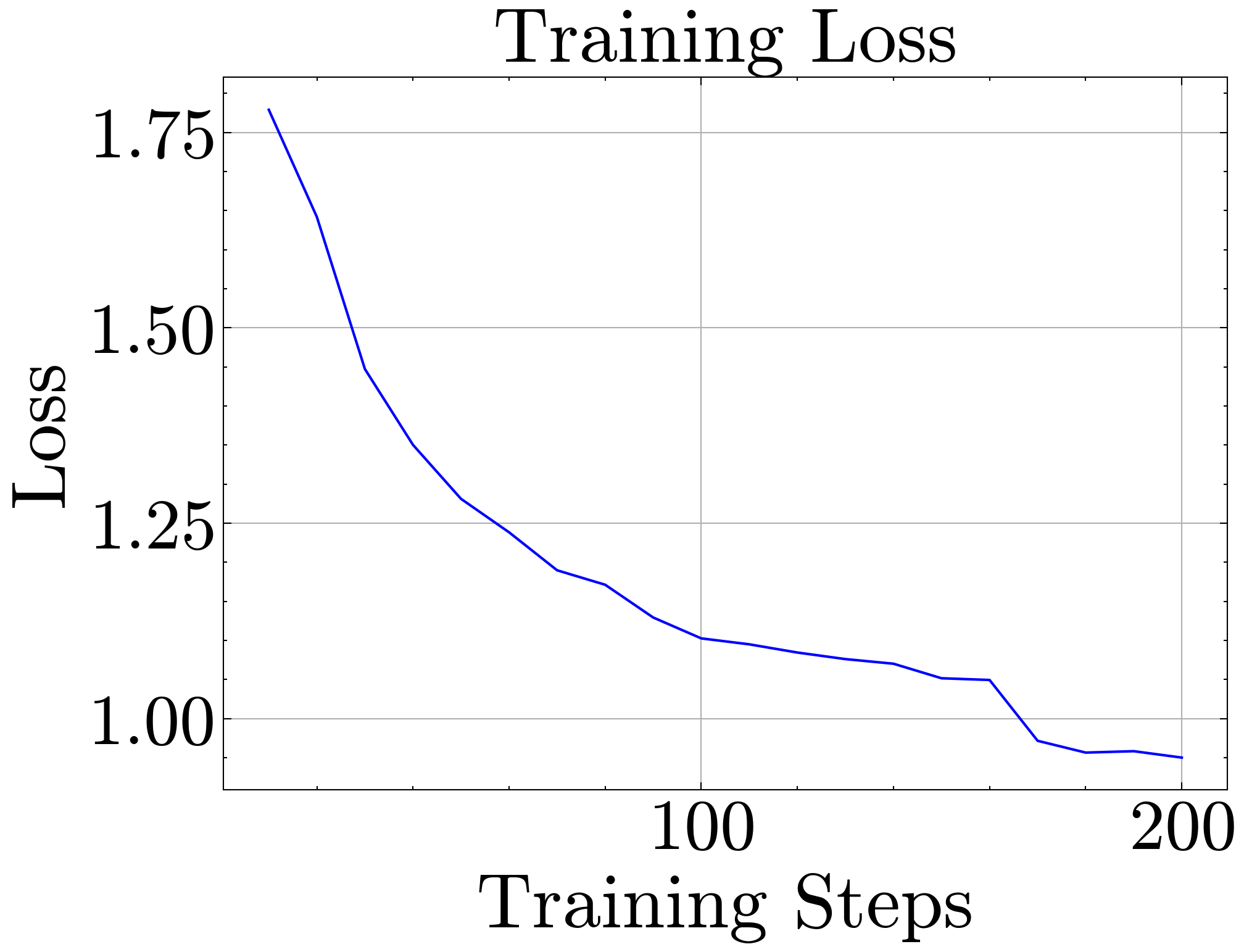}
    \end{minipage}
    \hfill
    \begin{minipage}{0.32\textwidth}
        \centering
        \includegraphics[width=\linewidth]{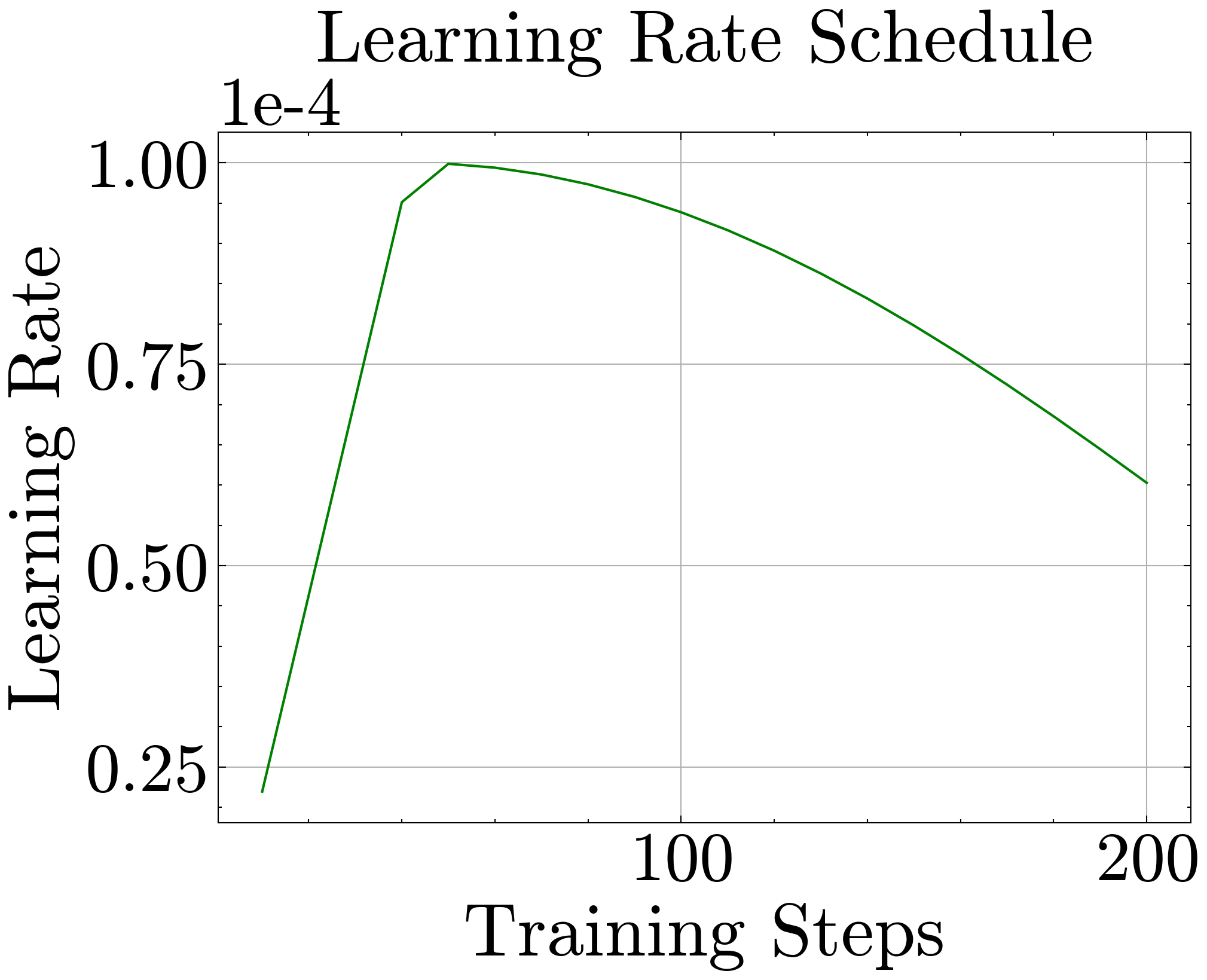}
    \end{minipage}
    \hfill
    \begin{minipage}{0.32\textwidth}
        \centering
        \includegraphics[width=\linewidth]{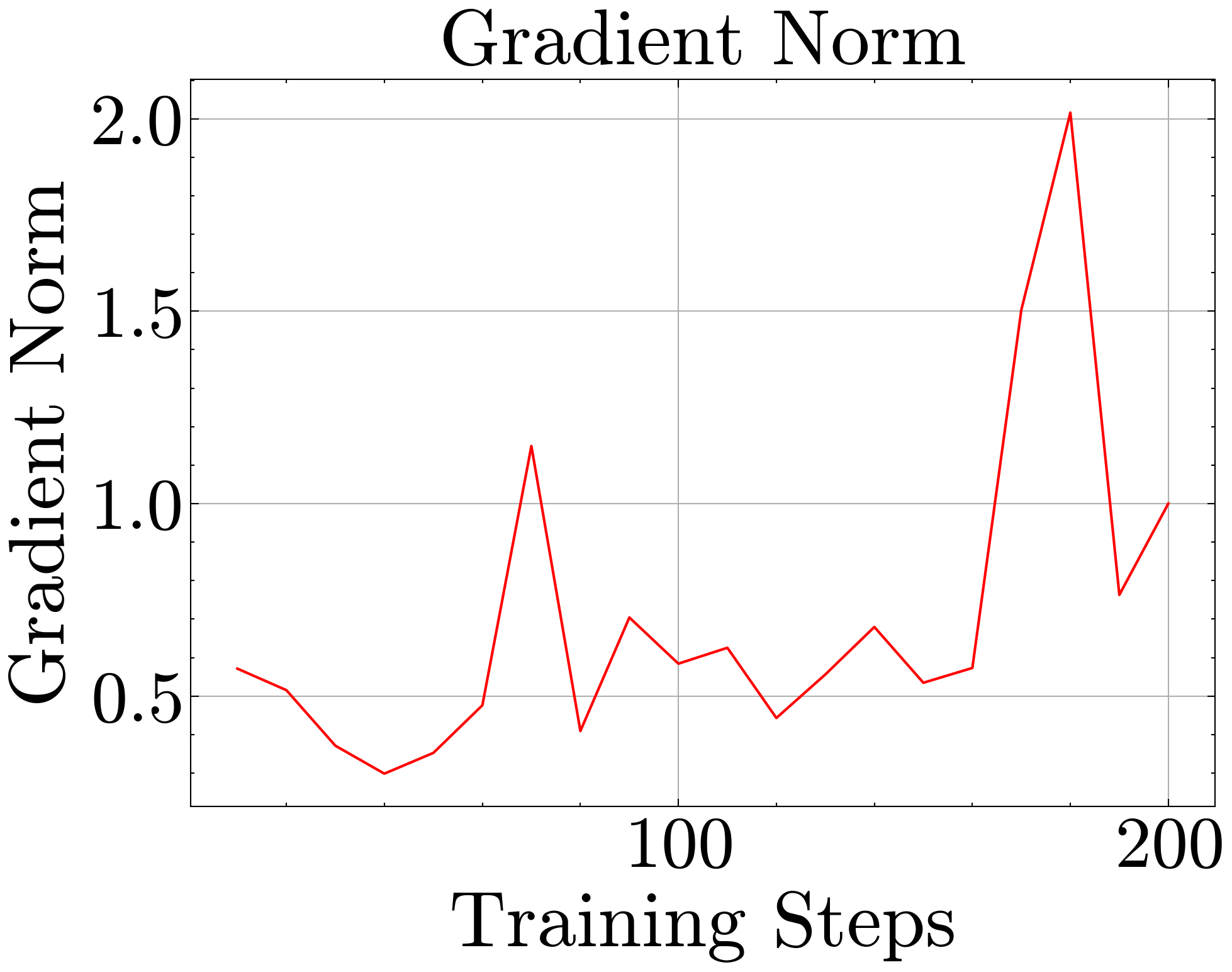}
    \end{minipage}\vspace{-1em}
    \caption{\textbf{IM dynamics of the 72B model.} Loss drops rapidly and then stabilizes, indicating effective early adaptation to humor-specific discourse. The gradient norm remains mostly low but exhibits occasional sharp spikes.}\vspace{-1em}
    \label{fig:CPT72B}
\end{figure*}

\begin{figure*}[!h]
\renewcommand{\thefigure}{E.4}
    \centering
    \begin{minipage}{0.32\textwidth}
        \centering
        \includegraphics[width=\linewidth]{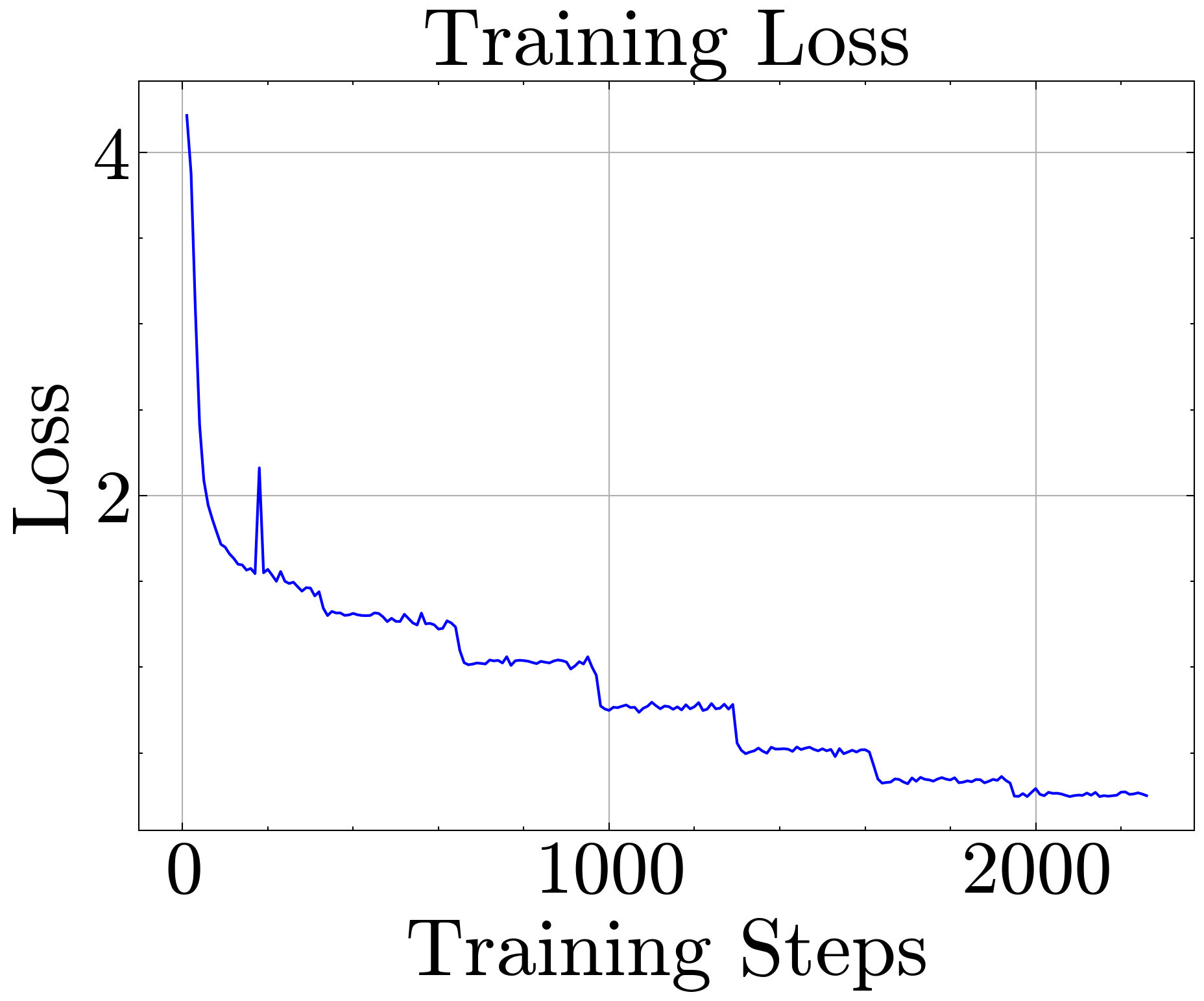}
    \end{minipage}
    \hfill
    \begin{minipage}{0.32\textwidth}
        \centering
        \includegraphics[width=\linewidth]{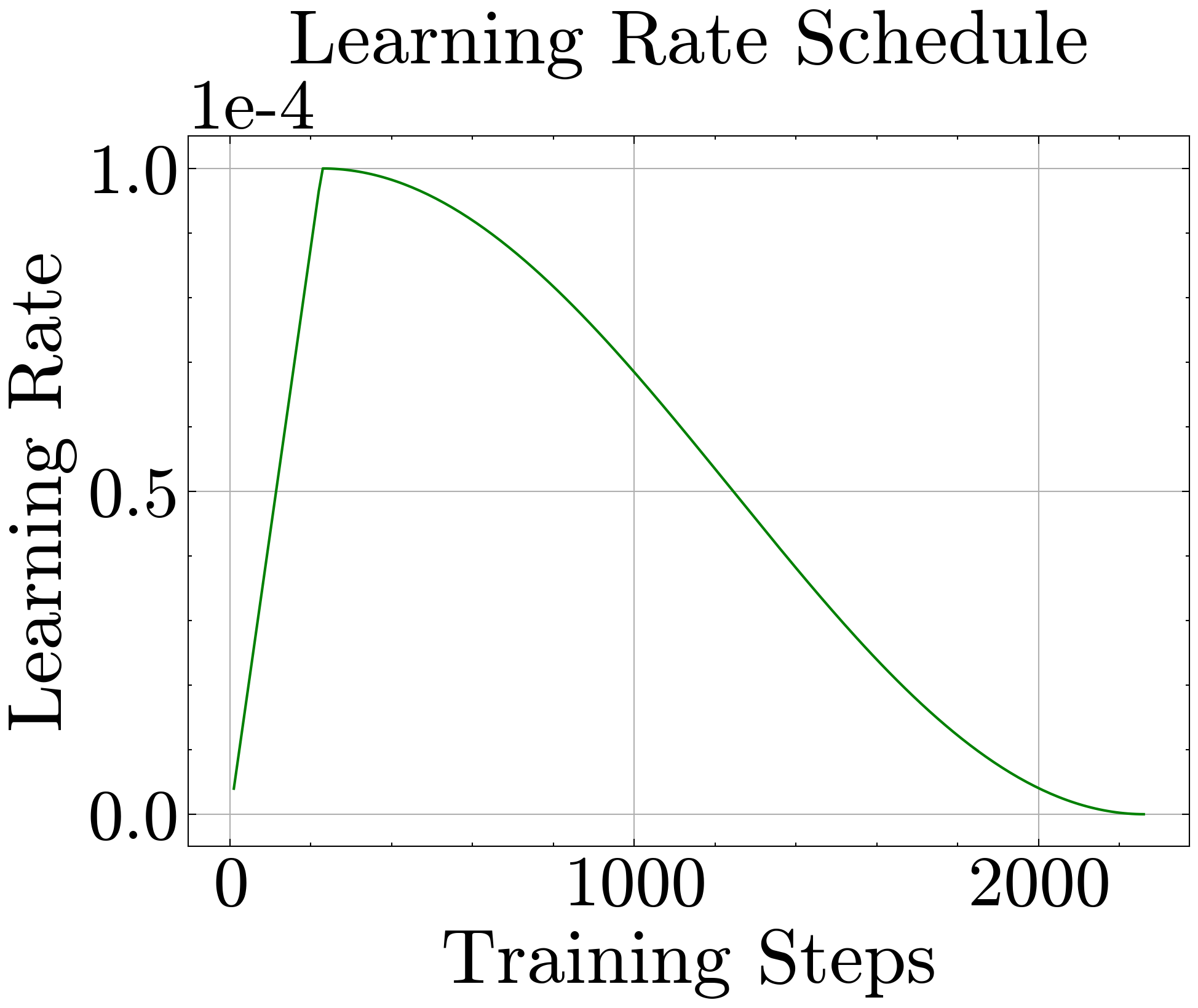}
    \end{minipage}
    \hfill
    \begin{minipage}{0.32\textwidth}
        \centering
        \includegraphics[width=\linewidth]{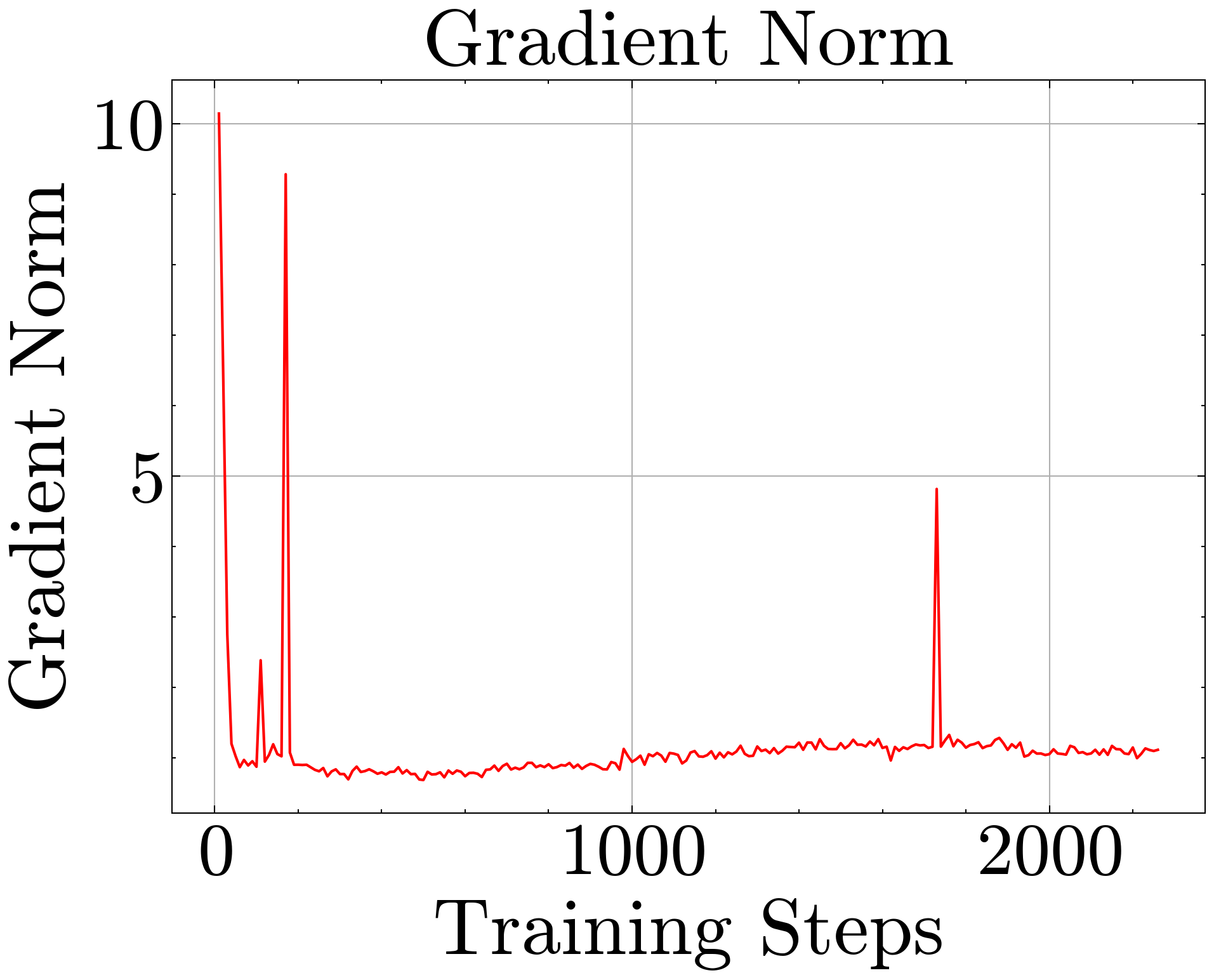}
    \end{minipage}\vspace{-0.5em}
    \caption{\textbf{RM dynamics of the 7B model.} Rapid loss reduction during warmup followed by stable convergence indicates effective learning of structured reasoning from captionist traces.}
    \label{fig:SFT}
\end{figure*}

\begin{figure*}[!h]
\renewcommand{\thefigure}{E.5}
    \centering
    \begin{minipage}{0.32\textwidth}
        \centering
        \includegraphics[width=\linewidth]{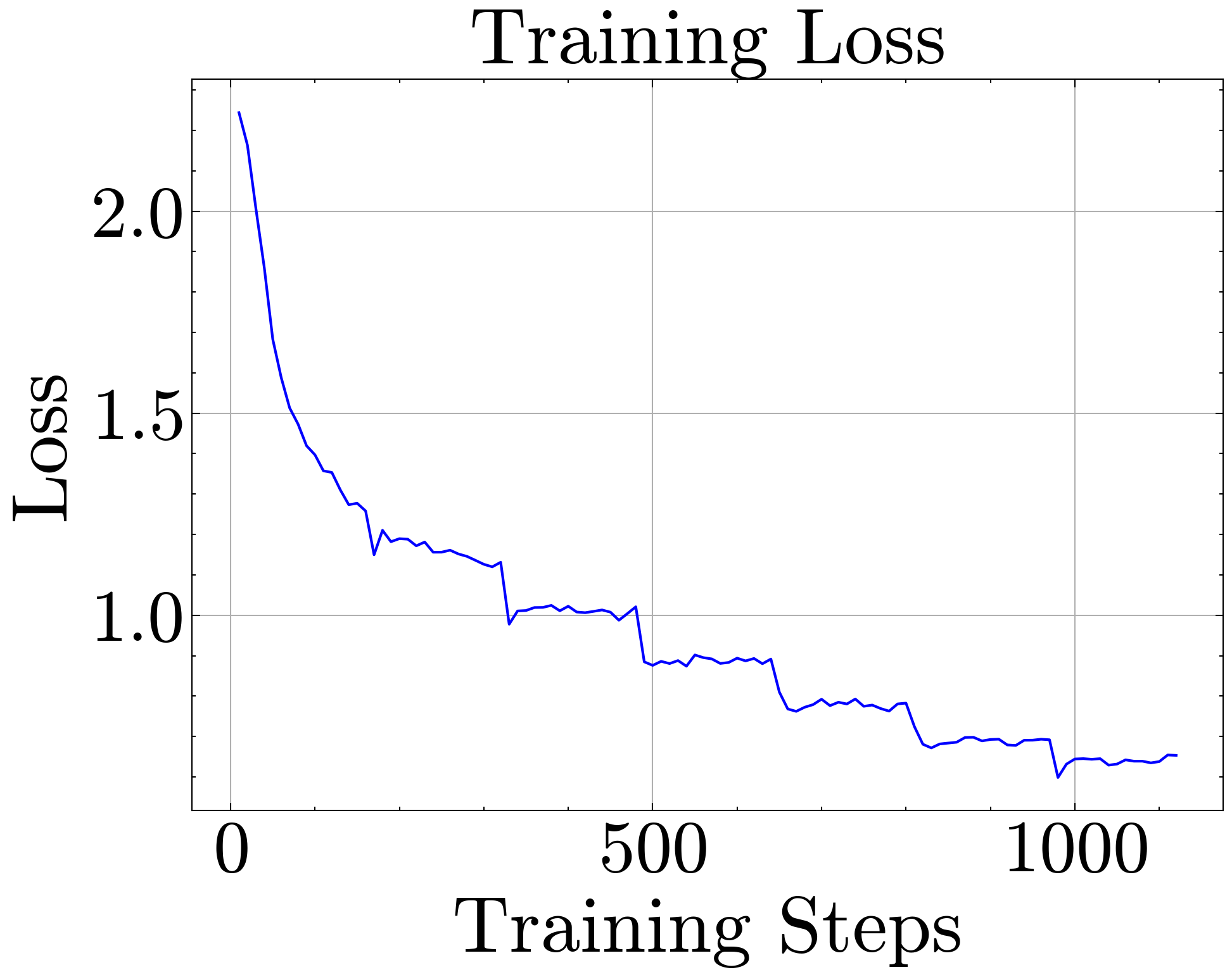}
    \end{minipage}
    \hfill
    \begin{minipage}{0.32\textwidth}
        \centering
        \includegraphics[width=\linewidth]{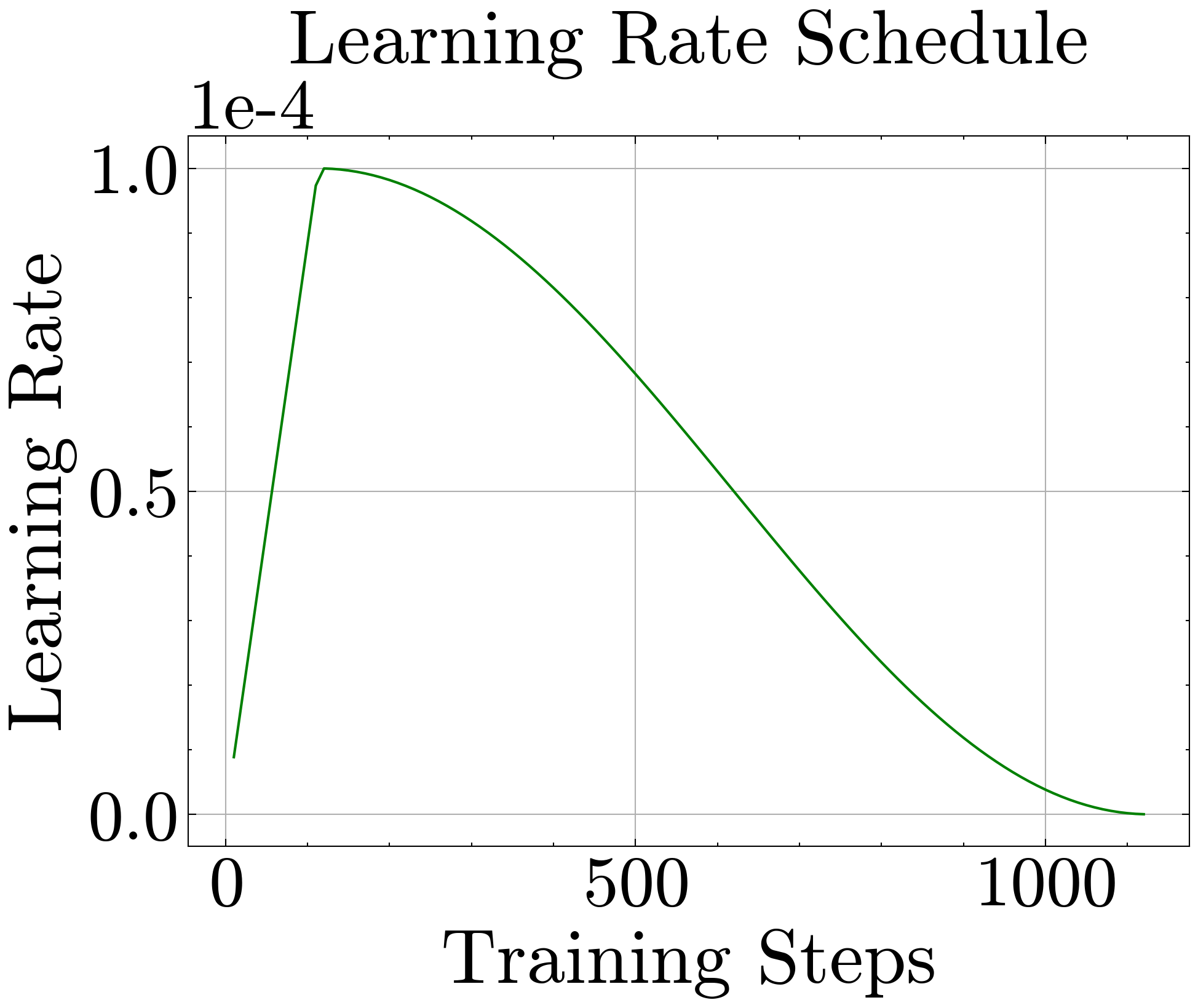}
    \end{minipage}
    \hfill
    \begin{minipage}{0.32\textwidth}
        \centering
        \includegraphics[width=\linewidth]{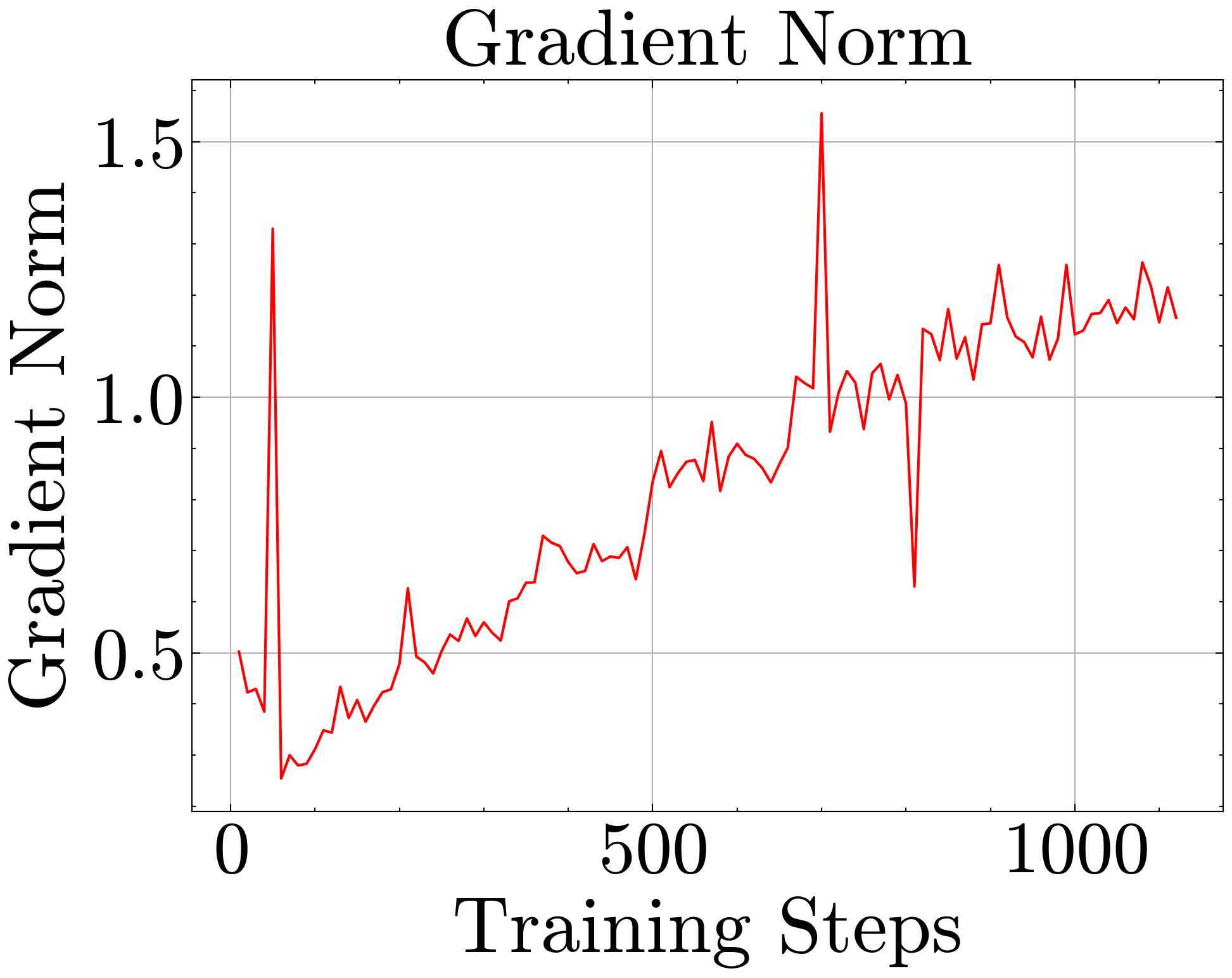}
    \end{minipage}\vspace{-0.5em}
    \caption{\textbf{RM dynamics of the 32B model.} Loss decreases smoothly after warmup, indicating effective adaptation to structured reasoning supervision. The gradual increase in gradient norm suggests continued refinement of reasoning patterns during training.}
    \label{fig:SFT32B}
\end{figure*}

\begin{figure*}[!h]
\renewcommand{\thefigure}{E.6}
    \centering
    \begin{minipage}{0.32\textwidth}
        \centering
        \includegraphics[width=\linewidth]{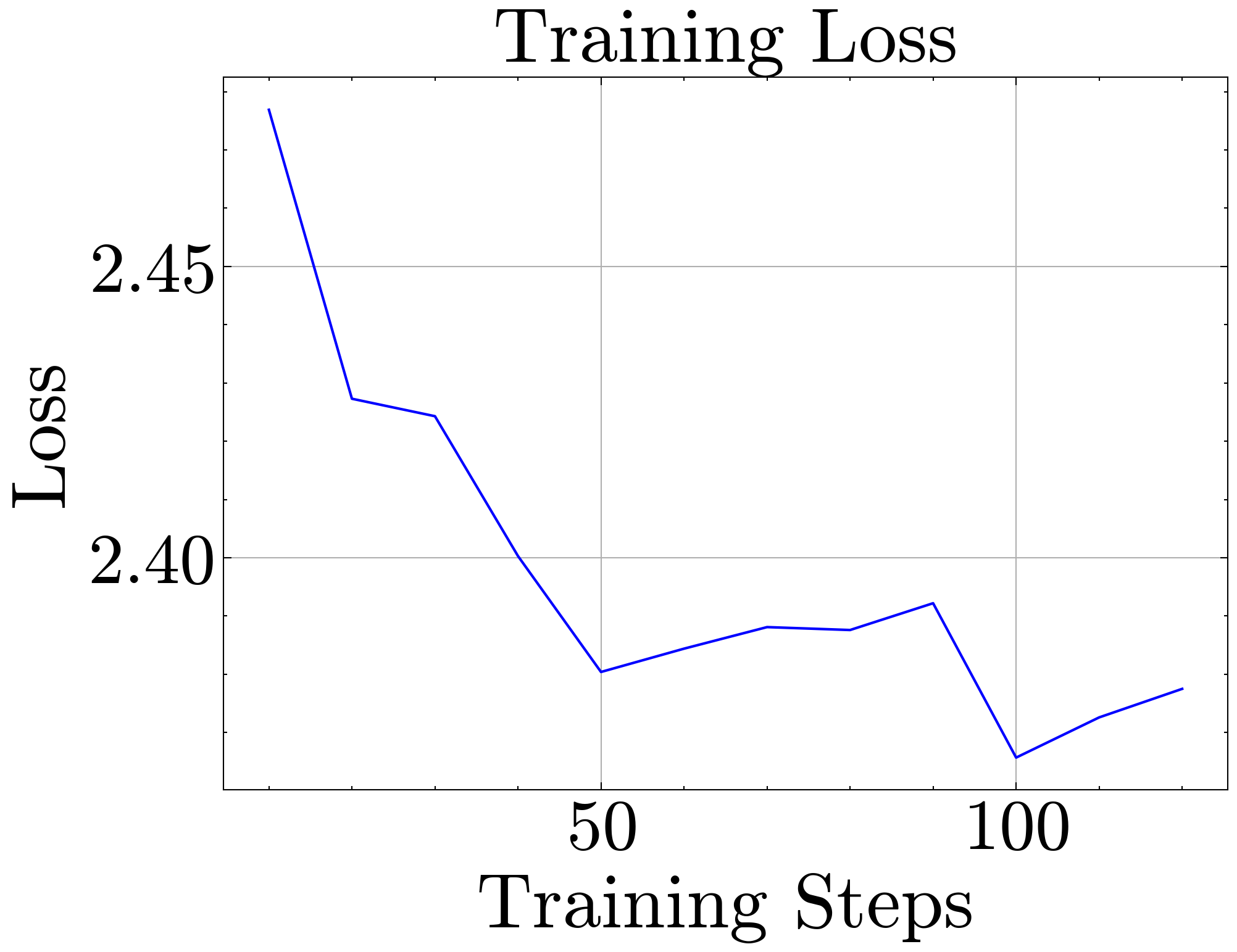}
    \end{minipage}
    \hfill
    \begin{minipage}{0.32\textwidth}
        \centering
        \includegraphics[width=\linewidth]{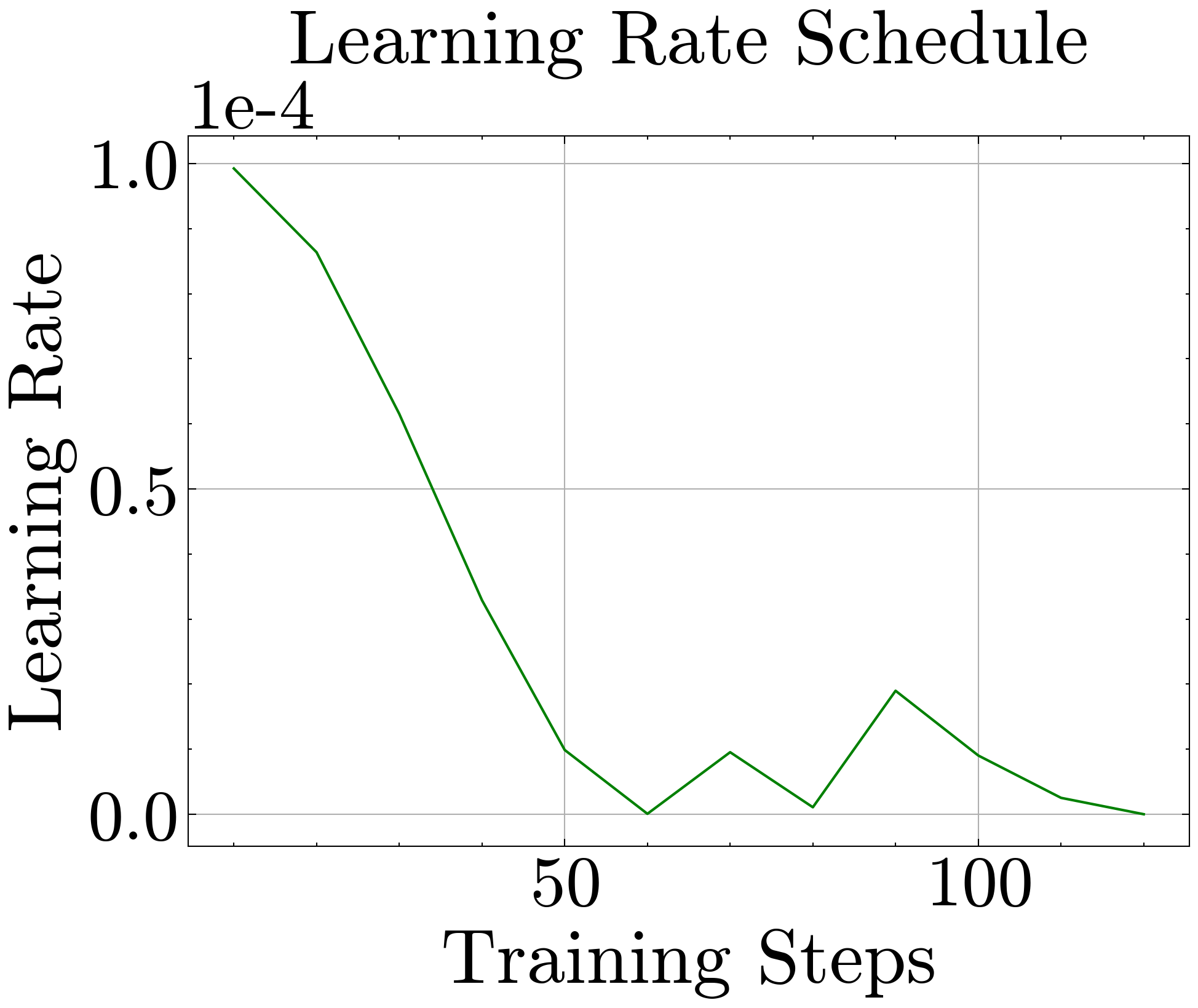}
    \end{minipage}
    \hfill
    \begin{minipage}{0.32\textwidth}
        \centering
        \includegraphics[width=\linewidth]{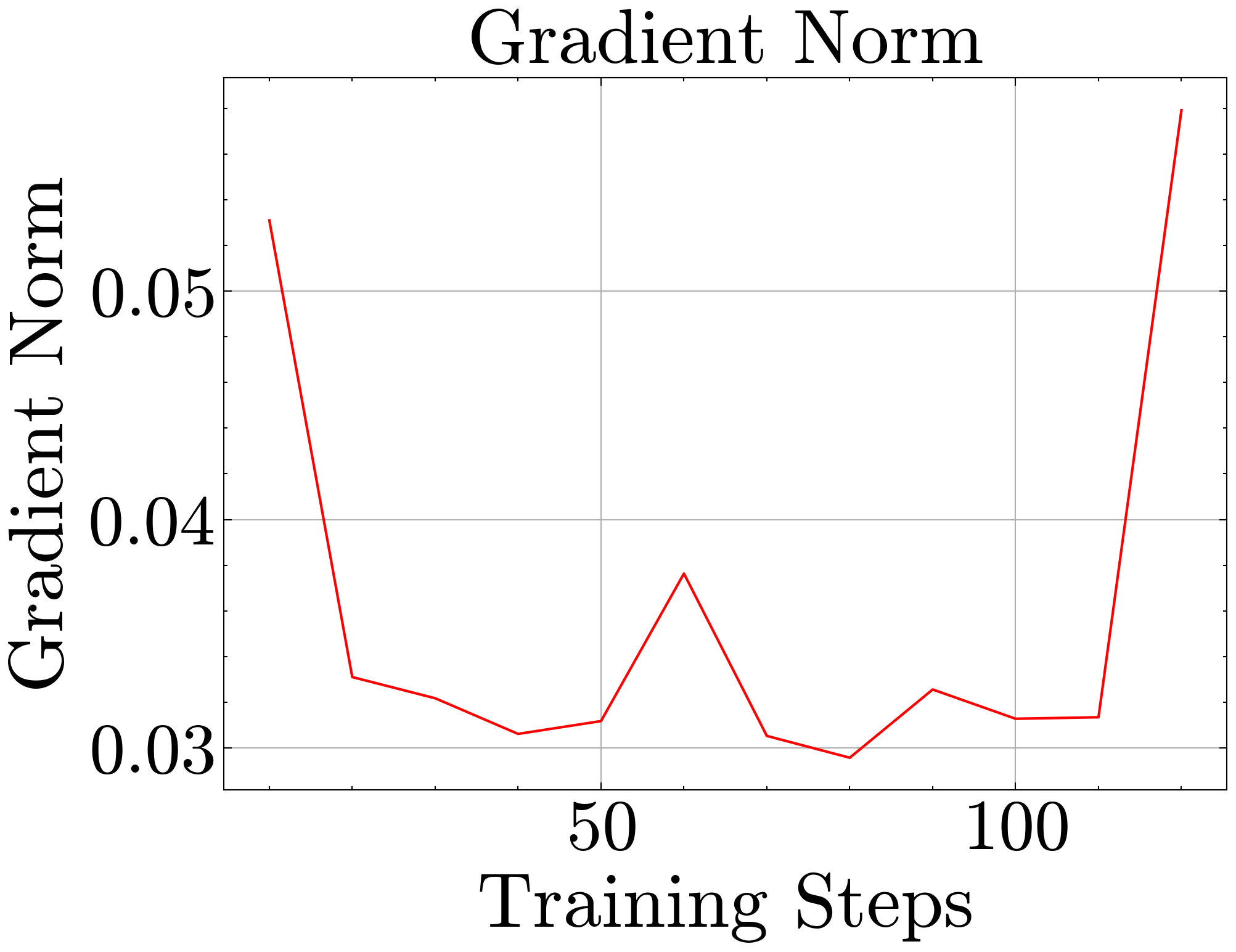}
    \end{minipage}\vspace{-0.5em}
    \caption{\textbf{RM dynamics of the 72B model.} Loss decreases more gradually with limited training steps, indicating early-stage adaptation to reasoning supervision. The gradient norm remains low overall with occasional spikes, suggesting stable but less extensive optimization.}
    \label{fig:SFT72B}
\end{figure*}

\parahead{Incongruity Modeling (IM)} 
The goal of IM is to adapt the backbone model to humor-relevant discourse. We train Qwen2.5-VL-7B-Instruct \citep{qwen2025qwen25technicalreport} using LoRA adapters (rank 32, scaling factor 64, dropout 0.05), and Qwen2.5-VL-32B-Instruct using LoRA with 4-bit quantization (rank 8, scaling factor 16), on the curated corpus described in Sec.~\ref{sec:cpt-corpora}. Training the 7B model runs for 50 epochs with batch size 1 and learning rate $1\times 10^{-4}$ on 2 NVIDIA H100 GPU ($\sim$4 hours). The 32B model is trained for 25 epochs with batch size 1 and learning rate $1\times 10^{-4}$ on 4 NVIDIA H100 GPUs ($\sim$6 hours). Similarly, the 72B model is trained for 10 epochs with batch size 1 and learning rate $1\times 10^{-4}$ on 4 NVIDIA H200 GPUs ($\sim$6 hours). Since the IM corpus is predominantly textual, we freeze the visual encoder and multimodal projection layers, updating only the language component. This prevents degradation of visual understanding while adapting the model to humor-specific reasoning patterns. Fig.~\ref{fig:CPT}-\ref{fig:CPT72B} show loss, learning rate, and gradient norm dynamics of our IM models.

\parahead{Resolution Modeling (RM)}
RM teaches the model to construct structured interpretations using captionist reasoning traces (Sec.~\ref{ssec:SFT}). We initialize from the IM-adapted model and train using \texttt{LLaMA-Factory} \citep{zheng2024llamafactory} with LoRA (rank 64 for 7B, rank 8 for 32B and 72B), keeping the vision encoder frozen. Training uses distilled reasoning traces and runs for 7 epochs with effective batch size 16, cosine learning-rate schedule with 10\% warmup, and peak rate $1\times 10^{-4}$. The 7B model trains in $\sim$3 hours on 2×H100 GPUs, while the 32B model requires $\sim$6 hours on 4×H100 GPUs. Fig.~\ref{fig:SFT}-\ref{fig:SFT72B} show training dynamics. Loss decreases smoothly after warmup, with stable gradients, indicating effective learning of structured reasoning patterns.

\begin{figure*}[!h]
\vspace{-0.1cm}
    \centering
    \begin{minipage}{0.32\textwidth}
        \centering
        \includegraphics[width=\linewidth]{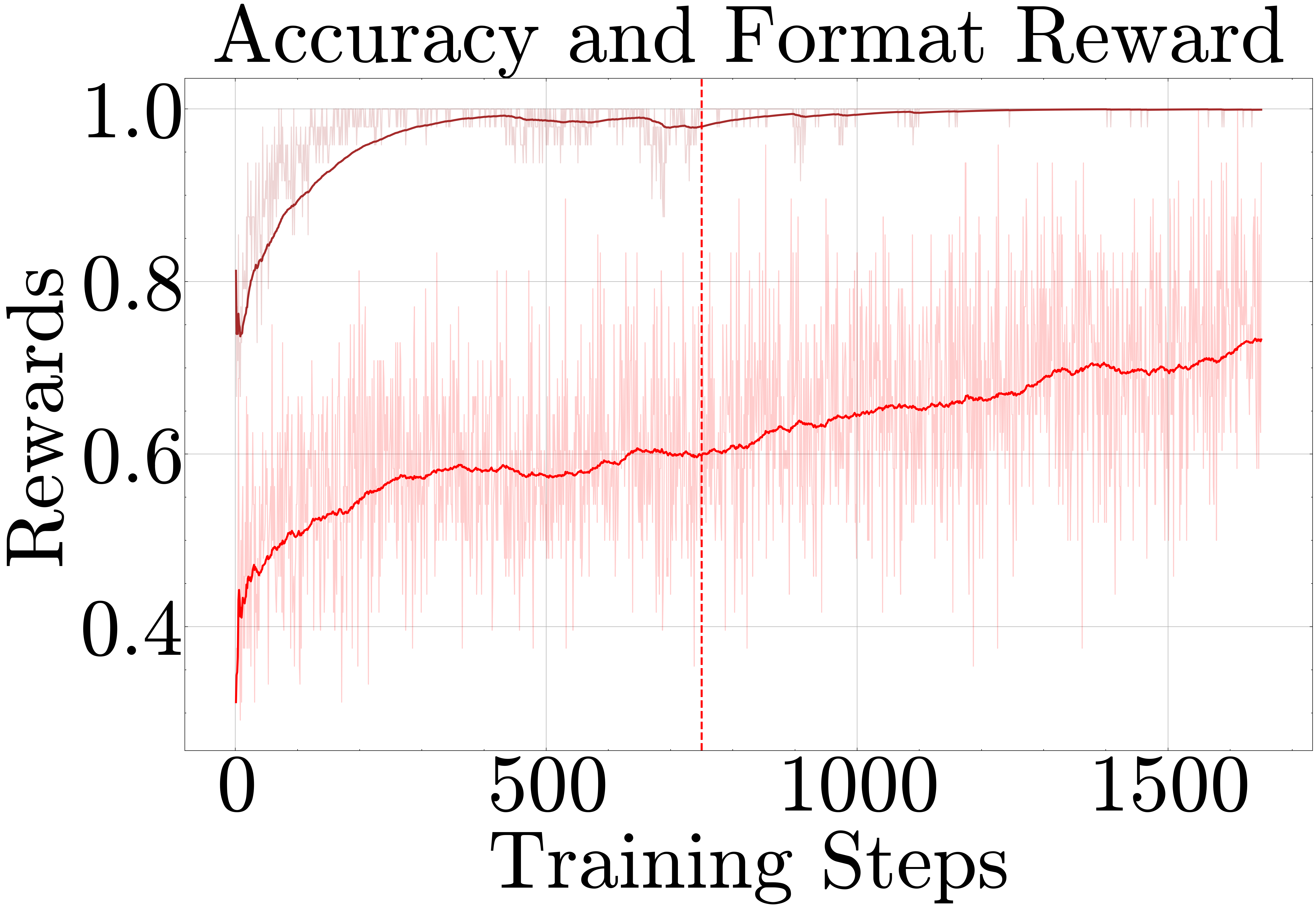}
    \end{minipage}
    \hfill
    \begin{minipage}{0.32\textwidth}
        \centering
        \includegraphics[width=\linewidth]{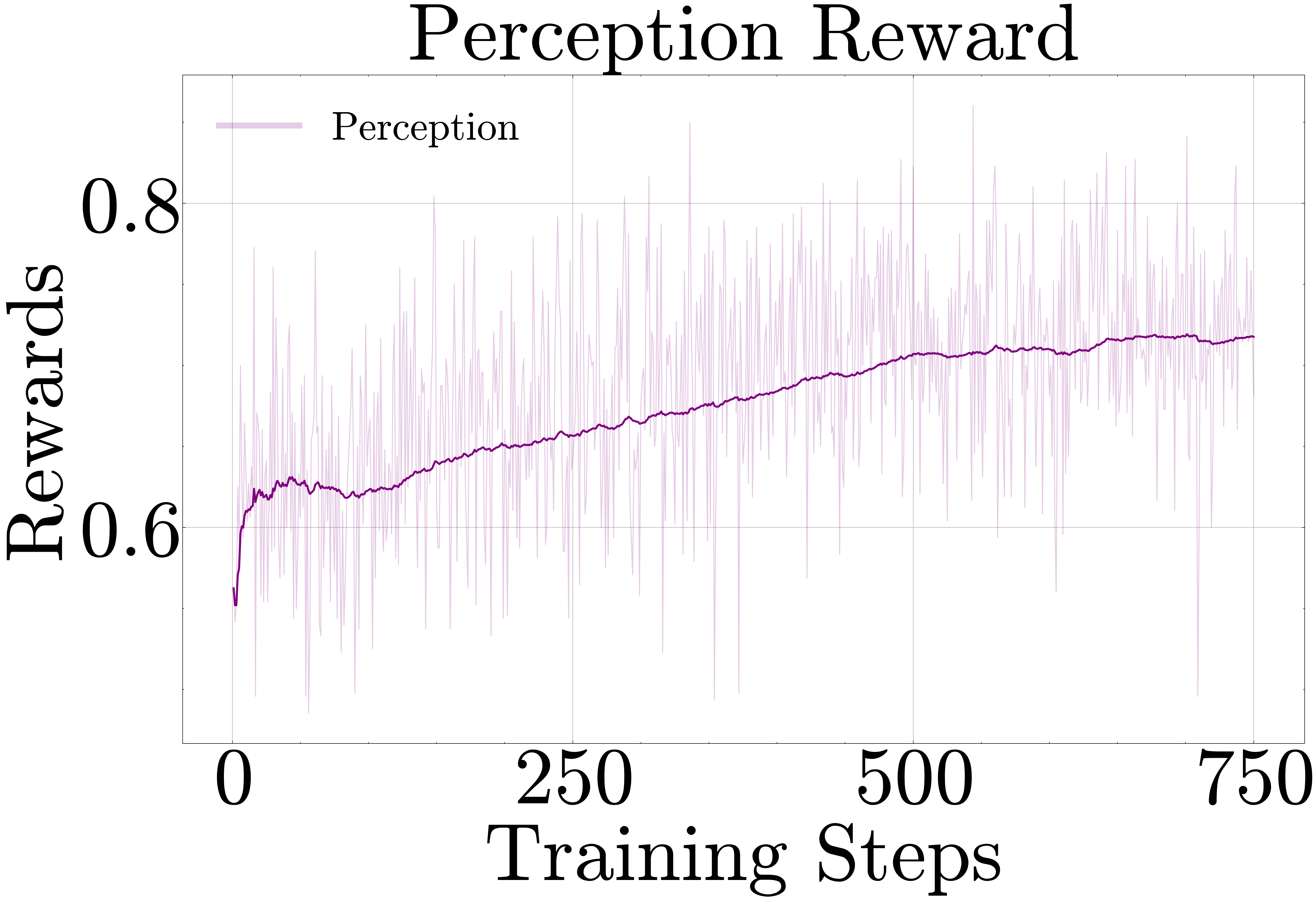}
    \end{minipage}
    \hfill
    \begin{minipage}{0.32\textwidth}
        \centering
        \includegraphics[width=\linewidth]{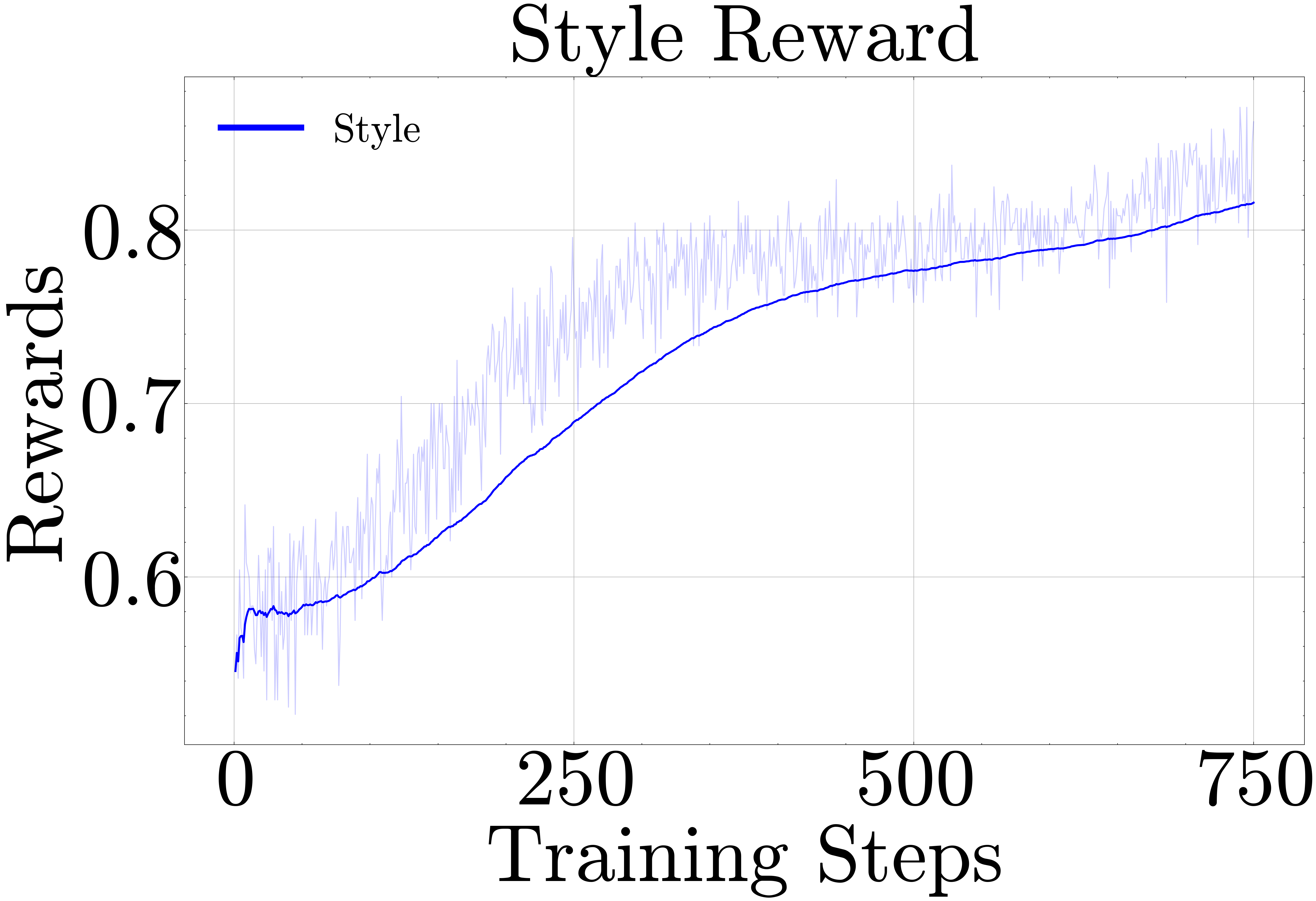}
    \end{minipage}\vspace{-0.5em}
    \caption{\textbf{PA dynamics of the 7B model.} Perception and style rewards increase rapidly and saturate early, while accuracy and format rewards continue improving, indicating progressive alignment with human judgment and reasoning quality.}
    \label{fig:RL}
\end{figure*}
\begin{figure*}[!h]
\vspace{-0.2cm}
    \centering
    \begin{minipage}{0.32\textwidth}
        \centering
        \includegraphics[width=\linewidth]{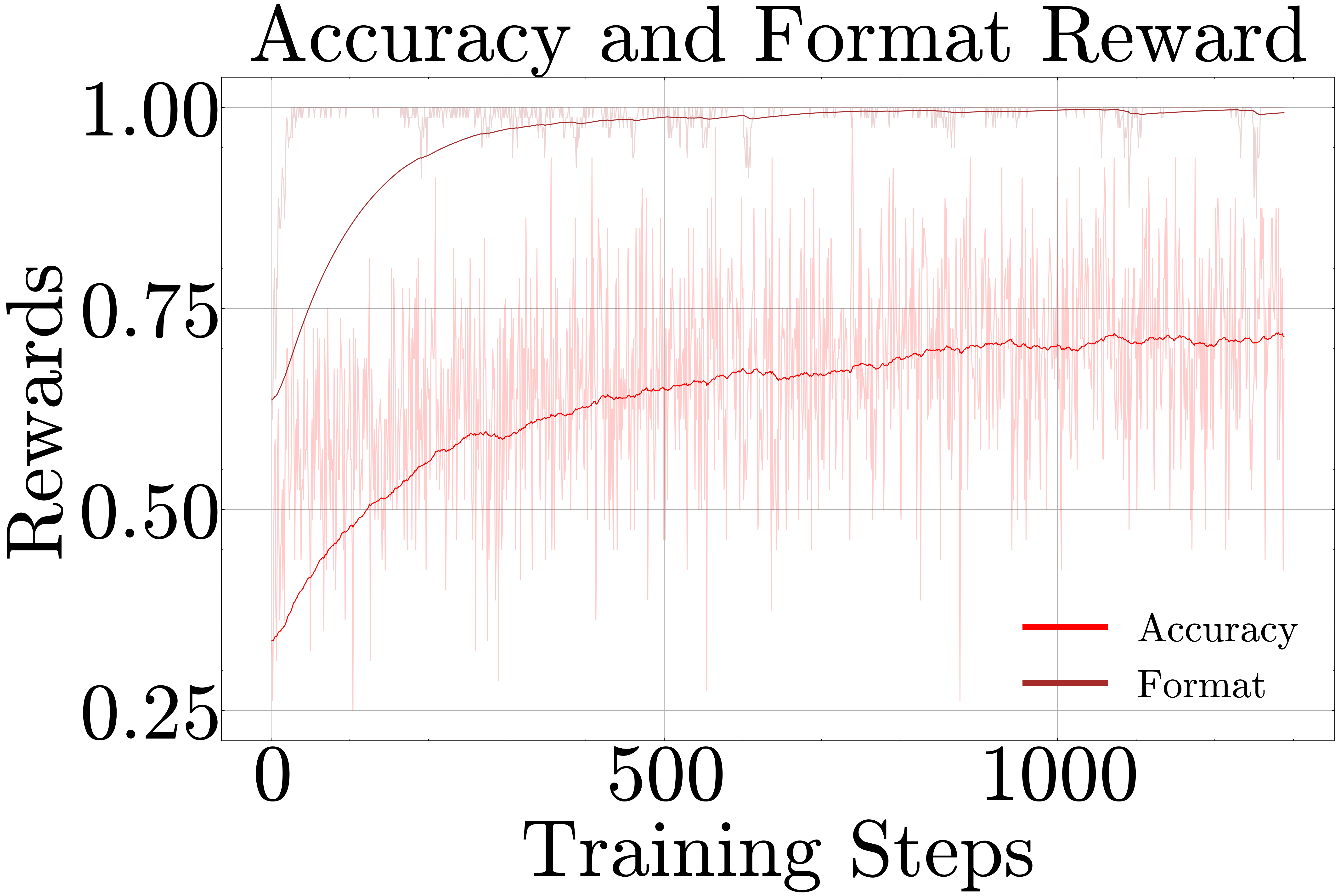}
    \end{minipage}
    \hfill
    \begin{minipage}{0.32\textwidth}
        \centering
        \includegraphics[width=\linewidth]{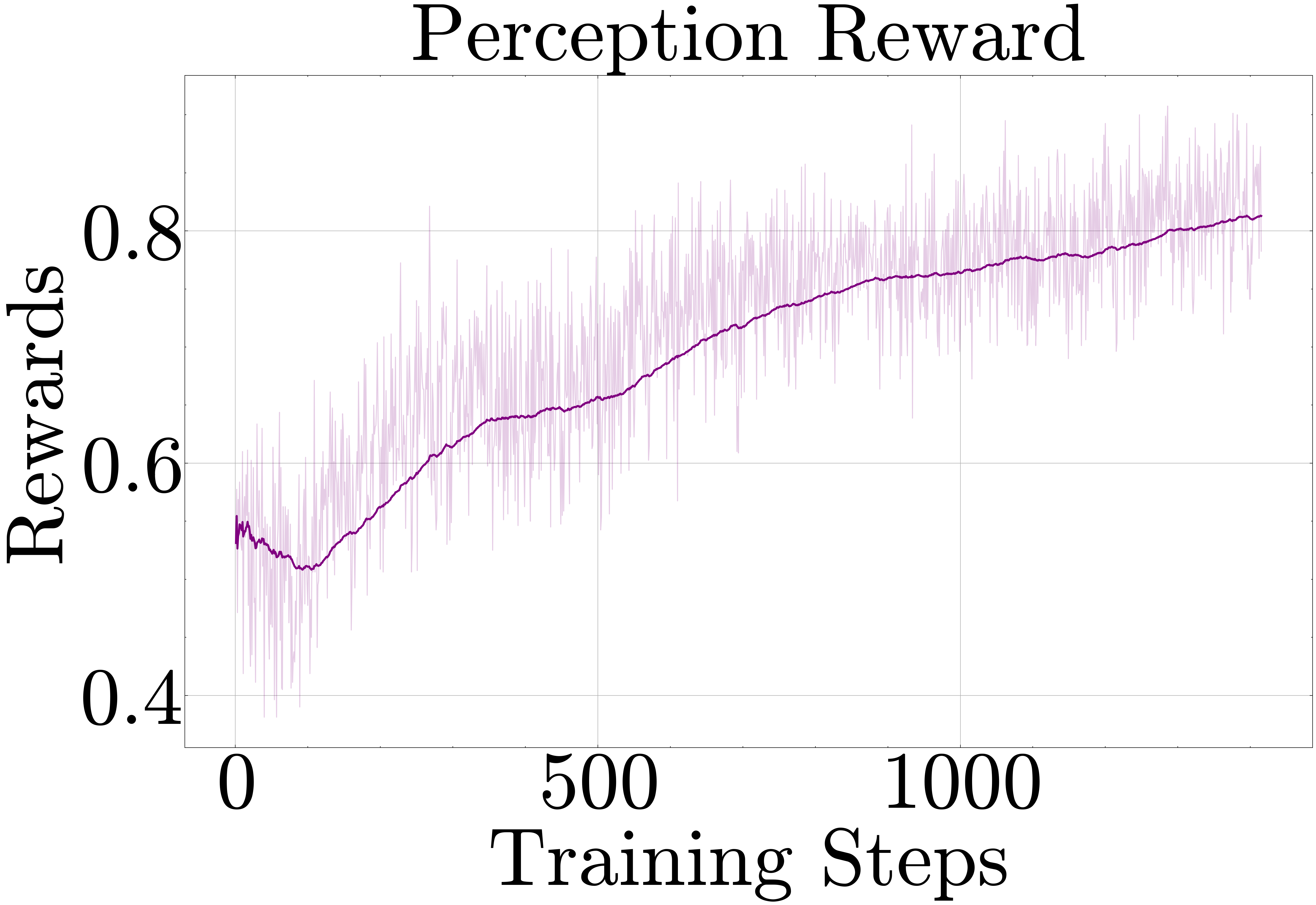}
    \end{minipage}
    \hfill
    \begin{minipage}{0.32\textwidth}
        \centering
        \includegraphics[width=\linewidth]{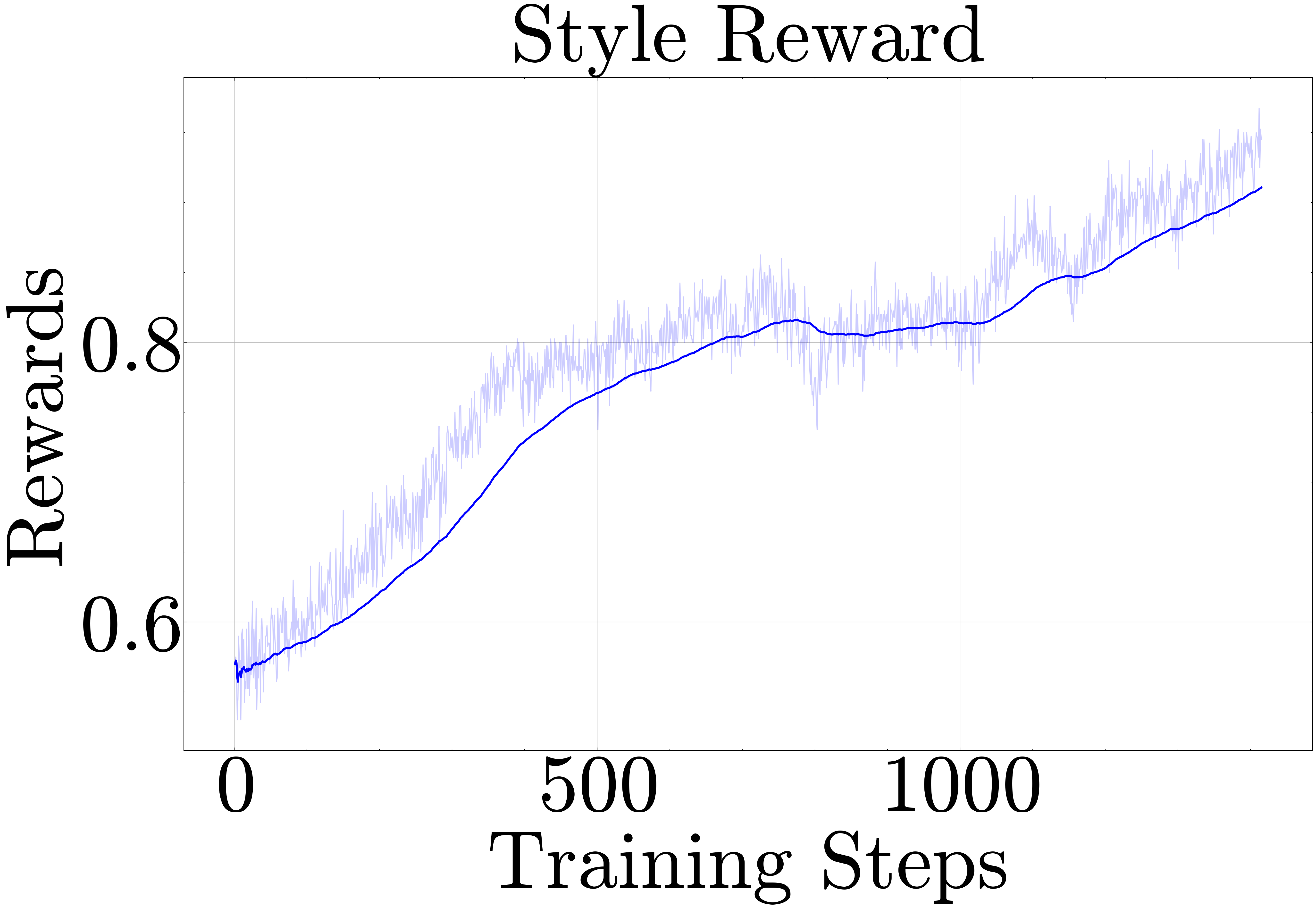}
    \end{minipage}\vspace{-0.5em}
    \caption{\textbf{PA dynamics of the 32B model.} Perception and style rewards increase steadily over training, while accuracy and format rewards improve more gradually, indicating stable and sustained alignment with human judgment and reasoning quality.}
    \label{fig:RL32B}
\end{figure*}

\begin{figure*}[!h]
\vspace{-0.2cm}
    \centering
    \begin{minipage}{0.32\textwidth}
        \centering
        \includegraphics[width=\linewidth]{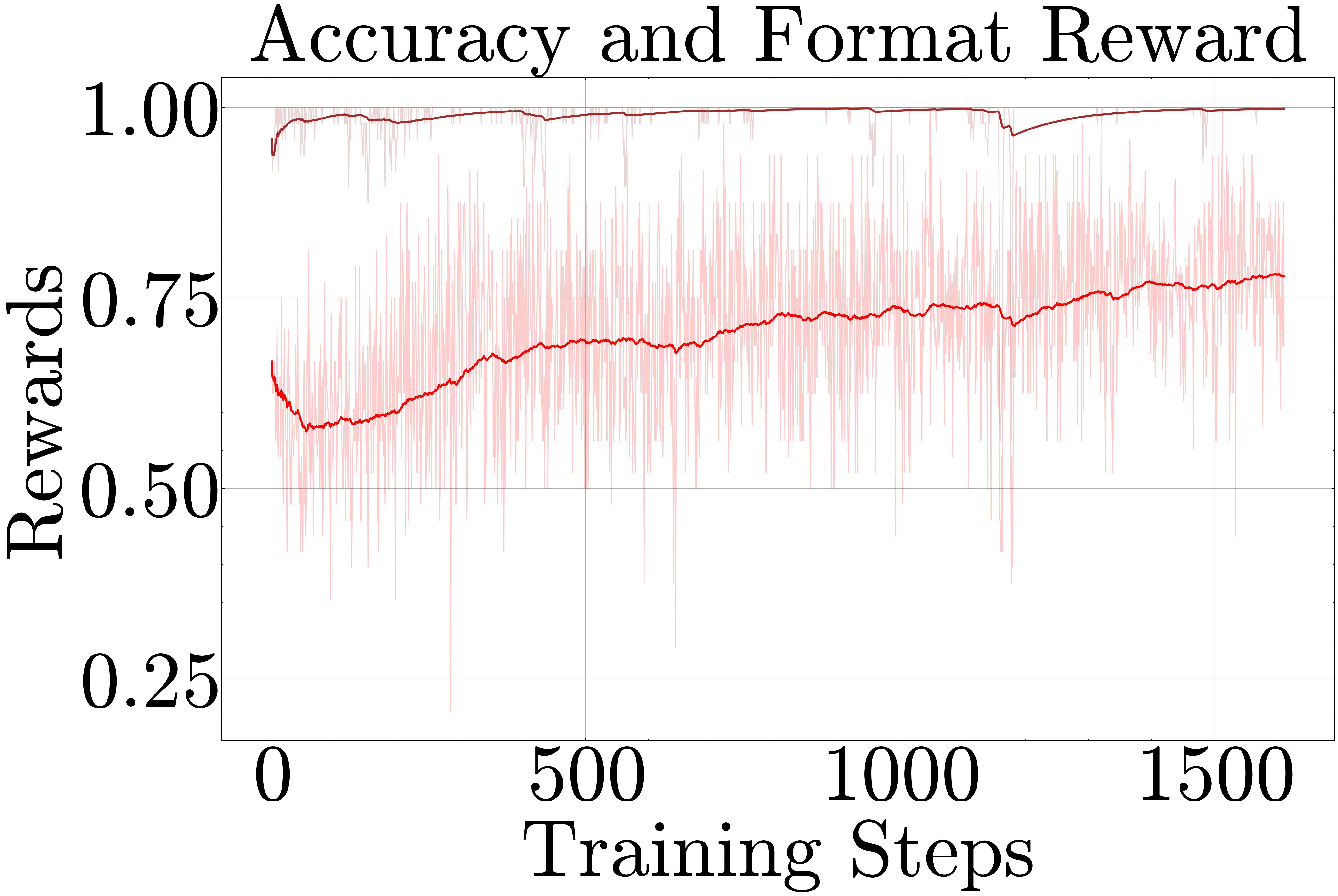}
    \end{minipage}
    \hfill
    \begin{minipage}{0.32\textwidth}
        \centering
        \includegraphics[width=\linewidth]{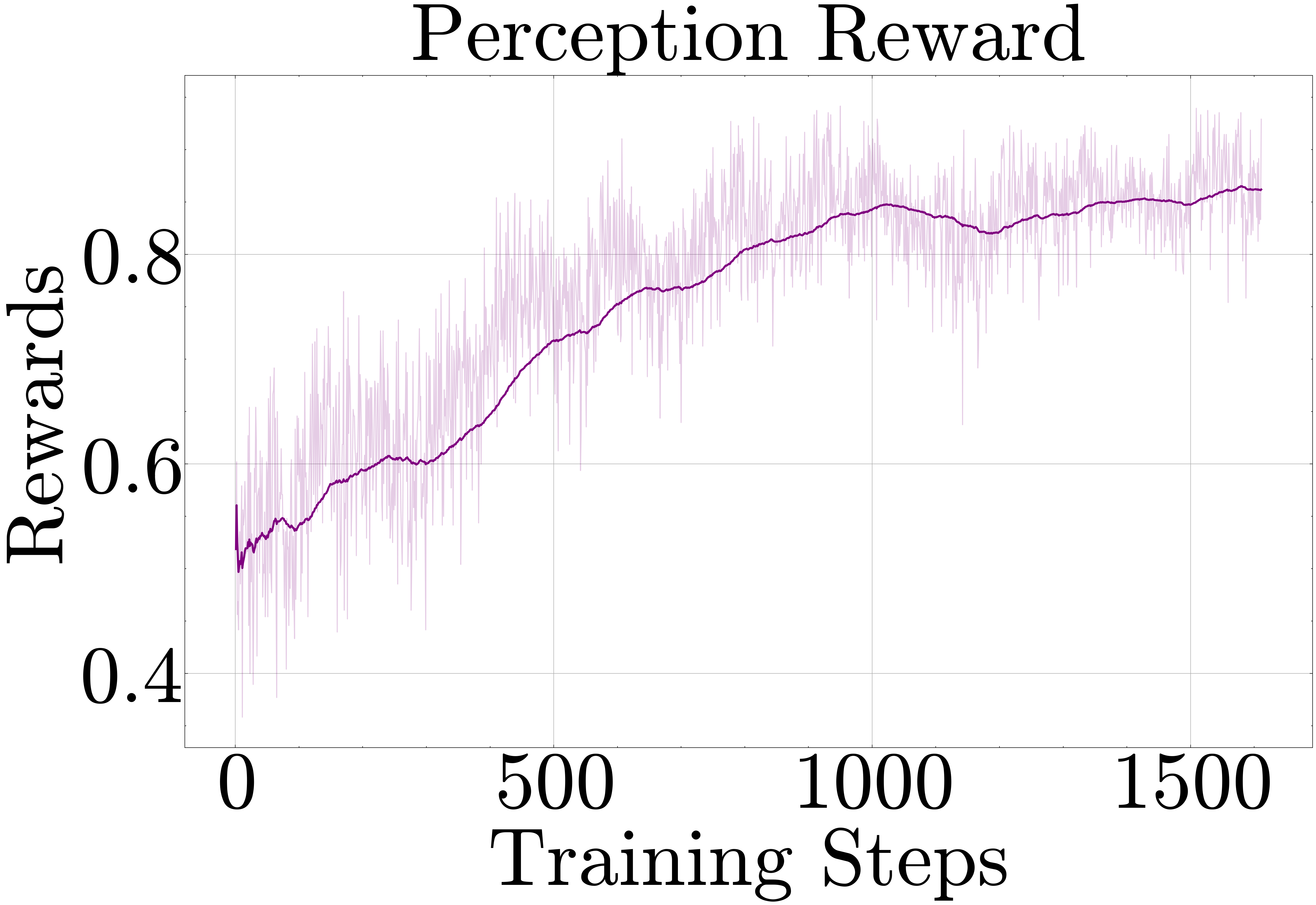}
    \end{minipage}
    \hfill
    \begin{minipage}{0.32\textwidth}
        \centering
        \includegraphics[width=\linewidth]{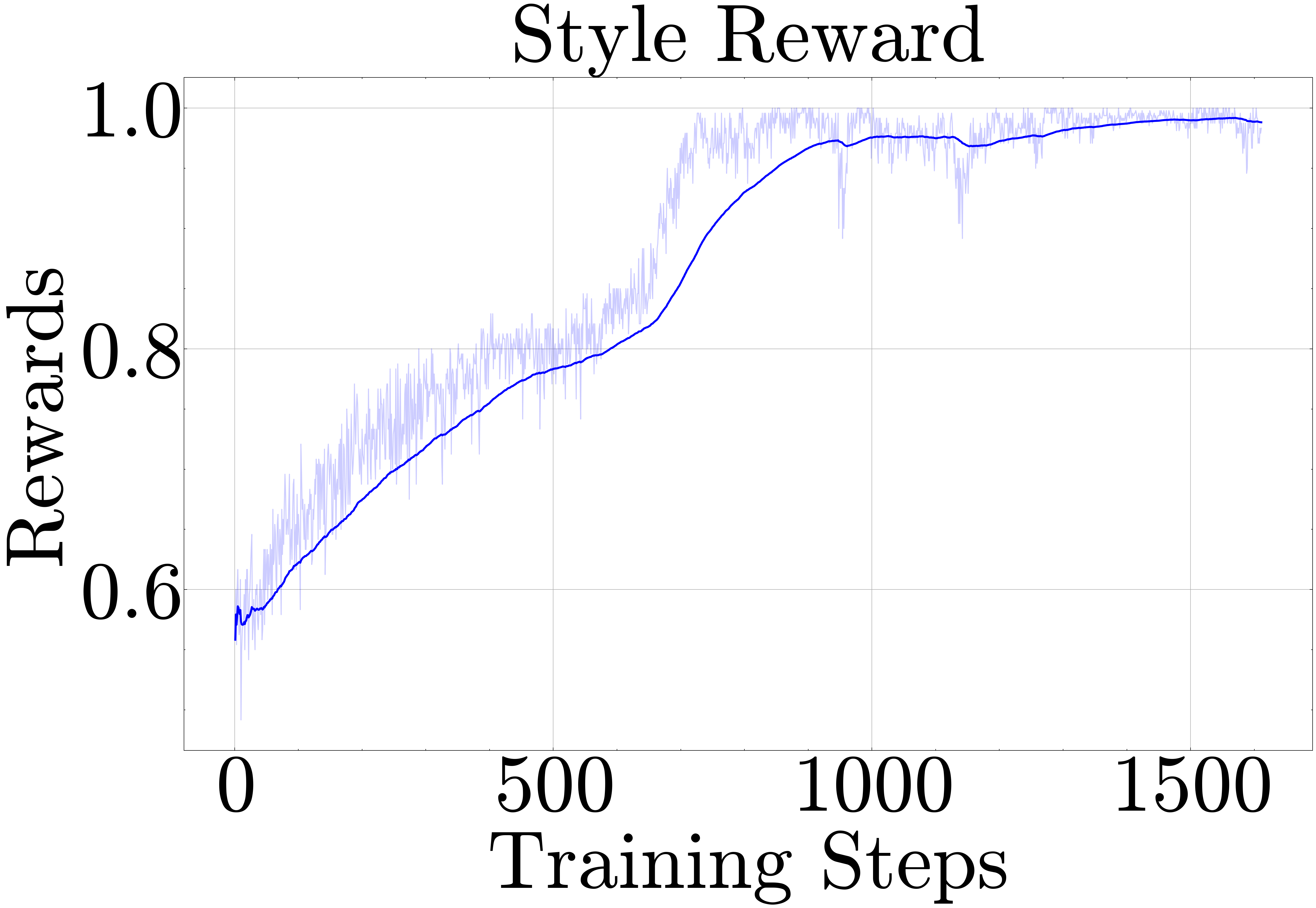}
    \end{minipage}\vspace{-0.5em}
    \caption{\textbf{PA dynamics of the 72B model.} Perception and style rewards increase rapidly and approach saturation, while accuracy and format rewards continue improving, indicating strong alignment with human judgment and well-calibrated reasoning quality at scale.}
    \label{fig:RL72B}
\end{figure*}

\parahead{Preference Alignment (PA)}
PA optimizes model outputs with humor-aware rewards (Sec.~\ref{ssec:RL}). We use EasyR1~\citep{zheng2025easyr1} with GRPO optimization~\citep{deepseekai2025deepseekr1incentivizingreasoningcapability}, sampling 3 rollouts per prompt with maximum length 512 and temperature 1.0. Training uses rollout batch size 16 and learning rate $1\times 10^{-6}$ on 4×H100 GPUs (8 for the 32B model, and 16 for 72B model). Fig.~\ref{fig:RL}-\ref{fig:RL72B} show reward dynamics. Perception and style rewards saturate early, while accuracy and format rewards continue improving.

\section{Prompts}
\setcounter{figure}{0}
\setcounter{table}{0}
\label{sec:prompts}
This section provides the exact prompts used to implement different components of \textit{Incongruity-Resolution Supervision (IRS)}. For reproducibility, we group them into three categories: (i) reasoning-trace generation, which supports structured interpretation; (ii) reward-judge prompts, which guide preference alignment; and (iii) evaluation prompts for both text-only and multimodal models.

\renewcommand{\bottomfraction}{0.8}

\renewcommand{\textfraction}{0.1}
\setlength{\textfloatsep}{10pt plus 1pt minus 2pt} 

\begin{figure*}[!h]
    \centering
    \includegraphics[width=\linewidth]{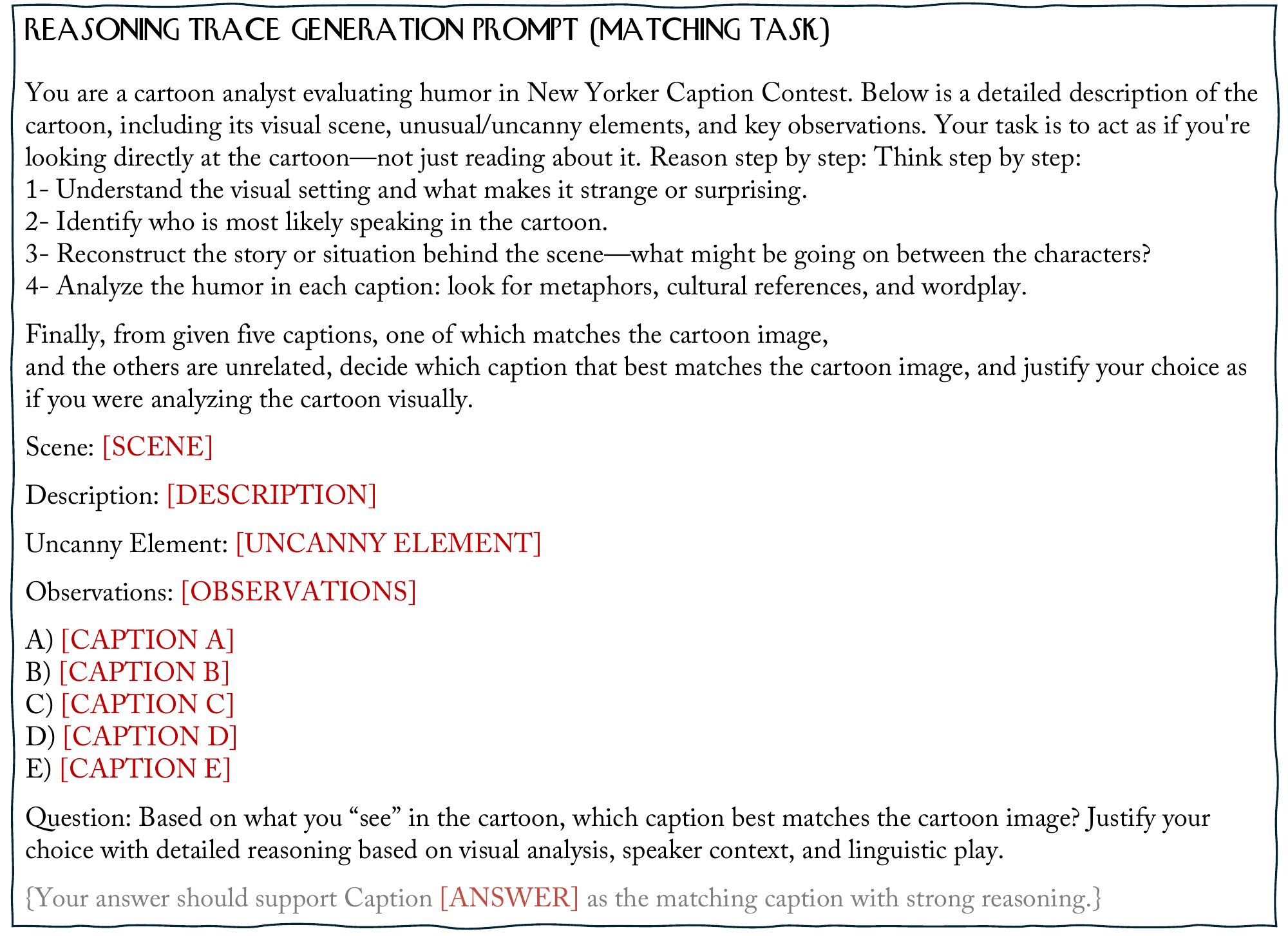}\vspace{-0.5em}
    \caption{\textbf{Prompt for reasoning trace generation (matching).} Enforces structured, step-by-step reasoning that reconstructs the scene, identifies incongruities, infers speaker context, and justifies the correct caption through visual grounding and narrative interpretation. The correct answer is optionally provided.}    
    \label{fig:CoT-generation-matching}
\end{figure*}

\subsection{Reasoning Trace Generation.}
\label{ssec:trace-generatio}
We generate captionist reasoning traces from the structured cartoon annotations of \citet{hessel-etal-2023-androids}, which include scene descriptions, uncanny elements, and entity references that provide grounding for interpretation.\footnote{For cartoons missing these descriptions, we use the OpenAI o3 model~\citep{openai_o3_system_card_2025} to generate them automatically from the image.} We use DeepSeek-R1~\citep{deepseekai2025deepseekr1incentivizingreasoningcapability} as a teacher model, prompting it to reconstruct the scene, identify salient incongruities, infer speaker intent, and analyze humor through mechanisms such as wordplay and cultural references before selecting the correct caption. This process produces structured reasoning traces that explicitly model how captions are interpreted and evaluated.

The resulting traces are rephrased using GPT-4o~\citep{openai2024gpt4o} to match the style of professional captionist commentary, emphasizing concise, observational, and image-grounded language. This transformation replaces annotation-based phrasing (e.g., \textit{``the description says''}) with direct visual interpretation (e.g., \textit{``when I look at the cartoon''}), while preserving the underlying reasoning structure. This step ensures stylistic consistency between generated traces and real captionist discourse.

\begin{figure*}[!h]
    \centering
    \includegraphics[width=\linewidth]{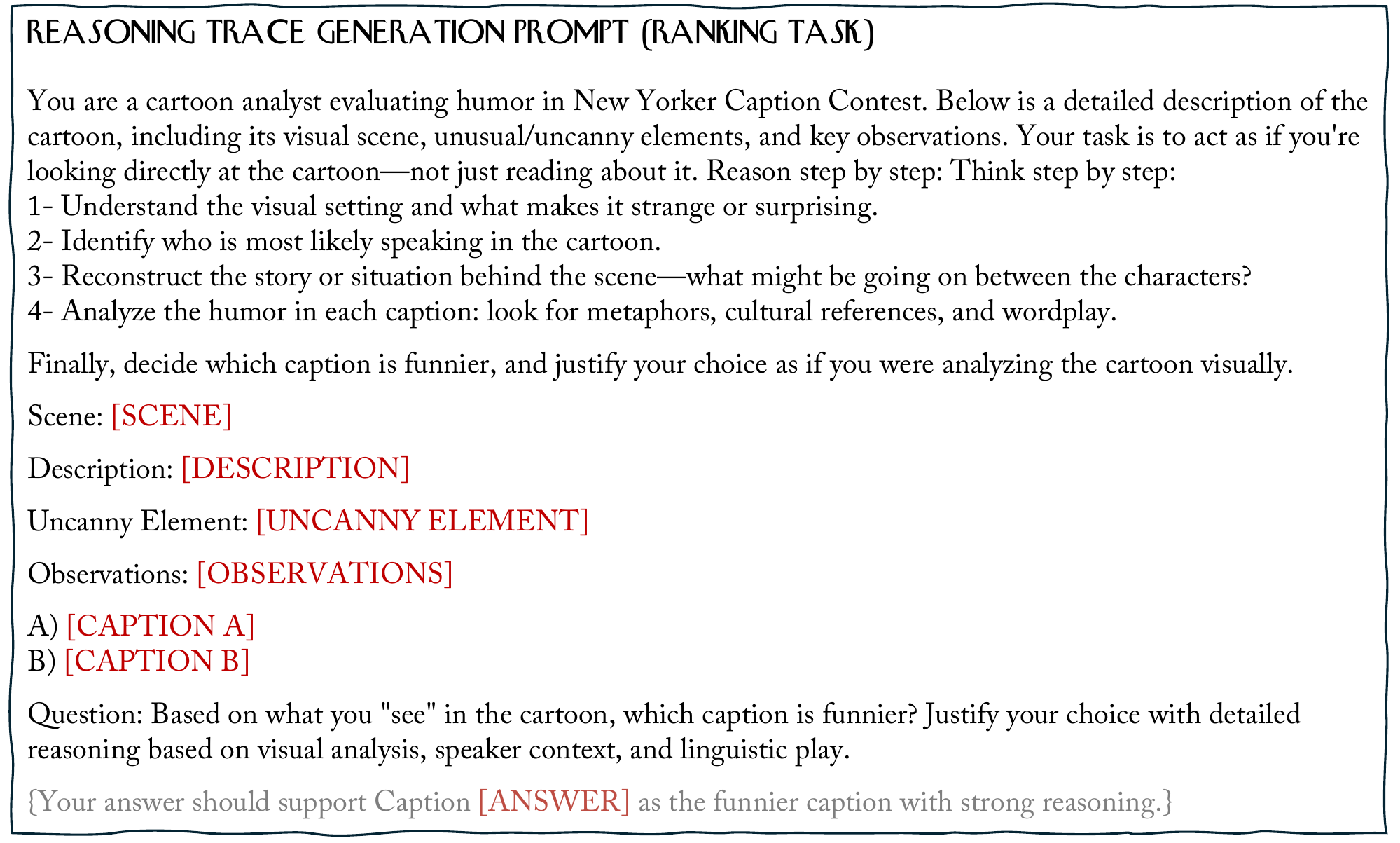}\vspace{-0.5em}
    \caption{\textbf{Prompt for reasoning trace generation (ranking).} Enforces structured, step-by-step reasoning that reconstructs the scene, identifies incongruities, infers speaker context, and compares competing captions through visual grounding and narrative interpretation. The correct answer is optionally provided.}
    \label{fig:CoT-generation-ranking}
\end{figure*}

\begin{figure*}[!h]
    \centering
    \includegraphics[width=\linewidth]{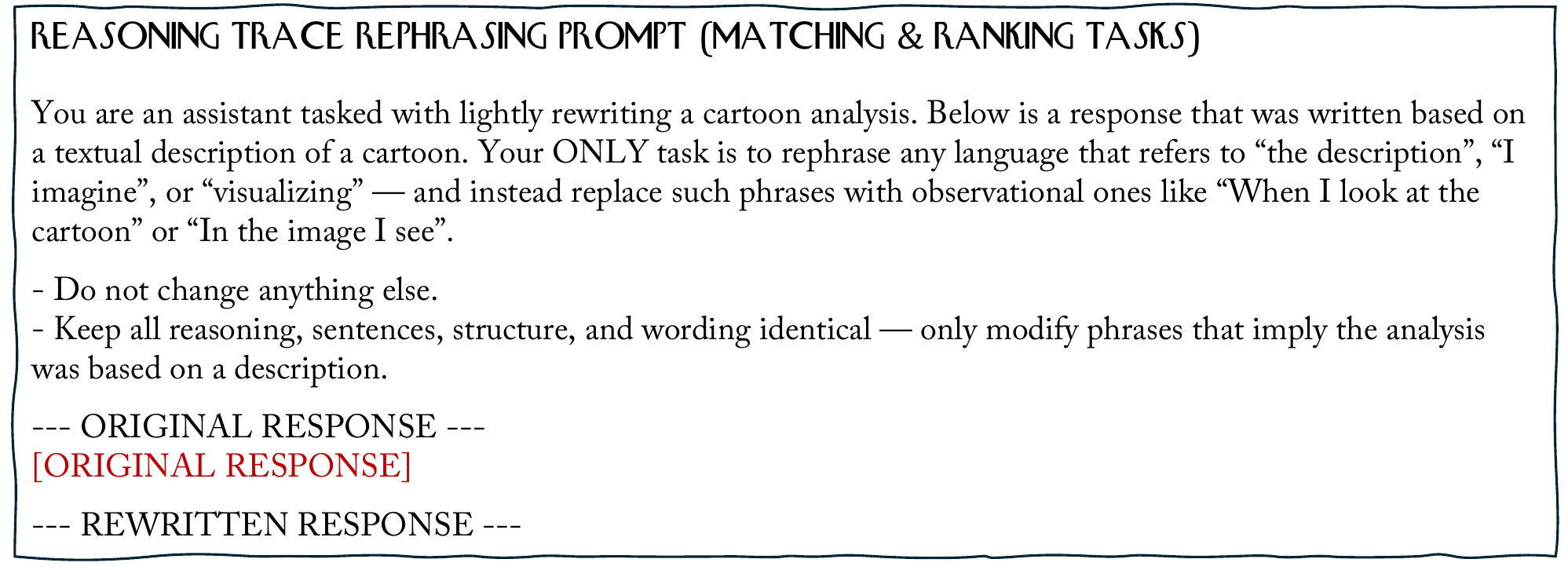}\vspace{-0.5em}
    \caption{\textbf{Prompt for reasoning-trace rephrasing.} Rewrites reasoning traces into captionist-style commentary by replacing description-based phrasing with image-grounded language, while preserving the original reasoning structure and content.}
    \label{fig:CoT-rephrasing}
\end{figure*}

Fig.~\ref{fig:CoT-generation-matching} and~\ref{fig:CoT-generation-ranking} shows the generation templates for the matching and ranking tasks, respectively, which structures the reasoning process around scene description, incongruity detection, and caption justification. Fig.~\ref{fig:CoT-rephrasing} shows the rephrasing template, used to refine traces into captionist-style commentary.

\subsection{Reward-Judge Prompts}
\label{ssec:judge-reward-prompts}
We design two LLM-as-judge templates that produce automatic, fine-grained reward signals during reinforcement learning. These prompts implement \emph{preference alignment} within IRS by evaluating whether model outputs are both visually grounded and stylistically consistent with captionist practice. Each judge returns binary scores for multiple criteria, which are aggregated into the reward function (Sec.~\ref{ssec:RL}). Together, these rewards capture two core dimensions of humor understanding: (i) perceptual grounding, encouraging reasoning to reference concrete visual incongruities in the cartoon, and (ii) stylistic fidelity, encouraging captions and explanations to reflect the linguistic qualities of professional humor writing.

\begin{figure*}[!t]
    \centering
    \includegraphics[width=\linewidth]{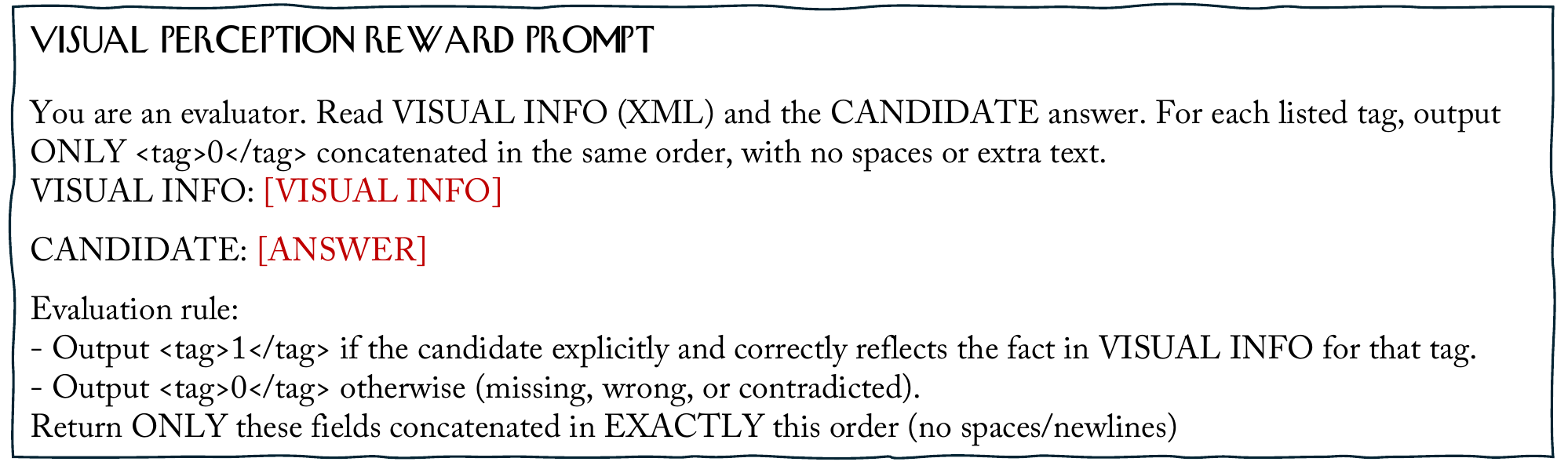}\vspace{-0.5em}
    \caption{\textbf{Prompt for visual perception judge.} Evaluates whether reasoning trace explicitly references curated visual details; outputs binary tags for each attribute.}
    \label{fig:prompt_perception}
\end{figure*}

\parahead{Visual Perception Judge}  
This prompt evaluates whether model reasoning is grounded in salient visual details. Each cartoon is associated with up to ten curated reference descriptions (entities, background elements, and incongruities). Qwen2.5-7B-Instruct \citep{qwen25technicalreport_pre} receives both the model’s reasoning and these references, and outputs a binary vector indicating whether each reference is explicitly reflected in the reasoning. This design directly links the perception reward ($R_p$) to the visual anchors curated in our dataset (Fig.~\ref{fig:visual_references}), encouraging explanations that are faithful to the observed scene rather than relying on abstract or unsupported interpretations. Fig.~\ref{fig:prompt_perception} shows the exact prompt.

\begin{figure*}[!t]
    \centering
    \includegraphics[width=0.99\linewidth]{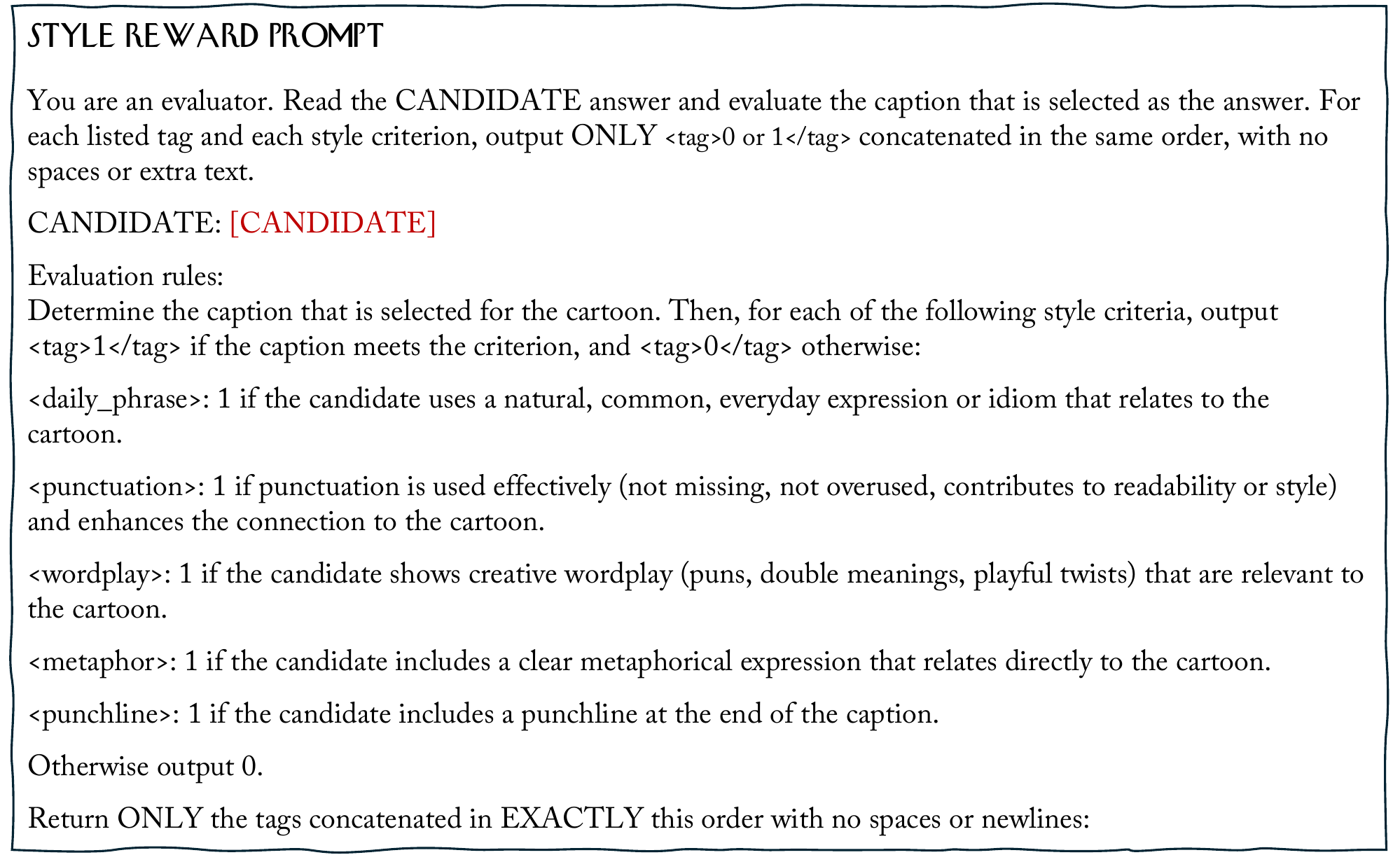} 
    \caption{\textbf{Prompt for style judge.} Evaluates captions against stylistic criteria such as phrasing, punctuation, wordplay, metaphor, and punchline placement; outputs binary tags for each dimension.}
    \label{fig:prompt_style}
\end{figure*}

\begin{figure*}[!t]
    \centering
    \includegraphics[width=\linewidth]{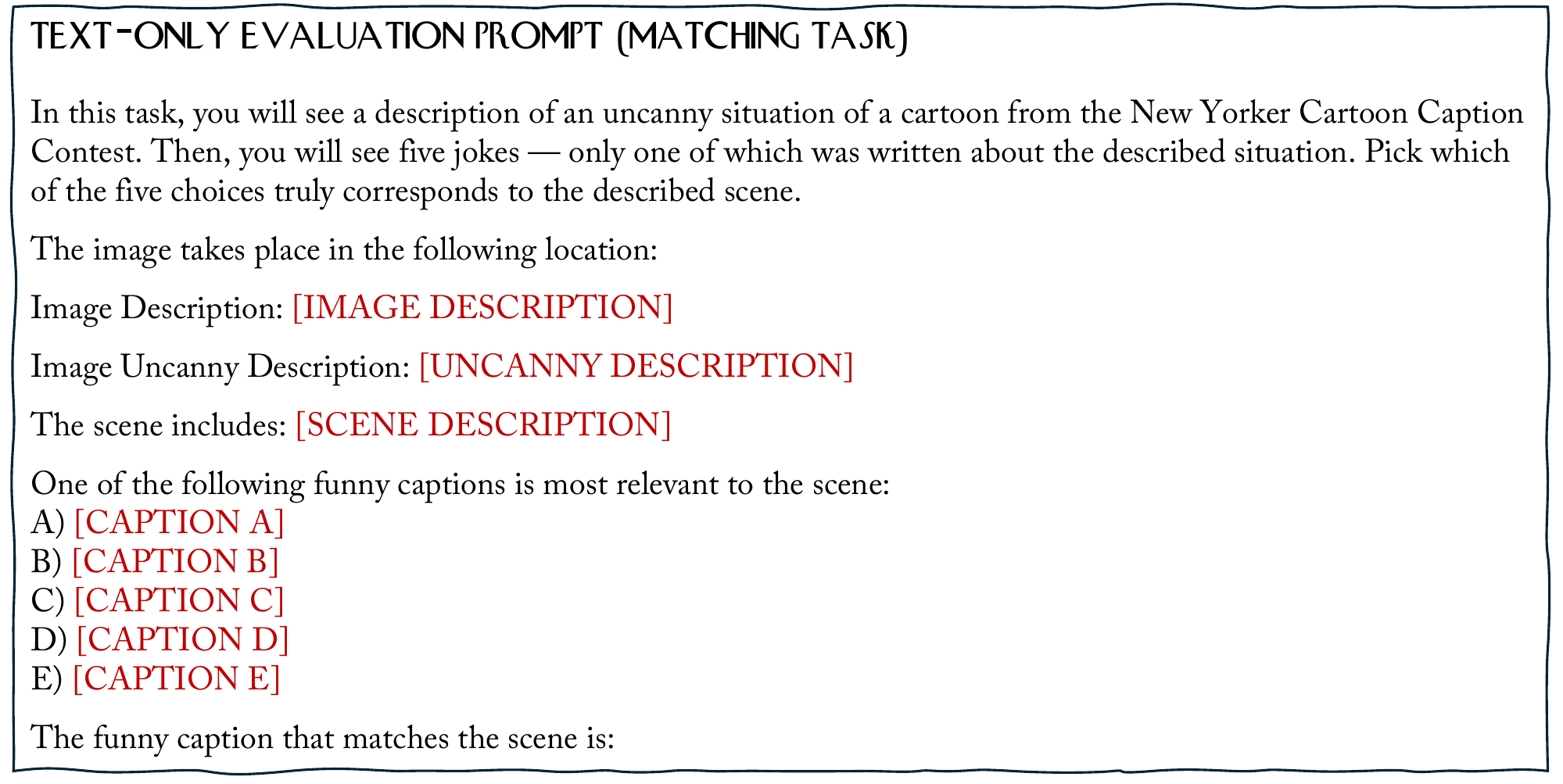} \vspace{-2em}
    \caption{\textbf{Prompt for text-only evaluation (matching).} Evaluates DeepSeek-R1 using textual annotations of the cartoon; model selects the caption that best matches the described scene.}
    \label{fig:prompt_textonly_matching}
\end{figure*}
\begin{figure*}[!t]
    \centering
    \includegraphics[width=\linewidth]{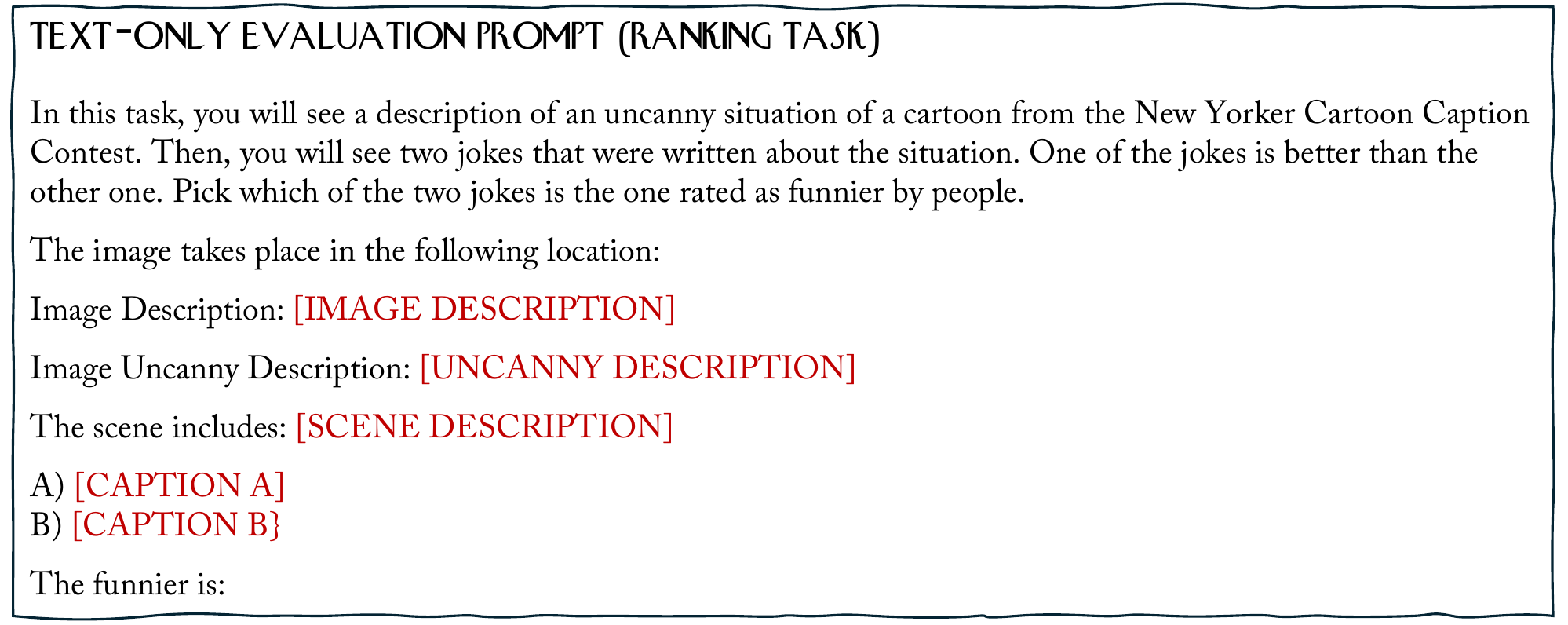} \vspace{-2em}
    \caption{\textbf{Prompt for text-only evaluation (ranking).} Evaluates DeepSeek-R1 using textual annotations; model selects the funnier caption between two options.}
    \label{fig:prompt_textonly_ranking}
\end{figure*}

\parahead{Style Judge}  
This prompt evaluates the linguistic quality of captions and explanations. Drawing on captionist guidelines~\citep{wood2024your}, the template checks five stylistic dimensions:
\begin{enumerate}[leftmargin=*]
\item \textbf{Natural phrasing} (use of idiomatic, everyday expressions),
\item \textbf{Punctuation} (effective, neither missing nor overused),
\item \textbf{Wordplay} (puns, double meanings, playful twists),
\item \textbf{Metaphor} (figurative expressions grounded in the cartoon),
\item \textbf{Punchline placement} (delivering the payoff at the end).
\end{enumerate}

The judge outputs a binary vector corresponding to these criteria without additional commentary. The aggregated score forms the style reward ($R_s$), encouraging models not only to select the correct caption but to justify it in a way that reflects the tone and structure of professional humor writing. Fig.~\ref{fig:prompt_style} shows the full template.

Collectively, these LLM-as-judge rewards complement correctness- and format-based signals, enabling reinforcement learning to capture both perceptual grounding and stylistic quality. This enriches training with humor-aware evaluation criteria that go beyond task accuracy alone.

\subsection{Text-only Evaluation}
\label{ssec:prompt_textual}
We evaluate DeepSeek-R1~\citep{deepseekai2025deepseekr1incentivizingreasoningcapability} as a text-only baseline using structured cartoon annotations from \citet{hessel-etal-2023-androids}. These annotations provide scene descriptions, uncanny elements, and key observations, enabling reasoning without access to raw visual inputs. This setting isolates reasoning ability under near-perfect perception, since the model is given curated descriptions that abstract away visual ambiguity. We use the original prompts from \citet{hessel-etal-2023-androids} for both matching and ranking tasks. The matching prompt presents five candidate captions with one correct option, while the ranking prompt presents two captions with one preferred by the crowd. In both cases, the model is instructed to reason solely from the textual annotations. Fig.~\ref{fig:prompt_textonly_matching} and Fig.~\ref{fig:prompt_textonly_ranking} show the full templates.

\begin{figure*}[!h]
\centering
\includegraphics[width=\linewidth]{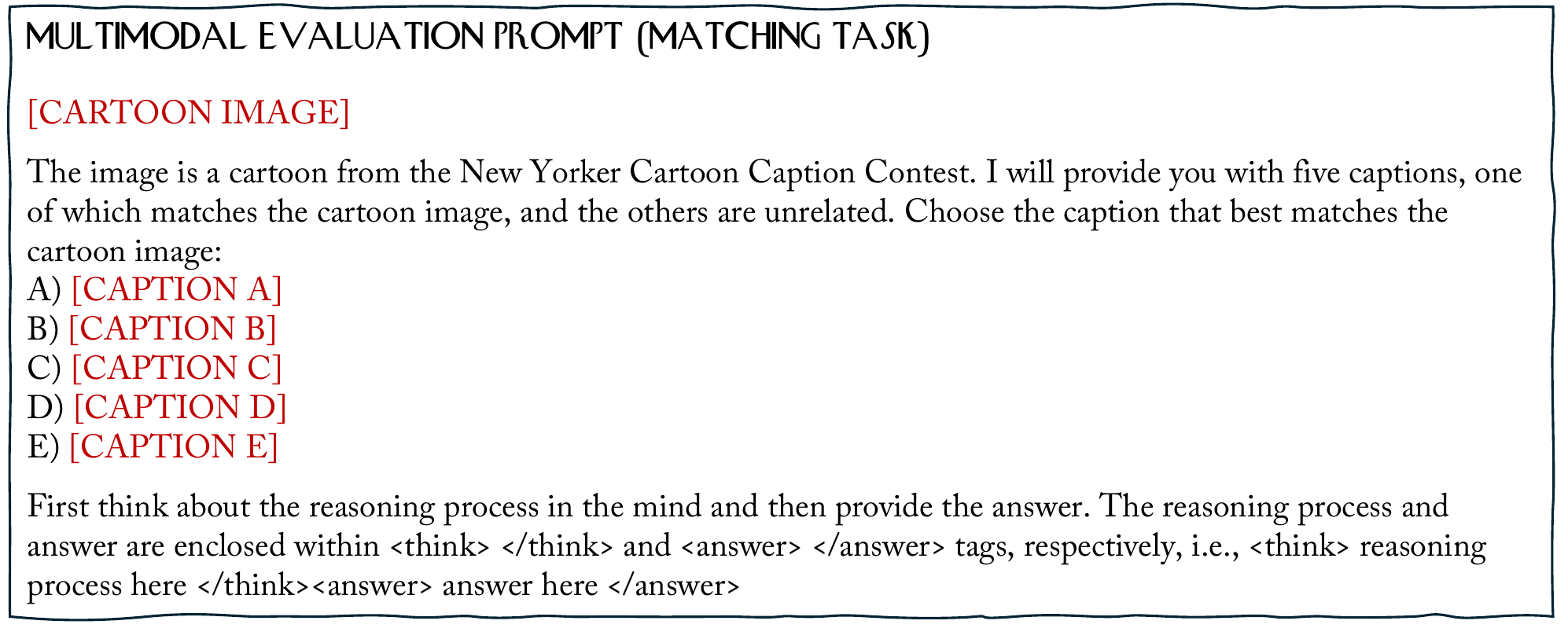}\vspace{-1em}
\caption{\textbf{Prompt for multimodal evaluation (matching).} Cartoon image plus five candidate captions (A-E). Model outputs reasoning and final choice in \texttt{<think>} and \texttt{<answer>} tags.}
\label{fig:prompt_mm_matching}
\end{figure*}
\begin{figure*}[!t]
    \centering
    \includegraphics[width=\linewidth]{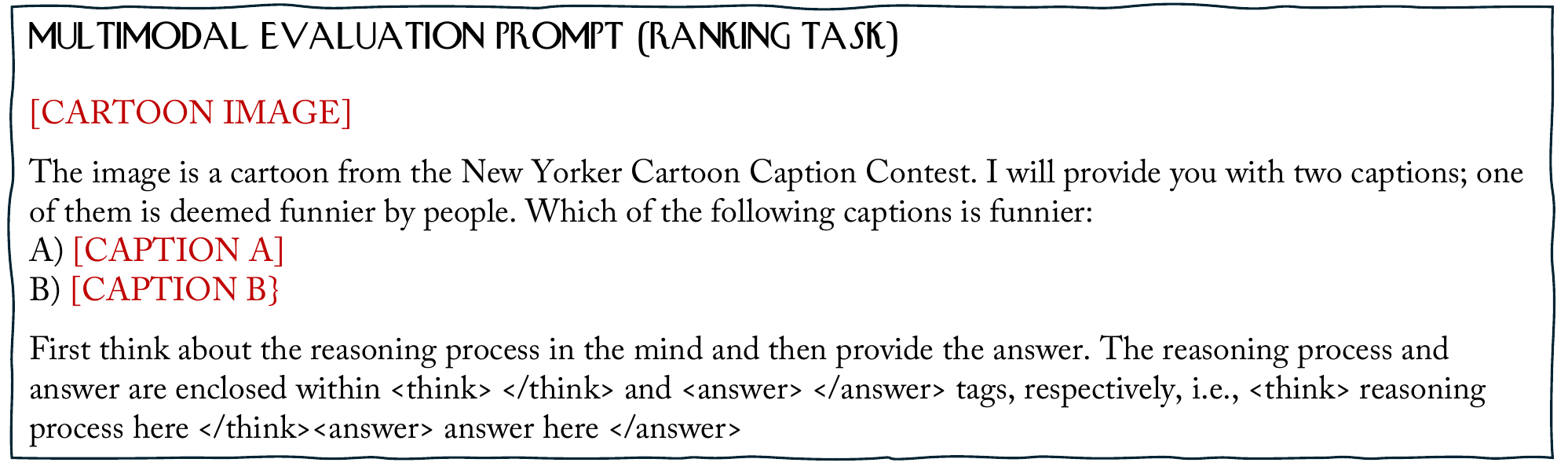}\vspace{-1em}
    \caption{\textbf{Prompt for multimodal evaluation (ranking).} Cartoon image plus two candidate captions (A-B). Model outputs reasoning and final choice in \texttt{<think>} and \texttt{<answer>} tags.}
\label{fig:prompt_mm_ranking}
    \label{fig:prompt_mm_ranking}
\end{figure*}
\subsection{Multimodal Evaluation}
\label{ssec:prompt-multimodal}
We standardize evaluation of vision-language models using task-specific prompts that present the cartoon image (\verb|<image>|) together with candidate captions. All models are required to produce outputs in the controlled format \verb|<think>...</think><answer>...</answer>|, which enforces explicit reasoning traces and enables automatic parsing and evaluation. Fig.~\ref{fig:prompt_mm_matching} and Fig.~\ref{fig:prompt_mm_ranking} show the full templates.

To ensure fair comparison, the same prompting protocol is applied to all competing multimodal reasoning models (e.g., GLM-4V, Qwen2.5-VL, Kimi-VL). Models are explicitly instructed to \emph{``think before answering''} and to follow the standardized output format. This helps ensure that performance differences reflect reasoning ability rather than prompt design.
\setlength{\tabcolsep}{5pt}
\begin{table*}[!t]
\renewcommand{\thetable}{G.1}
\centering
\caption{\textbf{Inference configurations.} Model identifiers, access methods, and inference configurations used for evaluation. API-accessed models use the providers' default inference settings.}
\label{tab:inference-config}
\begin{tabular}{llcl}
\toprule
\textbf{Model} & \textbf{Model Identifier} & \textbf{Access} & \textbf{Inference Configuration} \\
\midrule
GPT-4o & \texttt{gpt-4o} & OpenAI API & Default \\
o4-mini & \texttt{o4-mini} & OpenAI API & Default \\
o3 & \texttt{o3} & OpenAI API & Default \\
DeepSeek-R1 & \texttt{deepseek-reasoner} & DeepSeek API & Default \\
Gemini 2.5 Flash & \texttt{gemini-2.5-flash} & Gemini API & Default \\
\midrule
\begin{tabular}[t]{@{}l@{}}Qwen2.5-VL\\\{7B,32B,72B\}-Instruct\end{tabular} & \texttt{...} & Local & \begin{tabular}[t]{@{}l@{}}$T=0.5$,\\Max new tokens $=1024$\end{tabular} \\
GLM-4.1V-9B-Think & \texttt{...} & Local & \begin{tabular}[t]{@{}l@{}}$T=$Model Default,\\Max new tokens $=2048$\end{tabular}\\
Kimi-VL-A3B-Think & \texttt{...} & Local & \begin{tabular}[t]{@{}l@{}}$T=0.5$,\\Max new tokens $=1024$\end{tabular} \\
LlamaV-o1 & \texttt{...} & Local & \begin{tabular}[t]{@{}l@{}}$T=$Model Default,\\Max new tokens $=2048$\end{tabular} \\
NVLM-D-72B & \texttt{...} & Local & \begin{tabular}[t]{@{}l@{}}$T=0.5$,\\Max new tokens $=$ Model Default\end{tabular}\\
\bottomrule
\end{tabular}
\end{table*}

\section{Inference Configurations}
\label{sec:inference-conf}

Table~\ref{tab:inference-config} summarizes the models used for evaluation together with their access method and inference settings. We use the default inference settings for models accessed through APIs, while the decoding settings used for locally hosted models are reported in the table. These include temperature and maximum generation length where applicable. All reported results are obtained from a single evaluation run for each model.

\section{Statistical Analysis}
\label{sec:stat-analysis}

\begin{table*}[!t]
\renewcommand{\thetable}{H.1}
\centering
\caption{\textbf{Performance with 95\% bootstrap confidence intervals.} Accuracy and standard error across the in-domain evaluation splits.}
\begin{tabular}{llcccr}
\toprule
\textbf{Model} & \textbf{Task} & \textbf{$n$} &
\textbf{Accuracy (\%) [95\% CI]} & \textbf{Standard Error} \\
\midrule
\multirow{4}{*}{\textbf{IRS-7B}}
& Matching & 300 & 59.67 [54.00, 65.33] & 0.0288 \\
& Ranking & 385 & 64.42 [59.48, 69.09] & 0.0244 \\
& Ranking (10-vs-1000) & 350 & 56.29 [51.14, 61.43] & 0.0264 \\
& Ranking (30-vs-300) & 350 & \textbf{53.14 [47.99, 58.29]} & 0.0268 \\
\midrule
\multirow{4}{*}{\textbf{IRS-32B}}
& Matching & 300 & 62.67 [57.33, 68.33] & 0.028 \\
& Ranking & 385 & 68.05 [63.38, 72.47] & 0.0238 \\
& Ranking (10-vs-1000) & 350 & \textbf{62.86 [57.71, 68.00]} & 0.0258 \\
& Ranking (30-vs-300) & 350 & \textbf{53.14 [47.71, 58.29]} & 0.0268 \\
\midrule
\multirow{4}{*}{\textbf{IRS-72B}}
& Matching & 300 & \textbf{69.33 [64.00, 74.33]} & 0.0267 \\
& Ranking & 385 & \textbf{76.10 [71.69, 80.26]} & 0.0217 \\
& Ranking (10-vs-1000) & 350 & 62.57 [57.43, 67.71] & 0.0260 \\
& Ranking (30-vs-300) & 350 & 50.86 [45.43, 56.00] & 0.0269 \\
\bottomrule
\end{tabular}
\label{tab:bootstrap_ci}
\end{table*}

To evaluate model performance, we report accuracy along with 95\% confidence intervals (CI) and standard errors. Confidence intervals were calculated using non-parametric percentile bootstrapping with $B = 10,000$ resamples drawn with replacement. For each dataset split of size $n$, $B$ independent bootstrap resamples were drawn, and sample accuracy was computed per resample. The 95\% confidence interval was defined by the $2.5^{\text{th}}$ and $97.5^{\text{th}}$ percentiles of the empirical bootstrap distribution, and standard error was computed as the standard deviation of the resampled means.

Table~\ref{tab:bootstrap_ci} presents the comparative evaluation of IRS-7B, IRS-32B, and IRS-72B across all benchmark splits. IRS-72B improves over IRS-7B by 9.66 and 11.68 percentage points on matching and ranking, respectively; the ranking improvement is accompanied by non-overlapping 95\% bootstrap confidence intervals.

A clear performance drop is observed as split difficulty increases. On 30-vs-300 Ranking split, all three models operate near random-chance accuracy ($50\%$), indicating that subtle visual-textual incongruities in this subset remain challenging even for the 72B parameter model.

\begin{figure*}[!h]  
\renewcommand{\thefigure}{I.1}
\includegraphics[width=\linewidth]{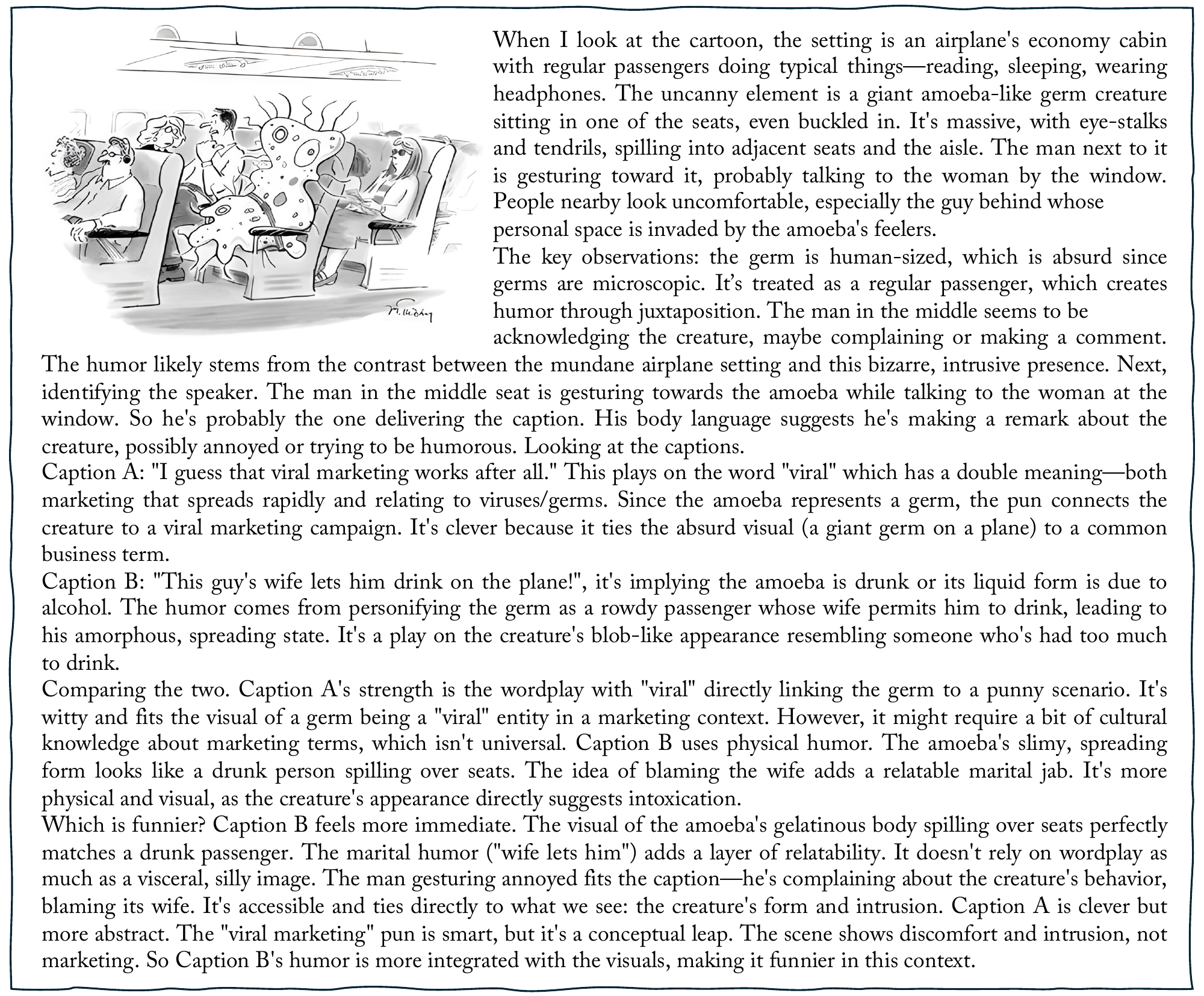}\vspace{-0.5em}
\caption{\textbf{Example reasoning trace for a caption ranking task.} The trace compares two captions against the visual setup (a giant amoeba crowding an airplane seat) and explains why Caption B (``This guy’s wife lets him drink on the plane!'') better resolves the incongruity and captures the visual humor. This example illustrates how ranking traces encode comparative reasoning over humor mechanisms such as visual grounding and punchline strength.}
\label{fig:cot_ranking_example}
\end{figure*}
\begin{figure*}[!h]  
\renewcommand{\thefigure}{I.2}
\centering
\includegraphics[width=\linewidth]{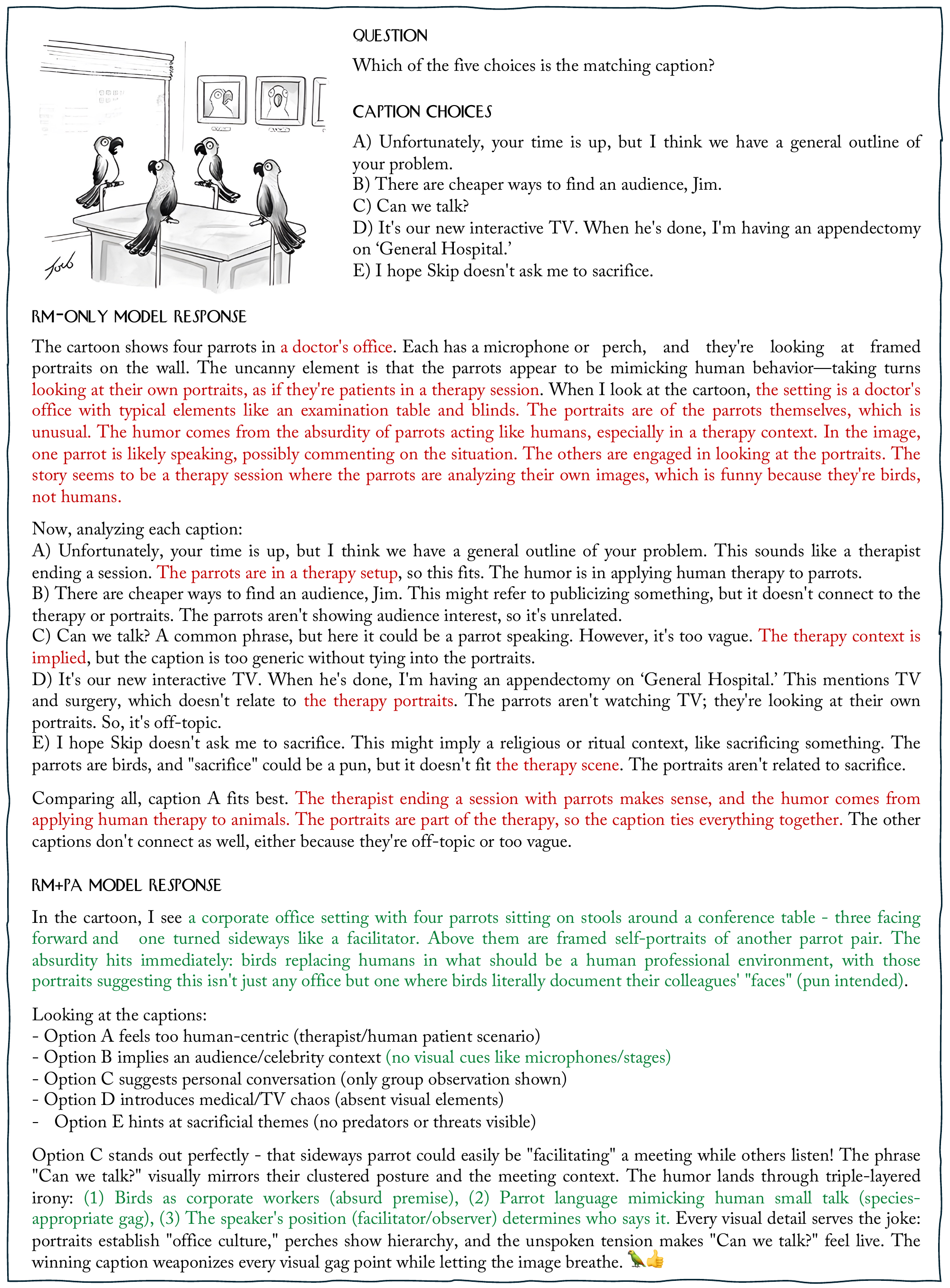}\vspace{-0.5em}
\caption{\textbf{Example reasoning trace for a caption matching task.} The RM-only model produces fluent but superficial descriptions, often framing the parrots’ scene as a ``therapy session'' without fully connecting this to the core incongruity of the humor. The RM+PA model, by contrast, grounds its reasoning in salient visual cues, filters irrelevant distractors, and highlights layered humor mechanisms such as species-appropriate wordplay, speaker positioning, and irony, producing commentary closer to professional captionist style.}
\label{fig:response-example1}
\end{figure*}

\section{Additional Examples: Ground Truth Traces, Evolution of Model Reasoning, and Judge Responses}
\label{sec:examples}
\setcounter{figure}{2}
\setcounter{table}{0}
To complement the quantitative results in the main paper, we provide qualitative examples that illustrate how reasoning traces, model outputs, and reward signals interact to shape humor-aware reasoning under IRS.

\subsection{Ground-Truth Traces} 
In addition to the matching example in the main text (Fig.~\ref{fig:cot_matching_example}), we include a ground-truth reasoning trace for a ranking task. As illustrated in Fig.~\ref{fig:cot_ranking_example}, the trace demonstrates how captionists compare alternatives by identifying which caption better exploits the visual incongruity and delivers a more effective punchline. These traces form the structured supervision used during Resolution Modeling.

\subsection{Evolution of Model Reasoning.}  
To examine how IRS shapes reasoning, we compare outputs from RM-only and RM+PA models on matching and ranking tasks.

Fig.~\ref{fig:response-example1} presents a caption matching example. The RM-only model produces fluent but shallow explanations, often identifying surface-level themes (e.g., a ``therapy session'') without fully connecting them to the central incongruity. Its reasoning is verbose and only partially grounded in the image. In contrast, the RM+PA model anchors its explanation in salient visual cues, filters distractors, and highlights multiple interacting humor mechanisms, including wordplay, speaker roles, and ironic contrasts. The output reads closer to a captionist’s commentary: concise, well-structured, and attuned to comedic effect.

\begin{figure*}[!h]  
\includegraphics[width=\linewidth]{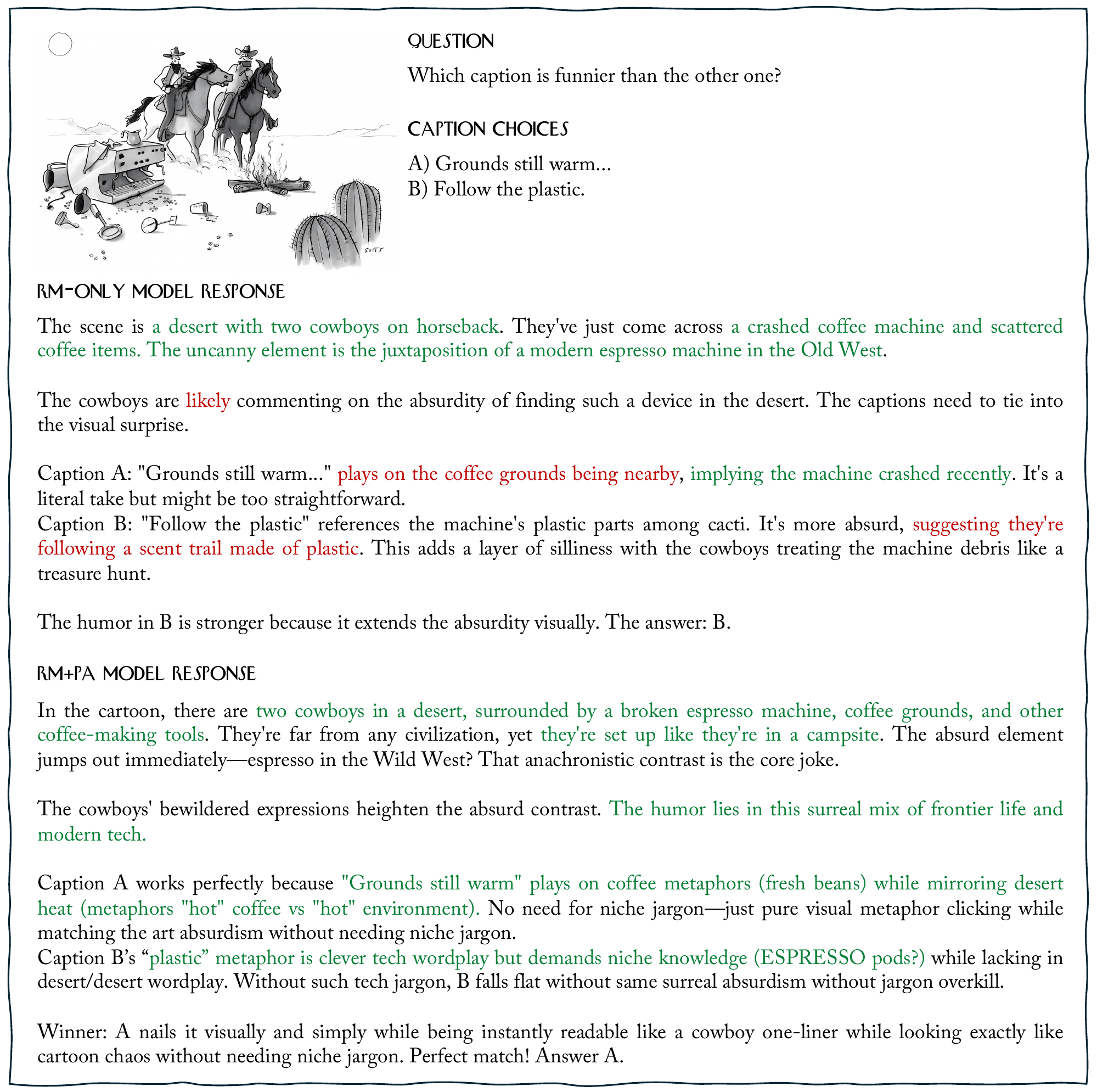}\vspace{-0.5em}
\caption{\textbf{Example reasoning on a caption ranking task.} The RM-only model fixates on literal associations (e.g., coffee grounds as evidence of recency), while the RM+PA model integrates the broader absurdity of encountering a modern espresso machine in a cowboy setting. By grounding its reasoning in visual context and avoiding reliance on niche jargon, the RM+PA model produces a more coherent and humor-aware judgment that better reflects the cartoon’s visual gag.}
\label{fig:response-example2}
\end{figure*}

Fig.~\ref{fig:response-example2} shows a ranking example, where the difference is more pronounced. The RM-only model focuses on literal associations (e.g., coffee grounds as evidence of recency), while missing the broader incongruity of cowboys interacting with modern espresso machines. The RM +PA model instead integrates visual context and stylistic considerations, selecting the caption that better captures the underlying joke without relying on overly specific cues. 

These examples are consistent with the quantitative results: PA improves not only accuracy but also the structure and style of reasoning, shifting outputs toward more grounded and economical explanations.

\subsection{Judge Responses}  
\begin{figure*}[!h]
  \centering
  \includegraphics[width=\linewidth]{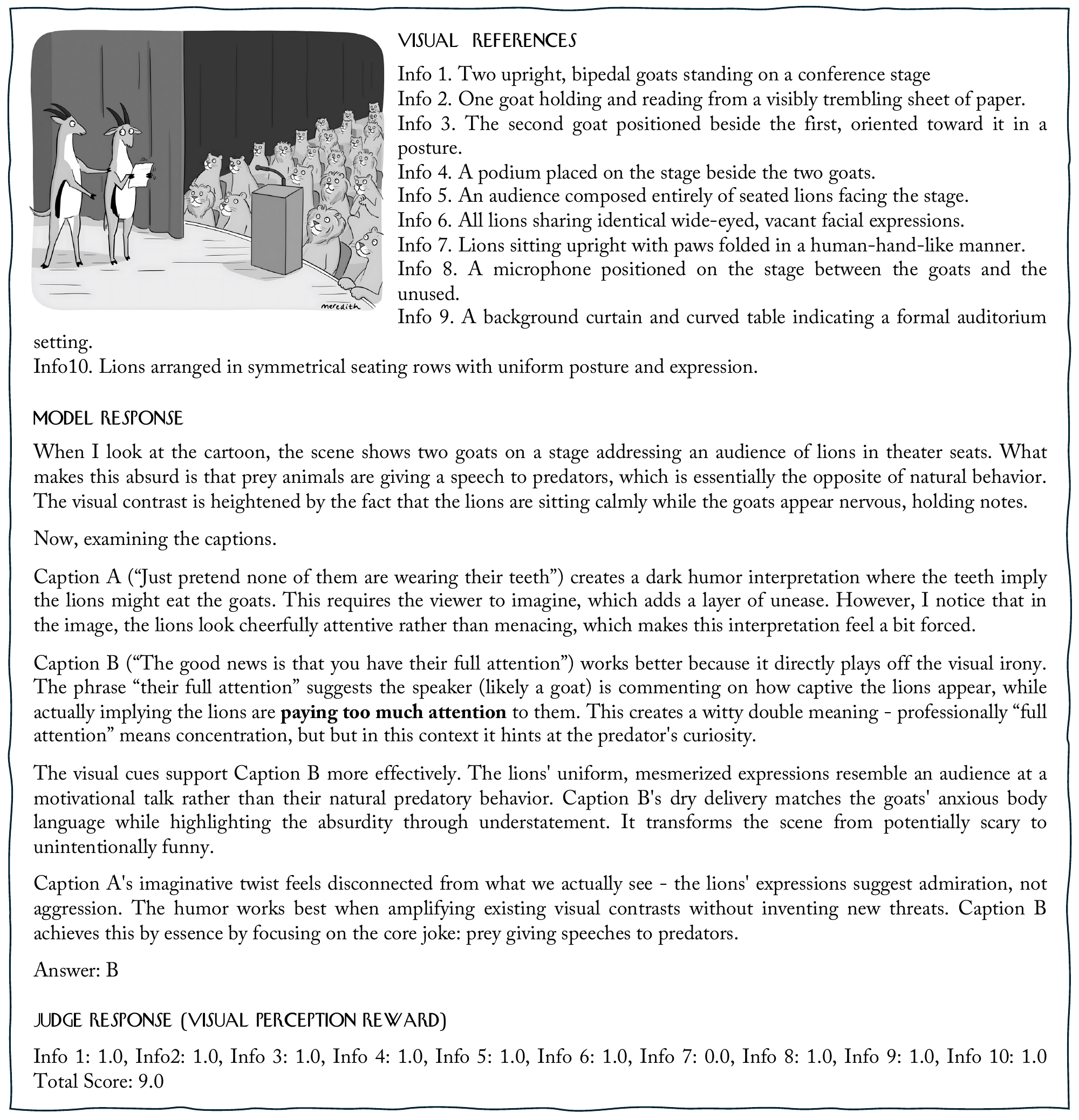}\vspace{-0.5em}
  \caption{\textbf{Visual perception judge response.} Binary scores indicating whether the reasoning reflects curated visual references, providing a direct measure of grounding in salient scene elements.}
  \label{fig:judge_perception}
\end{figure*}

\begin{figure*}[!h]
  \centering
  \includegraphics[width=\linewidth]{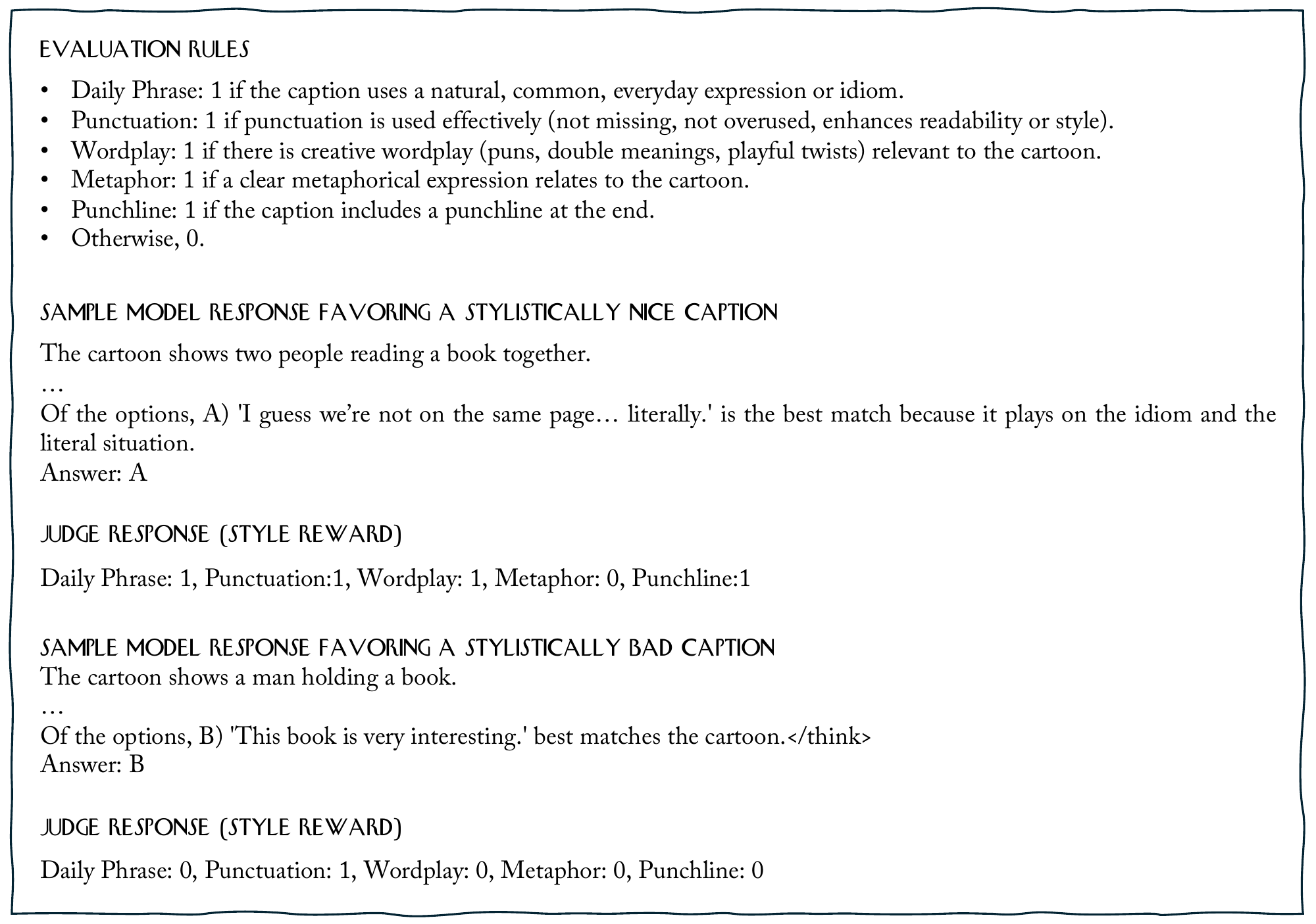}\vspace{-0.5em}
  \caption{\textbf{Style judge response.}  Binary scores across five stylistic dimensions (natural phrasing, punctuation, wordplay, metaphor, and punchline placement), capturing linguistic qualities associated with effective caption writing.}
  \label{fig:judge_style}
\end{figure*}

Fig.~\ref{fig:judge_perception} and~\ref{fig:judge_style} illustrate the outputs of the LLM-as-judge components used to compute perception and style rewards. In Fig.~\ref{fig:judge_perception}, the perception judge compares the model’s reasoning against curated visual references and returns per-attribute binary scores. This ensures that explanations are explicitly grounded in salient scene elements. In Fig.~\ref{fig:judge_style}, the style judge resolves the model’s \texttt{<answer>} to the selected caption and evaluates it along five dimensions: natural phrasing, punctuation, wordplay, metaphor, and punchline placement. These binary signals provide structured feedback on linguistic quality and humor delivery. Together, these reward components translate abstract notions of grounding and stylistic quality into measurable signals used during PA.

\subsection{Comparison with Expert Traces}

\begin{figure*}[!h]
  \centering
  \includegraphics[width=\linewidth]{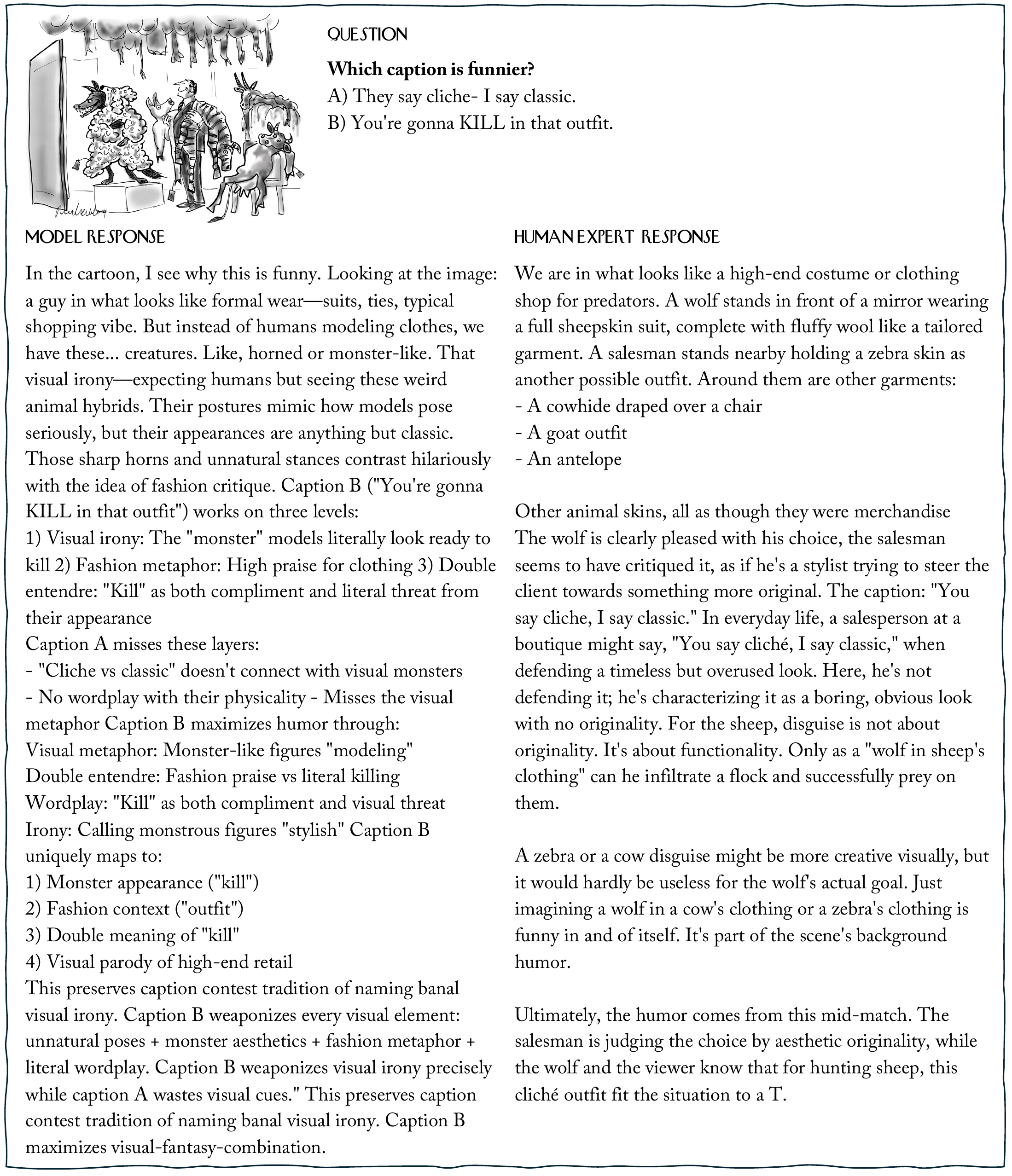}
  \caption{\textbf{Model vs.\ expert reasoning.} The model and expert reach different conclusions but follow similar reasoning steps, indicating alignment in how the cartoon is analyzed even when preferences differ.}
  \label{fig:comparison_model_expert1}
\end{figure*}

We include an example where the model’s final choice differs from that of a professional captionist. As shown in Fig.~\ref{fig:comparison_model_expert1}, the model nonetheless follows a similar reasoning process, identifying the central incongruity, evaluating alternative interpretations, and providing a structured justification. This alignment in reasoning structure, despite differing conclusions, suggests that the model has learned aspects of the captionist analysis process beyond surface-level pattern matching.

\subsection{Discussion}
The examples in Appendix~\ref{sec:examples} provide qualitative evidence of how IRS enables humor-aware reasoning. Ground-truth traces illustrate the structured supervision signals used during training, while model outputs show how reasoning evolves from surface-level associations to more grounded and stylistically consistent explanations during RM. The judge responses further reveal how PA translates qualitative criteria such as grounding and humor style into explicit reward signals. Together, these examples demonstrate that the improvements observed in quantitative evaluations are accompanied by systematic changes in reasoning behavior. Rather than treating humor as a black-box prediction task, IRS supports a decomposable and interpretable reasoning process, making it possible to analyze how models arrive at their decisions.

\subsection{Limitations}
\begin{figure*}[!t]
\begin{center}
\includegraphics[width=\linewidth]{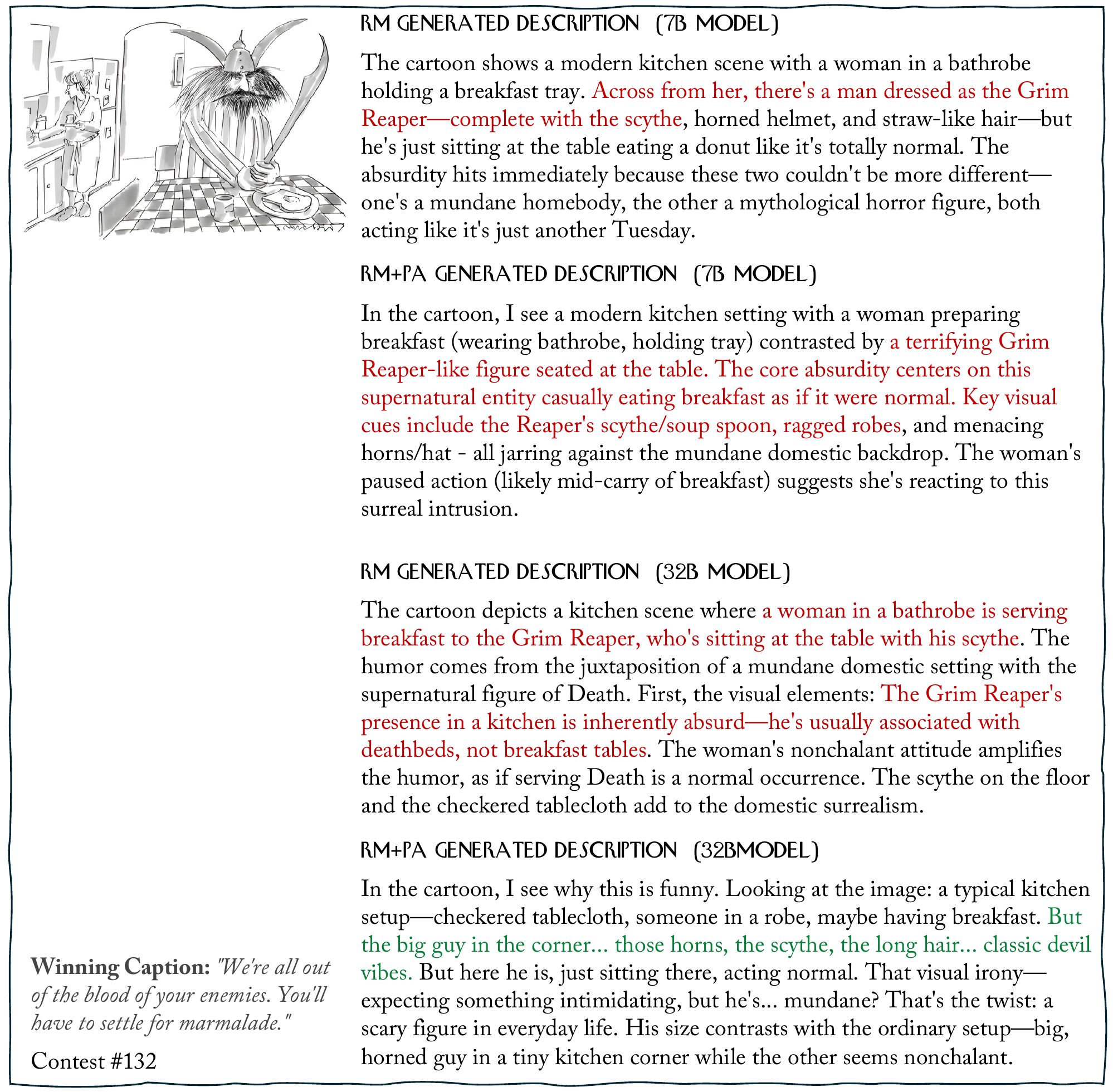}\vspace{-1.15em}
\end{center}

\caption{\textbf{Limitation: Visual perception errors.} All 7B-scale variants misidentify the foreground figure as the Grim Reaper (with scythe), though the humor relies on the incongruity of a Viking warrior in a domestic kitchen; this error cascades into mismatched reasoning and an incorrect caption. At 32B, the model misidentifies objects early in training but, with the two additional PA rewards, corrects this and identifies the objects correctly.\vspace{-0.275cm}}
\label{fig:lim-perception}
\end{figure*}

\begin{figure*}[!t]
\begin{center}

\includegraphics[width=\linewidth]{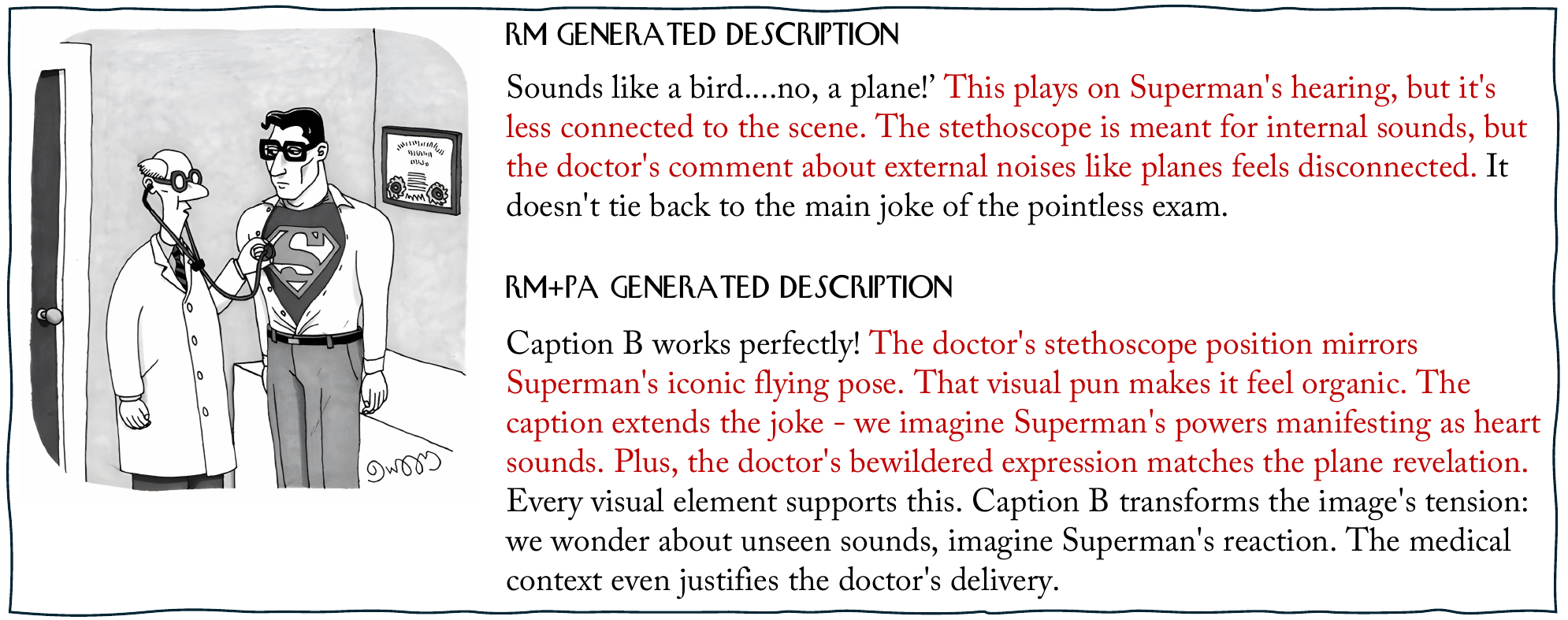}\vspace{-1.15em}
\end{center}
\caption{\textbf{Limitation: Shallow cultural understanding.} Sample question from ranking task with the choices A) "SORRY, YOU DON'T QUALIFY FOR OBAMACARE.", B) "Sounds like a bird....no, a plane!". The humor depends on a Superman reference (the \textit{``it’s a bird… it’s a plane…''} trope). The RM-only model fixates on superficial wordplay; RM+PA recognizes the pop-culture context but hallucinates a visual link and does not fully recover the cultural script.}

\label{fig:lim-culture}
\end{figure*}

We provide two qualitative examples (Figure~\ref{fig:lim-perception} and Figure~\ref{fig:lim-culture}) highlighting limitations of IRS and discuss the corresponding failure cases in the Section~\ref{sec:limitations}.

\section{Quantitative Error Taxonomy}
\label{sec:error_taxonomy}

To better understand failure cases, we analyze the incorrect predictions of IRS-72B across all benchmark splits. Each error is assigned to one of seven predefined categories by Gemini-2.5-Flash \citep{comanici2025gemini25pushingfrontier}. Table~\ref{tab:error_taxonomy_separated} summarizes the resulting error distribution. Errors are classified into seven failure categories:

\begin{itemize}[leftmargin=*]
    \item \textbf{Visual Misperception \& Hallucination:} Misidentifying physical objects, characters, or background elements, or hallucinating non-existent scene components.
    \item \textbf{Visual Incongruity Blindness:} Correctly identifying individual objects while failing to recognize the central uncanny visual anomaly or contradiction.
    \item \textbf{Humor Mechanism \& Pun Misinterpretation:} Failing to resolve idioms, metaphors, or wordplay in candidate captions.
    \item \textbf{Speaker Misdetection:} Failing to determine which character in the cartoon is speaking or receiving the dialogue line, leading to an inverted perspective on the joke.
    \item \textbf{Overthinking \& Post-Hoc Rationale Overfit:} Constructing an overly complex, post-hoc philosophical narrative to justify an incorrect choice.
    \item \textbf{Option Discrimination Failure:} Selecting a plausible caption without properly contrasting it against the funnier, visually-anchored punchline.
    \item \textbf{Cultural, Domain, or Anachronism Blindness:} Failing to recognize pop-culture references, historical contexts, brand names, or specific tropes.
\end{itemize}

\renewcommand{\arraystretch}{1.05}
\begin{table*}[!t]
\centering
\caption{\textbf{Error Taxonomy Across Matching and Ranking tasks for IRS-72B ($N = 487$ errors).} Columns report raw error counts separated by task types.}
\label{tab:error_taxonomy_separated}
\begin{tabular}{lrrrrrr}
\toprule
\multirow{2}{*}{\textbf{Error Category}} & \multicolumn{2}{c}{\textbf{(Hessel et al., 2023)}} & \multicolumn{2}{c}{\textbf{(Zhou et al., 2025)}} \\
& \textbf{Matching} & \textbf{Ranking} & \textbf{10-vs-1000} & \textbf{30-vs-300} \\
\midrule
Humor Mechanism \& Pun Misinterpretation & 0 & 45 & 49 & 73 \\
Visual Misperception \& Hallucination & 81 & 18 & 34 & 19 \\
Cultural / Domain / Anachronism Blindness & 3 & 20 & 24 & 24 \\
Overthinking \& Post-Hoc Rationale Overfit & 5 & 7 & 20 & 28 \\
Option Discrimination Failure & 1 & 1 & 2 & 14 \\
Visual Incongruity Blindness & 1 & 1 & 2 & 13 \\
Speaker Misdetection & 1 & 0 & 0 & 1 \\
\bottomrule
\end{tabular}
\end{table*}

In the Matching task, visual misperception and hallucination account for 81 of 92 errors. These errors often involve incorrectly perceived objects, attributes, or other visual details, which can cause the model to select a distractor caption. In contrast, humor interpretation errors are more common in the ranking task. Since both candidate captions are related to the visual content, these errors are more closely related to distinguishing the intended humorous interpretation.

Cultural and domain knowledge errors account for 71 errors overall, relatively more errors in the Ranking splits. These cases involve captions that depend on information beyond the visual content, such as references to historical figures, cultural concepts, or brands. As the difficulty level increases in the Ranking task, overthinking errors become more frequent. These cases are characterized by increasingly elaborate reasoning that leads to an incorrect caption choice. The remaining categories occur much less frequently and increase as the ranking questions become harder.

Overall, the error distribution shows a clear difference between the two task types: Matching errors are dominated by visual grounding failures, whereas Ranking errors more often involve humor interpretation and comparative reasoning.

\section{Reasoning-Chain Quality Audit}
\label{sec:audit}
\setcounter{figure}{0}
\setcounter{table}{0}
To assess the quality of the resulting reasoning traces, we perform a structured reasoning-chain audit using DeepSeek \citep{deepseekai2025deepseekr1incentivizingreasoningcapability}, NVLM-D-72B \citep{dai2024nvlmopenfrontierclassmultimodal}, and Gemini-2.5-Flash \citep{comanici2025gemini25pushingfrontier} as external evaluators. Rather than evaluating only answer correctness, the audit scores traces along several dimensions that are listed below.  

\parahead{Matching-task audit dimensions}
For matching tasks, where the model selects one caption from five candidates, we evaluate the following dimensions:
\begin{itemize}[leftmargin=*]
    \item \textbf{Incongruity:} whether the reasoning identifies the central mismatch, tension, or absurdity that drives the humor.
    \item \textbf{Humor Insight:} whether the explanation captures specific humor mechanisms such as wordplay, metaphor, reversal, or social convention.
    \item \textbf{Caption Alignment:} whether the reasoning clearly explains why the chosen caption matches the cartoon.
    \item \textbf{Option Discrimination:} whether the reasoning explains why the selected caption is better than the alternatives.
\end{itemize}

\parahead{Ranking-task audit dimensions}
For ranking tasks, where the model compares two captions, we evaluate:
\begin{itemize}[leftmargin=*]
    \item \textbf{Incongruity Alignment:} whether it identifies the primary source of humor.
    \item \textbf{Caption-Scene Fit:} whether the reasoning explains how each caption interacts with the visual elements.
    \item \textbf{Humor Sensitivity:} whether the reasoning captures subtle differences in timing, understatement, absurdity, or cultural framing.
    \item \textbf{Funniness Justification:} whether the rationale explains why one caption is funnier, not merely more coherent.
    \item \textbf{Comparative Evaluation:} whether the two captions are explicitly compared rather than discussed in isolation.
\end{itemize}

As shown in Table~\ref{tab:audit_rm}, we conduct a cross-model quality audit using three independent judges from distinct model families, parameter scales, and training origins: DeepSeek-R1, NVLM-D-72B, and Gemini-2.5-Flash to eliminate potential self-assessment bias inherent in evaluating reasoning traces using the model family that generated them. Absolute scores vary across judges due to differences in internal calibration applied by each model family to the evaluation rubric. Across both matching and ranking tasks, all three independent judges consistently assess the reasoning traces as high quality. 

Crucially, the relative ordering across rubric dimensions is strictly preserved under every evaluator: higher-order visual-semantic metrics, such as \textit{Incongruity} and \textit{Incongruity Alignment}, consistently score highest, whereas discriminative metrics like \textit{Option Discrimination} and \textit{Funniness Justification} remain consistently lowest. This convergent ranking across three model families, despite calibration differences, is stronger evidence of trace quality than any single judge's absolute scores, and indicates the audit reflects genuine properties of the traces rather than the idiosyncrasies of one evaluator.

To evaluate whether IRS enhances underlying reasoning mechanisms rather than merely optimizing final-answer accuracy, we complement our performance metrics with a component-level rationale quality assessment on inference-time rationales generated on the held-out test set. We compare IRS (32B) against its underlying base architecture (Qwen2.5-VL-32B) configured with zero-shot and few-shot CoT prompting. All rationales are independently assessed using Gemini-2.5-Flash. 

Table~\ref{tab:audit_test} summarizes the results. IRS achieves superior scores across all evaluated dimensions and overall averages in both matching and ranking tasks.

Overall, these results suggest that IRS improves not only answer accuracy but also the quality and structure of the generated reasoning itself. The strong incongruity-related scores are particularly consistent with the central objective of IRS: explicitly supervising incongruity-resolution reasoning rather than relying solely on implicit reasoning emerging from scale.

We provide full audit prompts below.

\parahead{Audit prompts}
The audit prompts instruct models to act as a skeptical expert reviewer rather than to regenerate a new interpretation. The evaluator is asked to identify weaknesses such as superficial humor analysis, and to return only a structured JSON object containing scores, flaws, strengths, reflection, revised answer, and revised rationale.

\begin{tcolorbox}[colback=gray!5,breakable,colframe=black,title=Matching Task Audit Prompt]
\small
You are a senior cartoon-caption analyst reviewing a model's reasoning trace for a New Yorker Caption Contest matching task.
You are given:
\begin{itemize}[leftmargin=*]
\item a cartoon scene description,
\item the unusual or uncanny element,
\item key observations,
\item five candidate captions,
\end{itemize}
and a step-by-step reasoning chain produced by a model, and the model's selected answer.

Your job now is not to generate a new interpretation from scratch, but to critically audit the provided reasoning trace as an expert reviewer.

You must assess whether the model's reasoning:
\begin{enumerate}[leftmargin=*]
\item identified the central incongruity that actually drives the humor,
\item explained why the selected caption matches better than the alternatives,
\item avoided superficial, generic, or unsupported interpretations.
\end{enumerate}

\textbf{Important:}
\begin{itemize}[leftmargin=*]
\item Do not simply restate the model's reasoning.
\item Look for overreach, hallucinated visual assumptions, weak humor analysis, and poor discrimination among options.
\item Be strict and skeptical.
\item If the reasoning is weak, say so clearly.
\item If needed, revise the final choice.
\end{itemize}

\textbf{Input:}

Scene: [SCENE]

Description: [DESCRIPTION]

Uncanny Element: [UNCANNY ELEMENT]

Observations: [OBSERVATIONS]

Candidate Captions:

A) [CAPTION A]

B) [CAPTION B]

C) [CAPTION C]

D) [CAPTION D]

E) [CAPTION E]

Model's Reasoning:

[REASONING CHAIN]

Model's Selected Answer:

[ANSWER]

Perform a professional audit along the following dimensions:

1. Incongruity Diagnosis: Did it identify the real mismatch, tension, or absurdity that makes the cartoon funny?

2. Caption-Image Alignment: Did it clearly explain why the chosen caption matches the cartoon?

3. Humor Interpretation: Did it explain the humor in a specific and non-trivial way, including metaphor, reversal, social convention, or wordplay where relevant?

4. Option Discrimination: Did it explain why the chosen caption is better than the other four, rather than merely sounding plausible on its own?

Respond ONLY in valid JSON with the following format:
\begin{verbatim}
{
  "scores": {
    "incongruity": 0.0,
    "humor_insight": 0.0,
    "caption_alignment": 0.0,
    "option_discrimination": 0.0,
  },
  "major_flaws": [
    "...",
    "..."
  ],
  "minor_flaws": [
    "...",
    "..."
  ],
  "best_supported_elements": [
    "...",
    "..."
  ],
  "reflection": "A concise expert assessment 
  of whether the original reasoning is 
  trustworthy.",
  "revised_answer": "A/B/C/D/E",
  "revised_rationale": "A short explanation 
  of whether the original answer should 
  stand or be changed."
}
\end{verbatim}
\normalsize
\end{tcolorbox}

\begin{tcolorbox}[colback=gray!5,breakable,colframe=black,title=Ranking Task Audit Prompt]
\small
You are a senior cartoon-caption editor reviewing a model's reasoning trace for a New Yorker Caption Contest ranking task.

You are given:
\begin{itemize}[leftmargin=*]
\item a cartoon scene description,
\item an uncanny or unusual element,
\item key observations,
\item two candidate captions,
\end{itemize}
and a step-by-step reasoning chain produced by a model, and the model's selected answer.

Your task now is to critically audit the provided reasoning trace as an expert evaluator of humor.

You must assess whether the model's reasoning:
\begin{enumerate}[leftmargin=*]
\item correctly identified the central incongruity driving the humor,
\item evaluated both captions comparatively (not in isolation),
\item captured subtle differences in tone, timing, and relevance,
\item justified why one caption is \textit{funnier}, not just plausible.
\end{enumerate}

\textbf{Important:}
\begin{itemize}[leftmargin=*]
\item Do not simply restate the model's reasoning.
\item Be skeptical and look for weaknesses.
\item Focus on comparative humor judgment, not absolute interpretation.
\item If the reasoning is weak or biased, say so clearly.
\item Revise your decision if necessary.
\end{itemize}

\textbf{Input:}

Scene: [SCENE]

Description: [DESCRIPTION]

Uncanny Element: [UNCANNY ELEMENT]

Observations: [OBSERVATIONS]

Captions:

A) [CAPTION A]

B) [CAPTION B]

Model's Reasoning:

[REASONING CHAIN]

Model's Selected Answer:

[ANSWER]

Perform a professional audit along the following dimensions:

1. Incongruity Alignment: Did it correctly identify the \textit{main} source of humor in the cartoon?

2. Caption-Scene Fit: Did it clearly explain how each caption interacts with the visual?

3. Humor Sensitivity: Did it capture nuances such as timing, understatement, absurdity, or cultural framing?

4. Funniness Justification: Did it justify \textit{why one is funnier}, not just more coherent?

5. Comparative Evaluation: Did it explicitly compare A vs B, or just explain one in isolation?

Respond ONLY in valid JSON:
\begin{verbatim}
{
  "scores": {
    "incongruity_alignment": 0.0,
    "comparative_evaluation": 0.0,
    "humor_sensitivity": 0.0,
    "caption_scene_fit": 0.0,
    "funniness_justification": 0.0,
  },
  "major_flaws": [
    "...",
    "..."
  ],
  "minor_flaws": [
    "...",
    "..."
  ],
  "strengths": [
    "...",
    "..."
  ],
  "reflection": "Concise expert judgment on 
  whether the original reasoning is 
  reliable.",
  "revised_answer": "A/B",
  "revised_rationale": "Short explanation 
  of whether and why the choice should 
  change."
}
\end{verbatim}
\normalsize
\end{tcolorbox}

\begin{table}[!t]
\centering
\caption{\textbf{Cross-Model Reasoning chain audit of RM supervision traces.}
DeepSeek-R1, NVLM-D-72B, and Gemini-2.5-Flash evaluate the traces across multiple rationale-quality dimensions. The generated traces consistently exhibit stronger higher-level humor reasoning, particularly in incongruity alignment, humor sensitivity, caption-scene fit, and funniness justification.}
\label{tab:audit_rm}
\small
\resizebox{\linewidth}{!}{
\begin{tabular}{@{}l@{}c@{$\;$}c@{}c@{}}
\toprule
\textbf{Matching Tasks} & & &\\
\midrule
\textbf{Metric} & \textbf{DeepSeek-R1} & \textbf{NVLM-D-72B} & \textbf{Gemini-2.5-Flash} \\
\midrule
Incongruity & 0.7525 & 0.9892 & 0.9629 \\
Caption Alignment & 0.6310 & 0.8469 & 0.9120 \\
Humor Insight & 0.6072 & 0.8322 & 0.8786 \\
Option Discrimination & 0.4903 & 0.7349 & 0.8516 \\
\textbf{Overall Score} & \textbf{0.6202} & \textbf{0.8508} & \textbf{0.9013} \\
\midrule 
\\ \midrule
\textbf{Ranking Tasks} & & & \\
\midrule
\textbf{Metric} & \textbf{DeepSeek-R1} & \textbf{NVLM-D-72B} & \textbf{Gemini-2.5-Flash} \\
\midrule
Incongruity Alignment & 0.7175 & 0.9305 & 0.9303 \\
Caption Scene Fit & 0.6554 & 0.7202 & 0.8334 \\
Humor Sensitivity & 0.5464 & 0.7817 & 0.7224 \\
Funniness Justification & 0.4532 & 0.6064 & 0.6914 \\
Comparative Evaluation & 0.5088 & 0.5428 & 0.8761 \\
\textbf{Overall Score} & \textbf{0.5762} & \textbf{0.7163} & \textbf{0.8107} \\
\bottomrule
\end{tabular}
}
\end{table}

\begin{table}[!t]
\centering
\small
\caption{\textbf{Reasoning chain audit on the test set.}
We report the Gemini-2.5-Flash scores of inference-time rationales satisfying component-level rubric criteria across matching and ranking tasks.}
\label{tab:audit_test}
\begin{tabular}{@{}lccc@{}}
\toprule
\textbf{Matching Tasks} & & &\\
\midrule
\textbf{Metric} & \textbf{Zero-shot} & \textbf{Few-shot} & \textbf{IRS} \\
\midrule
Incongruity & 0.3470 & 0.3637 & \textbf{0.4072} \\
Caption Alignment & 0.3030 & 0.3377 & \textbf{0.3551} \\
Humor Insight & 0.3671 & 0.3775 & \textbf{0.4042} \\
Option Discrimination & 0.3441 & 0.4065 & \textbf{0.4342} \\
\midrule
\textbf{Overall Score} & 0.3403 & 0.3714 & \textbf{0.4002} \\
\midrule
\\ \midrule
\textbf{Ranking Tasks} & & &\\
\midrule
\textbf{Metric} & \textbf{Zero-shot} & \textbf{Few-shot} & \textbf{IRS} \\
\midrule
Incongruity Alignment & 0.2914 & 0.3089 & \textbf{0.3452} \\
Caption Scene Fit & 0.3015 & 0.2711 & \textbf{0.3587} \\
Humor Sensitivity & 0.2693 & 0.2649 & \textbf{0.3160} \\
Funniness Justification & 0.2307 & 0.2227 & \textbf{0.2810} \\
Comparative Evaluation & 0.4985 & 0.5339 & \textbf{0.5976} \\
\midrule
\textbf{Overall Score} & 0.3183 & 0.3203 & \textbf{0.3797}\\
\bottomrule
\end{tabular}
\end{table}
\section{Caption Generation without Task-Specific Training}
\label{sec:captioning}
\setcounter{figure}{0}
\setcounter{table}{0}
Although IRS is designed for caption selection and ranking, we observe that models trained with IRS can also generate plausible captions in a zero-shot manner. This is notable because caption generation is not explicitly included in our training objectives. We evaluate this capability qualitatively by prompting the model to produce a single caption for a given cartoon, following the New Yorker Caption Contest format. 

\begin{figure*}[!h]
    \centering
    \includegraphics[width=\linewidth]{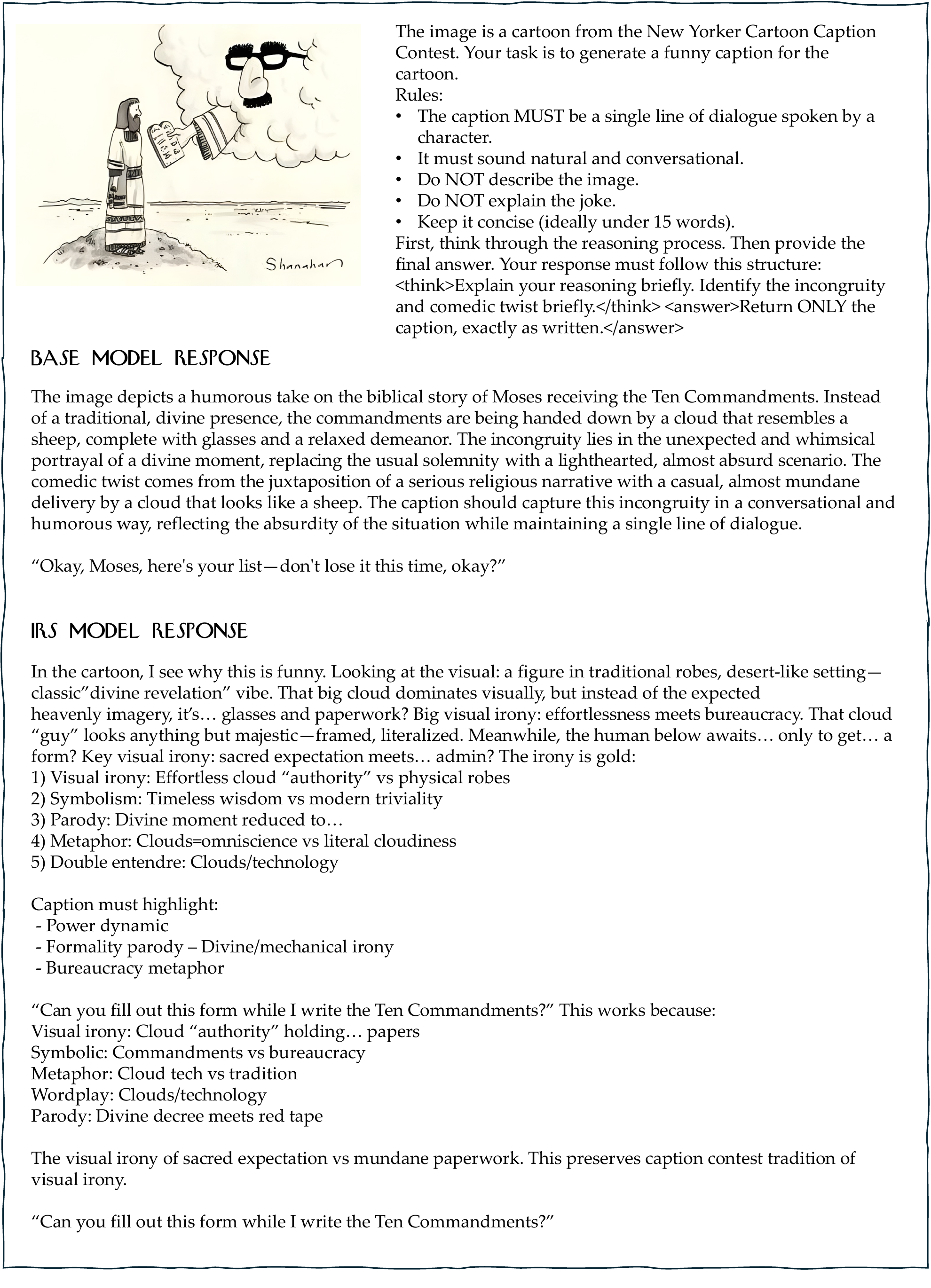}\vspace{-0.5em}
    \caption{\textbf{Zero-shot caption generation with IRS (divine-bureaucratic incongruity).} The IRS model captures the clash between a sacred revelation and mundane paperwork, grounding the humor in a reinterpretation of divine authority as bureaucratic process.}    
    \label{fig:caption-generation-1}
\end{figure*}

\begin{figure*}[!h]
    \centering
    \includegraphics[width=\linewidth]{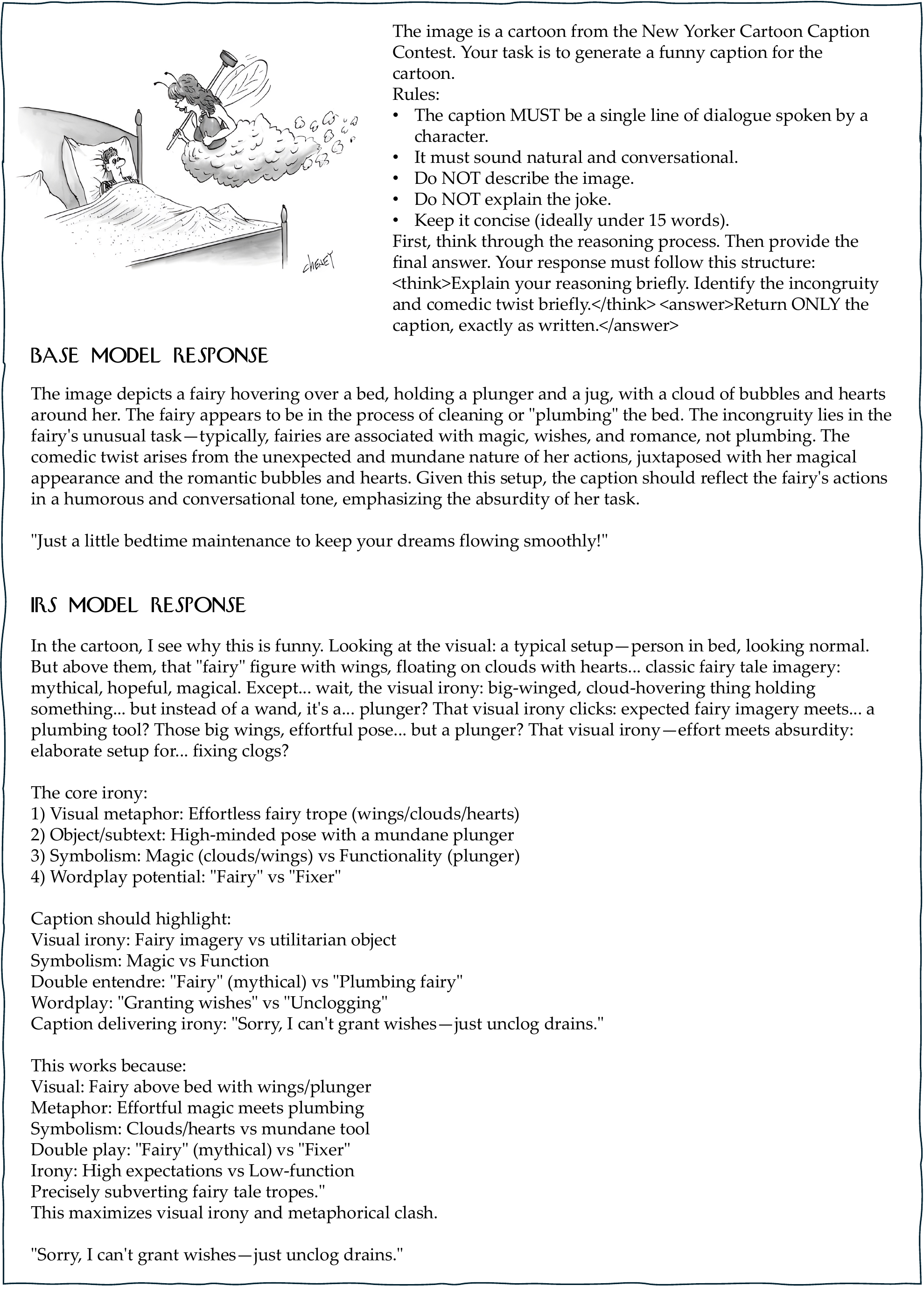}\vspace{-0.5em}
    \caption{\textbf{Zero-shot caption generation with IRS (magic-mundane incongruity).} The IRS model resolves the mismatch between fairy-tale imagery and a utilitarian task, producing a caption that reframes magical expectation as everyday functionality.}    
    \label{fig:caption-generation-2}
\end{figure*}

\begin{figure*}[!h]
    \centering
    \includegraphics[width=\linewidth]{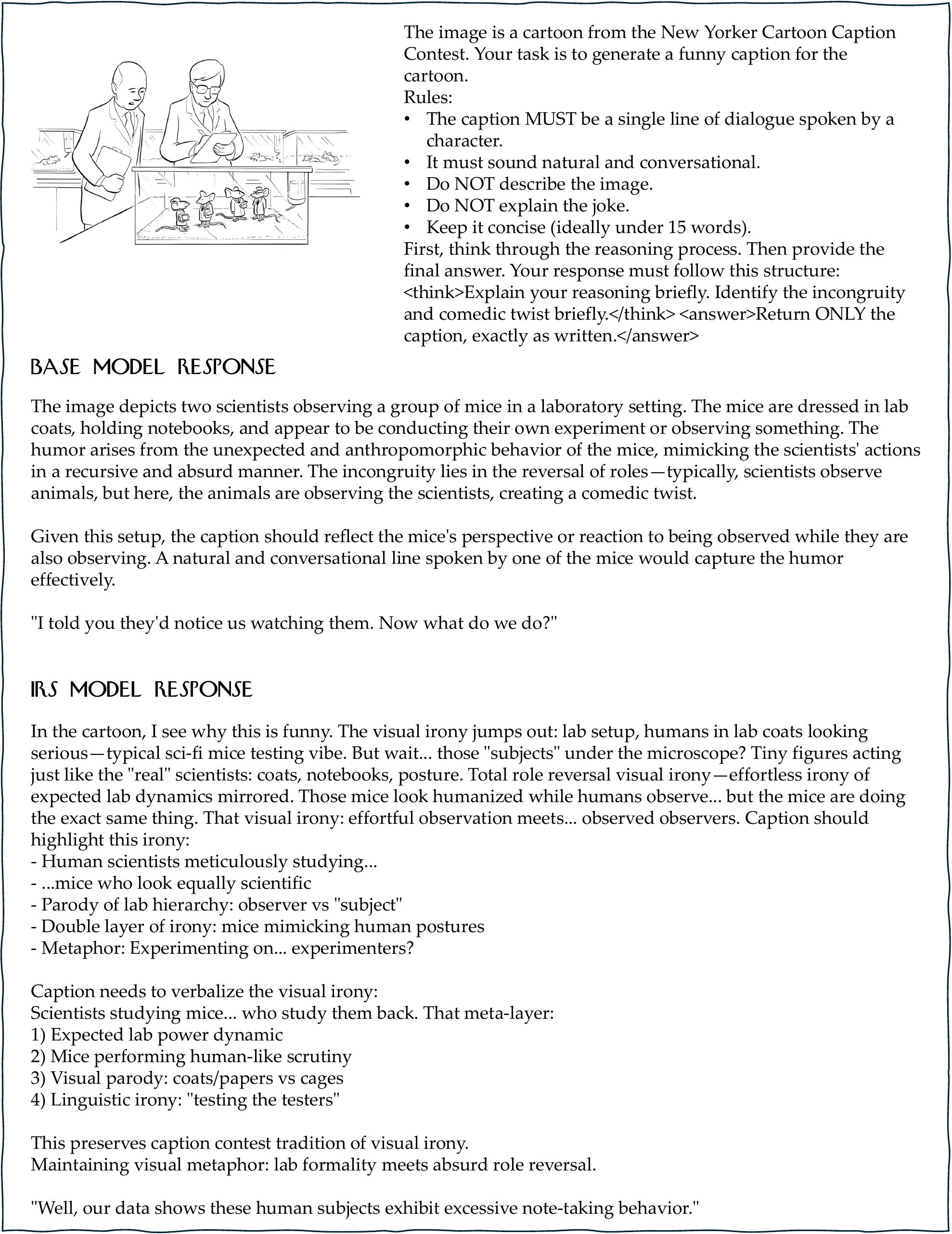}\vspace{-0.5em}
    \caption{\textbf{Zero-shot caption generation with IRS (role-reversal incongruity).} The IRS model captures the recursive reversal between observers and subjects, grounding the humor in a mirrored interpretation of scientific authority.}    
    \label{fig:caption-generation-3}
\end{figure*}

As shown in Fig.~\ref{fig:caption-generation-1}-\ref{fig:caption-generation-3}, the IRS-trained model generates captions that are concise, conversational, and grounded in the visual incongruity, often reflecting captionist-style reasoning patterns such as wordplay, irony, and narrative reinterpretation. Compared to the base model, which often produces plausible but less structured captions, the IRS model more consistently captures the underlying humorous mechanism and its connection to the visual incongruity. This suggests that supervising structured reasoning, rather than optimizing directly for generation, enables the model to internalize transferable patterns that extend naturally to captioning. These results further support our central claim: explicitly modeling and supervising reasoning structure leads to more generalizable capabilities, even in tasks not seen during training.

\section{User Study Details}
\label{app:userstudy}
\setcounter{figure}{0}
\setcounter{table}{0}
To establish a human reference point for the matching and ranking tasks, we conducted a small user study. Participants were shown New Yorker Cartoon Caption Contest cartoons and asked to select the most appropriate caption from the provided options, following the same task formats used in our automated evaluation (five-way selection for matching, two-way selection for ranking).

\begin{figure}[!t]
    \centering
    \includegraphics[width=\columnwidth]{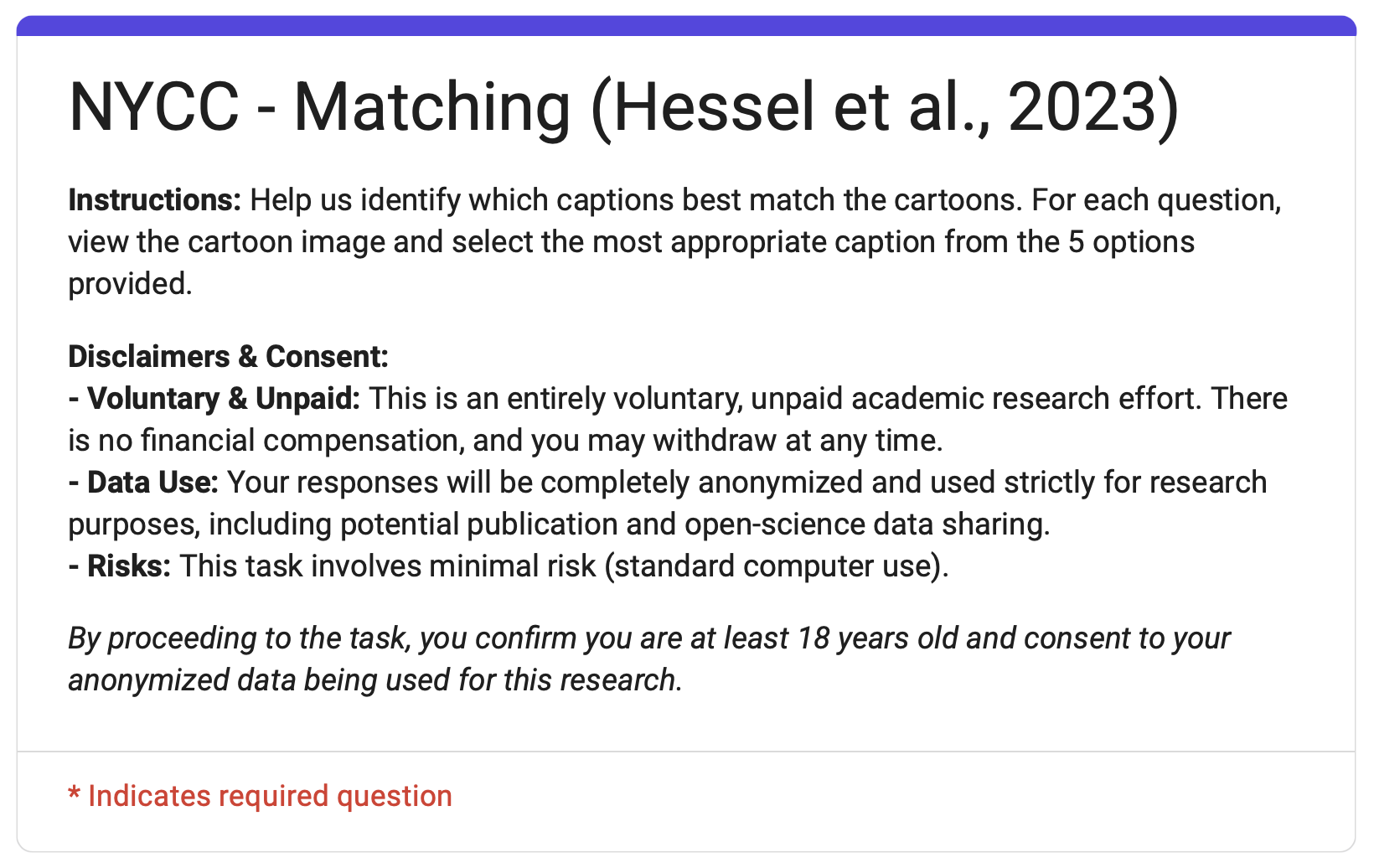}
    \caption{\textbf{Consent form and disclaimers shown to participants before the user study.}}
    \label{fig:user-study-consent}
\end{figure}

\begin{figure}[!htb]
    \centering
    \includegraphics[width=\columnwidth]{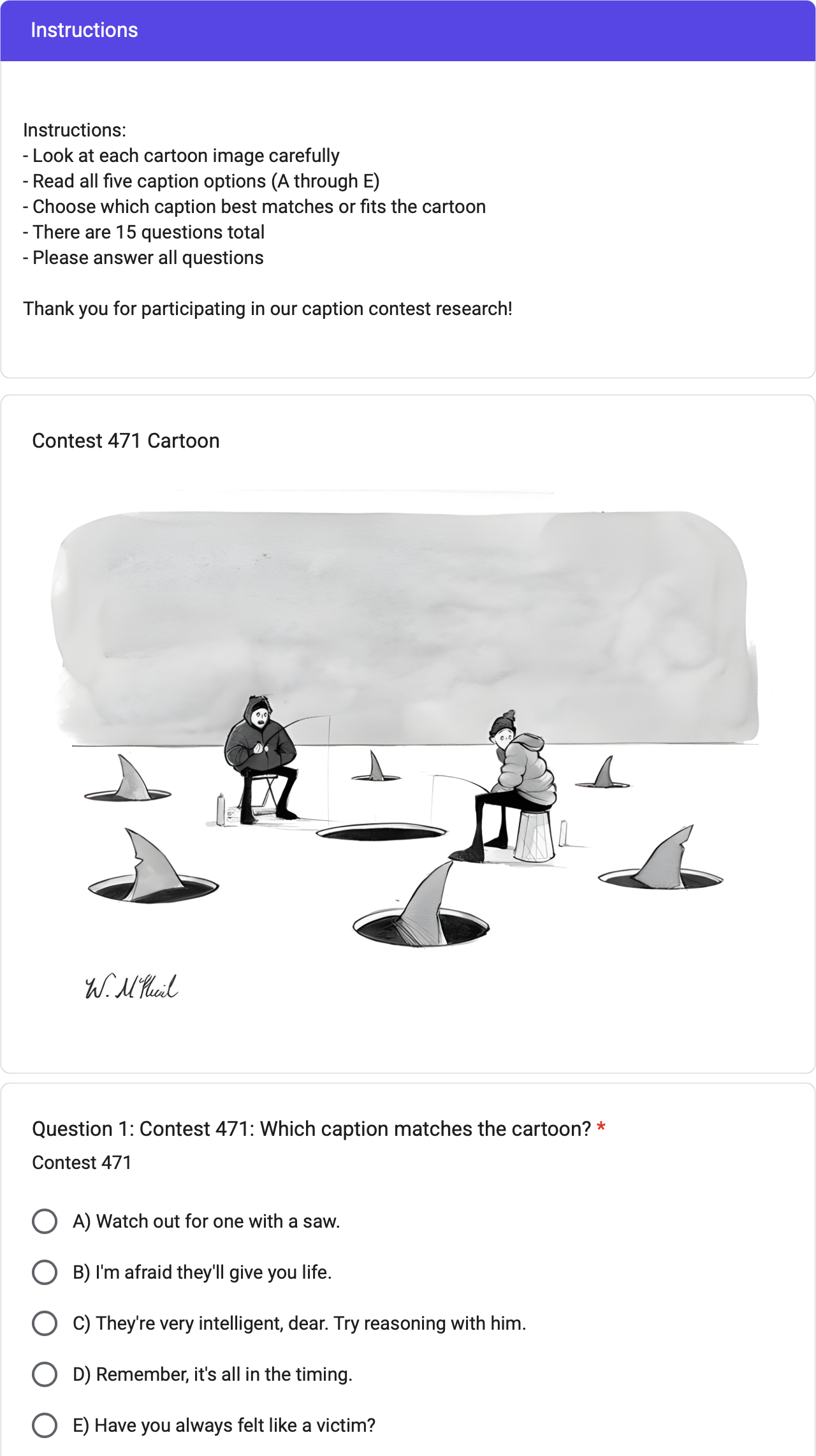}
    \caption{\textbf{Task instructions and an example matching question.}}
    \label{fig:user-study-question}
\end{figure}

\parahead{Instructions and interface} Participants completed the study through an online form. The full text of the instructions, disclaimers, and consent statement shown to participants, along with a representative matching question, is reproduced in Fig.~\ref{fig:user-study-consent} and~\ref{fig:user-study-question}. The instructions asked participants to view each cartoon carefully, read all candidate captions, and choose the one that best matches (matching task) or is funnier (ranking task). The matching and ranking interfaces were analogous, differing only in the number of candidate captions and the framing of the question. The study comprised 15 questions per task, and participants were asked to answer all of them.

\parahead{Recruitment and payment} The non-expert participants (21 in total) were recruited on a voluntary basis from among graduate students in our research group. Participation was entirely optional and unpaid, was not tied to course credit, compensation, or any form of academic evaluation, and participants were free to decline or withdraw at any time without consequence. As stated in the consent text shown to participants (Fig.~\ref{fig:user-study-consent}), the study was an entirely voluntary, unpaid academic research effort with no financial compensation. Because participation was voluntary and unpaid rather than conducted through a paid crowdsourcing platform, no payment-adequacy considerations relative to participant demographics apply. The task involved only minimal risk (standard computer use), as disclosed to participants. We note that, while these participants are not professional captionists, they have some familiarity with this line of research; we therefore interpret their results as a non-expert human reference rather than as a sample drawn from the general population.

\parahead{Expert reference} The expert human reference was provided by a domain expert with extensive professional experience evaluating New Yorker–style cartoon captions, who is a member of the research team for this work. The expert completed the same held-out matching and ranking items used in the automated evaluation, under the same conditions, with no access to gold captions or crowd preference statistics at the time of evaluation. Because the expert is affiliated with the authors, we report the expert results as a qualitative reference for what human-level humor reasoning looks like rather than as a controlled, independent population-level baseline.

\parahead{Consent and data use} Before beginning the task, participants were shown a consent statement explaining that their responses would be completely anonymized and used strictly for research purposes, including potential publication and open-science data sharing. By proceeding to the task, participants confirmed that they were at least 18 years old and consented to their anonymized data being used for this research. No personally identifying information was collected.

\section{LLM Usage}
\label{app:llm_usage}
LLMs were used to polish the writing and improve the clarity and flow of the text. All final revisions were reviewed and edited by the authors.

\end{document}